\documentclass{article}

\usepackage{microtype}
\usepackage{graphicx}
\usepackage{subcaption}
\usepackage{booktabs}
\usepackage{hyperref}
\usepackage{xurl}

\usepackage[accepted]{icml2026}

\usepackage{amsmath}
\usepackage{amssymb}
\usepackage{mathtools}
\usepackage{amsthm}

\usepackage[capitalize,noabbrev]{cleveref}

\usepackage{xcolor}
\usepackage{tcolorbox}
\usepackage{multirow}
\usepackage{pifont}
\usepackage{natbib}
\newcommand{\cmark}{\ding{51}}
\newcommand{\xmark}{\ding{55}}
\usepackage{placeins} 
\usepackage{float}
\usepackage{xspace}

\usepackage{makecell}

\theoremstyle{plain}
\newtheorem{theorem}{Theorem}[section]

\theoremstyle{definition}
\newtheorem{definition}[theorem]{Definition}
\newtheorem{assumption}[theorem]{Assumption}
\theoremstyle{remark}

\usepackage[textsize=tiny]{todonotes}

\icmltitlerunning{Revisiting Parameter-Based Knowledge Editing in Large Language Models: Theoretical Limits and Empirical Evidence}

\begin{document}

\twocolumn[
  \icmltitle{Revisiting Parameter-Based Knowledge Editing in Large Language Models: Theoretical Limits and Empirical Evidence}
  
  \icmlsetsymbol{equal}{*}

  \begin{icmlauthorlist}
    \icmlauthor{Wanying Ren}{equal,ECNU}
    \icmlauthor{Xin Song}{equal,ECNU}
    \icmlauthor{Futing Wang}{WU}
    \icmlauthor{Guoxiu He}{ECNU}
    \icmlauthor{Aixin Sun}{NTU}
  \end{icmlauthorlist}

  \icmlaffiliation{ECNU}{East China Normal University, Shanghai, China}
  \icmlaffiliation{WU}{Westlake University, Hangzhou, China}
  \icmlaffiliation{NTU}{Nanyang Technological University, Singapore}
  
  \icmlcorrespondingauthor{Guoxiu He}{gxhe@fem.ecnu.edu.cn}
  % \icmlcorrespondingauthor{Aixin Sun}{axsun@ntu.edu.sg}

  % You may provide any keywords that you find helpful for describing your
  % paper; these are used to populate the "keywords" metadata in the PDF but
  % will not be shown in the document
  % \icmlkeywords{Machine Learning, ICML}

  \vskip 0.3in
]

\printAffiliationsAndNotice{\icmlEqualContribution}

\begin{abstract}
Parameter-based knowledge editing updates the internal knowledge of large language models (LLMs) via localized weight modifications and has attracted significant attention. 
However, most existing methods overlook fundamental theoretical limitations and are rarely evaluated under realistic, practice-oriented settings. 
In this paper, we first present a theoretical analysis based on the dimensional Collapse Hypothesis, explaining how localized parameter edits can propagate along fragile directions in the representation space, inducing global interference and ultimately causing reasoning collapse. 
Building on this insight, we conduct a comprehensive empirical evaluation by systematically varying knowledge complexity, number of edits, evaluation dimensions, and baseline methods. 
Our results show that parameter-based editing methods consistently damage core LLM capabilities. 
In contrast, a simple retrieval-based baseline achieves consistently stronger performance than all parameter-editing methods across all evaluated conditions. 
These findings highlight that preserving the fundamental capabilities of LLMs after knowledge editing should be a central concern for future research. 
% \footnote{The code is available at \href{https://github.com/ECNU-Text-Computing/Revisiting-Parameter-Based-Knowledge-Editing}{GitHub Link}}
% The code is available at \href{https://github.com/ECNU-Text-Computing/Revisiting-Parameter-Based-Knowledge-Editing}{GitHub Link}.
% \href{https://github.com/ECNU-Text-Computing/Revisiting-Parameter-Based-Knowledge-Editing}{https://github.com/ECNU-Text-Computing/Revisiting-Parameter-Based-Knowledge-Editing}
% https://github.com/ECNU-Text-Computing/Revisiting-Parameter-Based-Knowledge-Editing

\end{abstract}

% \maketitle

\section{Introduction}
Large language models (LLMs) \citep{zeng2022glm, touvron2023llama, openai2023gpt} acquire extensive world knowledge \citep{jiang2020can, alkhamissi2022review, zhang2023large} and remarkable contextual reasoning abilities \citep{liu2023pre, lee2024reasoning} through large-scale pre-training and post-training \citep{brown2020language, ouyang2022training}. However, as world knowledge continuously evolves, some information encoded in LLMs inevitably becomes outdated or inaccurate \citep{mousavi2024your, ji2023survey}. To address this issue, knowledge editing methods \citep{sinitsin2020editable, zhumodifying} have been proposed to incorporate newly acquired knowledge into LLMs through minimal parameter modifications or external knowledge storage.

\begin{figure}[]
    \vskip 0.2in
    \centering
    \includegraphics[width=0.94\columnwidth]{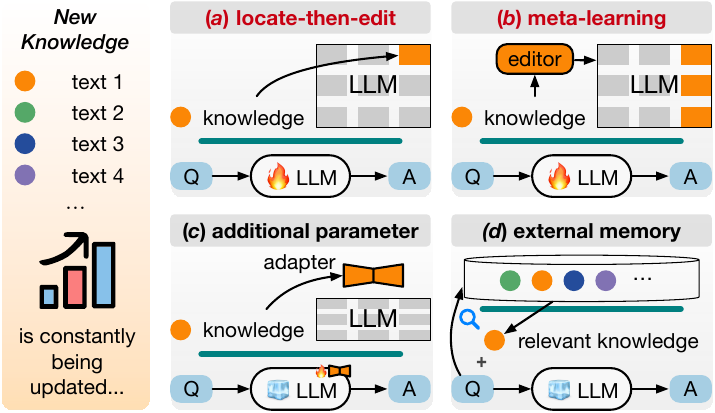}
    \caption{Training and inference workflows for four types of knowledge editing methods. The training process is shown above the \textcolor{teal}{green line}, while the inference stage is shown below. (a) and (b) perform knowledge editing by modifying the LLM parameters.} 
    \label{fig:model_intro}
    \vspace{-15pt}
\end{figure}

Knowledge editing methods can be broadly categorized into four main types, distinguished by how they adjust LLM parameters to incorporate collected textual knowledge. Figure \ref{fig:model_intro} illustrates their respective workflows during training and inference. (a) \textbf{Locate-then-edit} methods \citep{meng2022locating, mengmass, li2024pmet} assume that specific knowledge is associated with certain LLM parameters; they first \textit{locate} the neurons corresponding to the target knowledge and then \textit{edit} the relevant parameters to incorporate the new knowledge. (b) \textbf{Meta-learning}-based methods \citep{de2021editing, tanmassive} assume that the patterns of parameter changes during knowledge updates can be learned; an editor model is trained to modify the parameters in specific layers of the LLM associated with the target knowledge. (c) \textbf{Additional parameter}-based methods \citep{huangtransformer, yu2024melo, wang2024wise} assume that new knowledge can be stored in additional parameters external to the LLM; adapter layers or other auxiliary components are introduced to encode new knowledge while leaving the base LLM parameters unchanged. (d) \textbf{External memory}-based methods \citep{hartvigsen2024aging, mitchell2022memory} store updated knowledge as texts, embeddings, hidden states, or lightweight models; during inference, the most relevant information is retrieved to support LLM reasoning without modifying LLM parameters.

Methods that directly modify the internal parameters of LLMs have attracted substantial research attention, partly because they appear more principled, incur no additional inference-time overhead compared to the original model, and often achieve seemingly perfect knowledge updates under specific evaluation protocols. \textbf{From a theoretical perspective}, however, we argue that these advantages can be misleading. Although localized parameter edits may be effective for isolated factual updates, dimensional collapse in LLM representation spaces creates fragile low-variance directions where even small weight perturbations are disproportionately amplified. This amplification distorts representation geometry and propagates through subsequent layers, leading to catastrophic forgetting of unrelated knowledge, degraded generalization, and impaired reasoning. In this work, we develop a theoretical analysis in Section~\ref{sec:theo} to characterize the fundamental limitations of parameter-based knowledge editing methods.
\textbf{From an empirical perspective}, most existing studies have not fully evaluated knowledge editing methods in terms of LLMs' core capabilities and realistic application scenarios. While some work has highlighted deficiencies in specific evaluation dimensions, these discussions remain fragmented and have yet to form a widely accepted, systematic framework. LLMs encode not only isolated factual knowledge (\textit{e.g.}, triples) but also complex event-level knowledge, and their capabilities extend beyond answering simple facts to flexible knowledge application and deeper reasoning, such as reverse relation inference and multi-hop reasoning. Realistic editing scenarios should involve sequential, continuous updates rather than single-shot edits, and parameter-based editing methods should be systematically compared with external-knowledge-based paradigms (\textit{e.g.}, retrieval-augmented generation (RAG)). Empirical results are shown in Section \ref{sec:experiments}.

Our contributions are threefold \footnote{The code is available at \href{https://github.com/ECNU-Text-Computing/Revisiting-Parameter-Based-Knowledge-Editing}{GitHub Link}.}:
\begin{enumerate}
    \renewcommand{\labelenumi}{(\arabic{enumi})}
    % \item We develop a geometric theoretical framework explaining how localized parameter edits destabilize LLM representations.
    \item We develop a geometric theoretical framework that characterizes how localized parameter edits destabilize LLM representations.
    \item We perform a comprehensive empirical evaluation under realistic settings, showing that such edits consistently degrade core LLM capabilities.
    % \item We conduct a comprehensive and practical evaluation showing the resulting degradation of core LLM capabilities. % and demonstrating
    % \item Our results clarify the fundamental limitations of parameter-based knowledge editing and indicate that the choice of knowledge update mechanism should be guided by update scale.
    % \item Our results clarify the fundamental limitations of parameter-based knowledge editing and expose a fundamental stability--efficiency trade-off that should guide future knowledge update mechanisms.
    \item Our results clarify the fundamental limitations of parameter-based knowledge editing and highlight the stability--efficiency trade-off in the mechanisms.
\end{enumerate}

\section{Related Work}
\label{sec:related}
% 1. 已有研究指出知识编辑模型在追求单一事实准确性时，会引发涟漪效应、推理能力下降、模型崩溃等副作用，且现有评估方法可能高估其性能。  
% 2. 针对已有研究缺乏统一理论框架和评估标准的问题，本文提出“表征空间坍缩假说”并开展综合基准测试，覆盖复杂数据集和实际场景，以全面评估知识编辑方法的局限性。

% 1. 已有研究指出知识编辑模型在追求单一事实准确性时，会引发涟漪效应、推理能力下降、模型崩溃等副作用，且现有评估方法可能高估其性能。
% Knowledge editing has drawn increasing attention, with numerous methods proposed to improve the accuracy and efficiency of knowledge updates in LLMs 
Knowledge editing in LLMs has received increasing attention, with numerous methods aimed at improving accuracy and efficiency
\cite{li2025adaedit,fang2025cknowedit,scialanga2025sake,nie2025bmike,zhang2026conflict,liuedit}. While most work focuses on improving factual correctness for isolated updates, a few studies have also highlighted potential limitations of parameter-based editing. 
% For instance, \citet{pinter2023emptying} observed that optimizing for individual factual accuracy can conflict with the pre-training objectives of LLMs, and \citet{wang2024missing} quantified a \textit{ripple effect}, where local edits propagate through hidden representations and degrade overall model performance. Similarly, \citet{gu2024model,xie2025revealingdeceptiveness} showed that improvements in factual accuracy often come at the cost of reasoning, natural language inference, and question answering, and that even modifying just 1\% of parameters can harm unrelated tasks. 
% Other studies indicate that minor edits may reduce text coherence \citep{yang2024butterfly}, trigger model collapse in specific operations (\textit{e.g.}, ROME) \citep{yang2024fall}, amplify existing biases \citep{halevy2024flex}, or lead to overestimation of editing quality under teacher-forcing evaluation \citep{yang2025mirage}. Mechanistic analyses further reveal failure modes in locate-and-edit approaches and provide theoretical insights into key-value representations \cite{yan2024keys}. A more comprehensive discussion of related work can be found in Appendix \ref{sec:Arelated}.

% While most approaches focus on enhancing factual correctness for isolated updates, studies have identified limitations of parameter-based editing. 

For example, \citet{pinter2023emptying} noted conflicts between optimizing factual consistency and pre-training objectives, while \citet{wang2025missing} quantified the \textit{ripple effect} of local edits on hidden representations, leading to degraded model performance. Similarly, \citet{gu2024model,xie2025revealingdeceptiveness} observed that improving factual accuracy often undermines reasoning, natural language inference, and question answering, 
with edits to as little as 1\% of parameters harming unrelated tasks.
Furthermore, minor parameter modifications can reduce text coherence \citep{yang2024butterfly}, trigger model collapse in some specific methods such as ROME \citep{yang2024fall}, amplify existing biases \citep{halevy2024flex} or yield misleading evaluations under teacher-forcing settings \citep{yang2025mirage}. Mechanistic analyses have highlighted failure modes in locate-and-edit methods, offering theoretical insights into key-value representations \cite{yan2024keys}. A broader discussion appears in Appendix \ref{sec:Knowledge Editing Methods}.

% 2. 针对已有研究缺乏统一理论框架和评估标准的问题，本文提出“表征空间坍缩假说”并开展综合基准测试，覆盖复杂数据集和实际场景，以全面评估知识编辑方法的局限性。
While these studies have significantly enhanced empirical understanding of knowledge editing risks, their contributions remain largely phenomenological and scoped to specific phenomena: individual failure patterns, method variants, or isolated evaluation settings. 
They have not yet converged into a unified theoretical framework that explains why parameter edits can broadly compromise LLM performance. Nor have they established comprehensive and consistent evaluation standards (as summarized in Table \ref{tab:model_setting} in Appendix \ref{sec:sum_exp_setting}) that capture core LLM capabilities and realistic application scenarios.

% They have yet to propose a unified theoretical framework explaining systematic compromises in model capabilities or establish consistent evaluation standards for core LLM tasks and real-world applications (as summarized in Table \ref{tab:model_setting} in Appendix \ref{sec:sum_exp_setting}).

In contrast, this paper makes two complementary contributions, balancing theoretical analysis and empirical evaluation. 
First, we develop a mathematically grounded framework based on the dimensional Collapse Hypothesis, formally demonstrate that even minor parameter edits disrupt pre-training-induced knowledge geometry, resulting in systemic reasoning failures.
Second, 
% we conduct a comprehensive benchmark study that goes beyond prior benchmark studies \cite{chen2025uniedit}. Our evaluation systematically compares parameter-based and external-knowledge-based paradigms (such as an intuitive retrieval-augmented generation (RAG) baseline) across complex knowledge formats, reasoning-oriented LLMs, realistic multi-step editing scenarios, practical inference settings, and multiple performance metrics. 
We conduct a comprehensive benchmark comparing parameter-based and external-knowledge-based paradigms 
% (\textit{e.g.}, retrieval-augmented generation (RAG)) 
(\textit{e.g.}, RAG) 
across diverse knowledge formats, reasoning-oriented LLMs, multi-step editing scenarios, and practical inference tasks. This evaluation incorporates extensive metrics for robustness and capability preservation, surpassing prior benchmarks \cite{chen2025uniedit}.
Together, these contributions establish both theoretical underpinnings and empirical evidence for the limitations of parameter-based knowledge editing, uncovering potential oversights in critical application-relevant characteristics and 
% offering actionable insights for future research.
providing practical implications and actionable directions for future research.

\section{Preliminaries}
\label{sec:preliminaries}

% The goal of a triplet-based knowledge editing method $\mathcal{E}$ is to update a factual triple $e=(h, r, t)$ to a new triple $e'=(h, r, t_*)$, such as changing \textit{(Barack Obama, born in, Hawaii)} to \textit{(Barack Obama, born in, Kenya)}, within an original LLM $f^0$. The edited LLM $f^1$ is obtained by applying the editor to the original LLM and the target knowledge update:

Triplet-based knowledge editing focuses on updating a factual triple $e = (h, r, t)$ to a new triple $e'=(h, r, t_*)$ within an original language model $f^0$. For example, editing \textit{(Barack Obama, born in, Hawaii)} to \textit{(Barack Obama, born in, Kenya)}. The updated model $f^1$ is obtained by applying the editor $E$ to the initial model $f^0$ and the knowledge update $e'$:
\begin{equation}
       f^1 = \mathcal{E}(f^0, e'), \quad \text{such that} \ e' \in f^1
\end{equation}
Factual knowledge, such as triples, can be represented as a single prompt–answer pair $e=[(x, y)]$, whereas more complex event is represented using multiple prompt–answer pairs $e = [(x_1, y_1), \cdots, (x_m, y_m)]$, where $x$ denotes the prompt to elicit the knowledge, $y$ represents the answer, and $m$ denotes the number of pairs. 
% After editing the LLM $f^0$, it is expected that, given a knowledge-related prompt $x$, the LLM $f^1$ will produce the correct answer $y$.
After editing, the updated model $f^1$ should 
% produce correct answers $y$ for relevant prompts $x$.
outputs the ground truth $y$ for the corresponding prompts $x$.

\textbf{Editing Scenarios.} \textbf{Single Editing} refers to the process where the LLM updates only one piece of knowledge ($E = [e_1]$) at a time.
% The definition provided above corresponds to this single-editing scenario.
\textbf{Sequential Editing} also known as lifelong editing or continual editing \citep{hartvigsen2024aging,yu2024melo,wang2024wise}, refers to repeatedly applying the editing method to update a sequence of knowledge $E_t = [e_1, e_2, \ldots, e_t]$.
Consequently, the LLM evolves from $f^0$ to $f^t$ over multiple updates. Note that, in the sequential editing setting, each method operates differently: \textbf{locate-then-edit} methods modify a subset of the LLM's parameters with each update; \textbf{meta-learning}-based methods use a learned editor to update the parameters of the target model;
\textbf{additional parameter}-based methods repeatedly update external parameter components; \textbf{external memory}-based methods only update an external knowledge base, such as a text corpus or vector store, that stores newly introduced knowledge.

 \textbf{Evaluation Dimensions.} 
 % After incorporating $t$ pieces of new knowledge, \ie undergoing $t$ rounds of updates, the edited LLM $f^t$ is evaluated in four dimensions
 After $t$ updates, the edited model $f^t$ is evaluated across four dimensions
 ~\citep{zhang2024comprehensive}:

% \begin{itemize}
    % \item 
    $~~\bullet$ \textbf{Reliability:} 
    % The edited LLM should generate the updated target output for the prompts used for knowledge editing, \ie those available in $E_t$,  and ensure the persistence of the editing effects. This can be formally expressed as:
    The edited LLM should reliably produce the updated target outputs for the prompts in $E_t$, ensuring the persistence of the edits:
    \begin{equation}
        \mathbb{E}_{(x_i, y_i) \sim E_t} \mathbb{I} \{ \arg\max_y f^t(y \mid x_i) = y_i \}.
    \end{equation}

    % \item 
    $~~\bullet$ \textbf{Generalization:} The edited LLM should extend beyond the exact edits and correctly respond to paraphrased prompts, denoted as $N(E_t)$: 
    % Mathematically, this is formulated as:  
    \begin{equation}
        \mathbb{E}_{(x_i, y_i) \sim N(E_t)} \mathbb{I} \{ \arg\max_y f^t(y \mid x_i) = y_i \}.
    \end{equation}

    % \item 
    $~~\bullet$ \textbf{Locality:} 
    % It is imperative that the edited LLM should retain its original behavior when processing queries that are unrelated to the edits, denoted by $O(E_t)$. This criterion measures how well the edited LLM maintains the overall stability while keeping edits confined to the relevant scope. This requirement can be represented as:
    The edited LLM should retain its original behavior for queries unrelated to the edits, denoted $O(E_t)$, ensuring stability and minimal unintended changes:
    \begin{equation}
        \mathbb{E}_{(x_i, y_i) \sim O(E_t)} \mathbb{I} \{ f^t(y \mid x_i) = f^0(y \mid x_i) \}.
    \end{equation}

    % \item 
    $~~\bullet$ \textbf{Portability:} 
    % Furthermore, 
    % The edited LLM should effectively propagate the impact of the edited knowledge, correctly reasoning about its downstream implications
    % % , denoted by
    % $D(E_t)$, which encompasses three aspects: substituting the subject of the question with aliases, reasoning based on factual changes, and knowledge derived from reverse relationships. As shown in Table \ref{tab:model_setting}, this dimension has been ignored by a large number of studies. The requirement is defined as:
    The edited LLM should propagate the edits effectively, enabling downstream reasoning $D(E_t)$, including handling aliases, causal relationships, and reverse relations. Despite its importance, this dimension has been overlooked in many studies (shown in Table \ref{tab:model_setting}):
    \begin{equation}
        \mathbb{E}_{(x_i, y_i) \sim D(E_t)} \mathbb{I} \{ \arg\max_y f^t(y \mid x_i) = y_i \}.
    \end{equation}

\textbf{Evaluation for General Tasks.} 
In addition to the knowledge editing evaluation dimensions, it is also important to assess whether the edited LLM retains its ability to handle general tasks. 
% In this study, we use a mathematical reasoning dataset to evaluate the inference performance of the edited model. If the edited LLM satisfies these requirements, the knowledge editing process is considered successful, as it achieves the intended knowledge modification while preserving the LLM's core capabilities, including generalization, reasoning, and performance on unrelated knowledge.
% Beyond knowledge editing evaluation, it’s essential to assess whether $f^t$ retains general task capabilities. 
In this study, mathematical reasoning datasets are utilized to evaluate the edited model’s reasoning capabilities. A successful editing process achieves precise targeted updates while preserving the LLM’s core capabilities, including
generalization, reasoning, and performance on unrelated knowledge.

\section{Theoretical Analysis}
\label{sec:theo}
% 1. 本节基于表征空间几何结构提出理论框架，分析局部参数编辑通过误差放大机制损害模型泛化能力的原理，分三步推导表征空间脆弱性。
% 2. 通过维度坍缩、小扰动和表征稳定性三大假设形式化问题，定义编辑前后权重与隐向量关系，量化表征空间维度坍缩特征（有效秩极小）。
% 3. 提出坍缩方向扰动放大定理，证明微小编辑扰动在低方差主成分方向会被极端放大，导致语义扭曲和推理崩溃。
% 4. 连续编辑场景下，坍缩方向的微小扰动会线性累积放大，加剧语义扭曲风险，导致泛化能力随编辑次数增加而加速恶化。

This section develops a geometric framework for understanding why parameter-based knowledge editing can induce disproportionate side effects in large language models. Our analysis is motivated by a simple observation: if hidden representations are dimensionally collapsed, then some representation directions have extremely weak native signal scales, making them vulnerable to even localized parameter perturbations. 

We formalize this intuition in four steps. First, we introduce the setup and the working assumptions under which the mechanism is analyzable (Section~\ref{sec:Hypotheses-and-Formulation}). Second, we define a relative-sensitivity metric and show why perturbations can be strongly amplified in low-singular-value directions (Section~\ref{sec:Perturbation-Analysis}). Third, we discuss how such distortions accumulate under sequential editing and why practical behavior may become even worse once representation stability breaks down (Section~\ref{sec:Cumulative Effects in Sequential Editing}).
Finally, we bridge the representation-level analysis to model performance by exploring how excessive distortions correlate with downstream capability degradation, providing empirical evidence for the practical significance of this geometric vulnerability (Section~\ref{sec:From Representation Distortion to Capability Degradation}).

% Importantly, our goal is not to provide an exact end-to-end characterization of all edited LLM dynamics. Rather, we isolate a minimal vulnerability mechanism: even in a locally stable regime, dimensional collapse already makes edited representations disproportionately fragile in low-variance directions.

Our goal is not to fully characterize edited LLM dynamics but to isolate a minimal vulnerability: dimensional collapse makes representations fragile in low-variance directions, even in locally stable regimes.

% ===============================================================
\subsection{Working Hypotheses and Problem Formulation}
\label{sec:Hypotheses-and-Formulation}
% 通过维度坍缩、小扰动和表征稳定性三大假设形式化问题，定义编辑前后权重与隐向量关系，量化表征空间维度坍缩特征（有效秩极小）。
% ===============================================================
We consider a feed-forward network ($FFN$) layer in a Transformer model:

\begin{equation}
h=\phi(Wx),
\end{equation}

where
$x\in \mathbb{R}^{d_{in}}$ is the input vector, 
$W \in \mathbb{R}^{d \times d_{in}}$ is the weight matrix, 
$\phi:\mathbb{R}^d \to \mathbb{R}^d$ is an elementwise nonlinearity, and 
$h \in \mathbb{R}^d$ is the resulting hidden representation.

A parameter-editing operation modifies the layer locally by adding a perturbation $\triangle W$, producing
$W_{edited}=W+\triangle W$.

Given a set of $n$ input samples, let
$H=[h^{(1)},h^{(2)},\ldots,h^{(n)}]\in \mathbb{R}^{d \times n}$
denote the centered hidden representation matrix at this layer. We perform singular value decomposition: $H=U\Sigma V^T$.

where $U=[u_1,u_2,\ldots,u_d]\in \mathbb{R}^{d\times d}$ is an orthonormal basis of principal directions, and the singular values are ordered as
$\sigma_1 \geq \sigma_2 \geq \cdots \geq \sigma_d \geq 0.$

\begin{assumption}[Dimensional Collapse Hypothesis]
\label{hypo:1}
We assume that the hidden representation space of the pre-trained model exhibits dimensional collapse in a spectral sense: its singular-value spectrum is highly anisotropic, with a strongly compressed low-variance tail. 
% In particular, let $\sigma_{\min}$ denote the smallest non-zero singular value of the representation matrix. When $\sigma_{\min}$ is very small, any principal direction associated with it has an extremely weak native signal scale.
Let $\sigma_{\min}$ be the smallest non-zero singular value. When $\sigma_{\min}$ is small, the associated direction has a weak signal scale.

\end{assumption}

% This assumption is motivated by our empirical measurements of hidden-state spectra across multiple layers, which show low effective rank, strongly decaying tail singular values, and large non-zero-subspace condition numbers. As a representative example, Table~\ref{tab:dimensional_collapse_main} shows this behavior in Llama-3.1-8B-Instruct; broader results across model scales and architectures are reported in Appendix~\ref{sec:Hypothesis 1} (Tables~\ref{tab:dimensional_collapse_appendix_llama31}--\ref{tab:dimensional_collapse_t5_mend}).

% This assumption is supported by empirical measurements, which show low effective rank, rapidly decaying singular values, and large condition numbers (e.g., Llama-3.1-8B-Instruct in Table~\ref{tab:dimensional_collapse_main}; more results in Appendix~\ref{sec:Hypothesis 1}).

This assumption is supported by empirical measurements, which show low effective rank, rapidly decaying singular values, and large condition numbers (e.g., Llama-3.1-8B-Instruct in Table~\ref{tab:dimensional_collapse_main}). A detailed evaluation of these characteristics is provided in Appendix~\ref{sec:Hypothesis 1} (Tables~\ref{tab:dimensional_collapse_appendix_llama31}--\ref{tab:dimensional_collapse_t5_mend}). 

% \textbf{Table 1: Empirical evidence of dimensional collapse in the hidden representations of Llama-3.1-8B-Instruct.}

\begin{table*}[htbp]
    \centering
    \caption{Empirical evidence of dimensional collapse in the hidden representations of Llama-3.1-8B-Instruct. 
    The hidden representation space exhibits clear characteristics of dimensional collapse: the effective rank $r$ is significantly smaller than the feature dimension, the smallest singular value $\sigma_{\min}$ is extremely small ($\sim 10^{-5}$), and the condition number is very large ($>10^6$). 
    % These empirical patterns support the general validity of the dimensional collapse assumption.
    }
    \label{tab:dimensional_collapse_main}
    \begin{tabular}{c|cccrcc}
        \toprule
        layer & $d$ & $r_{\text{eff}}$ & $r/d$ (\%) & $\sigma_{1}$ & $\sigma_{\min}$ & $\text{cond}(\sigma_{1}/\sigma_{\min})$ \\
        \midrule
        5  & 4096 & 3249 & 79.3 & 194.1      & $3.38\times10^{-6}$ & $5.74\times10^{7}$ \\
        10 & 4096 & 3099 & 75.7 & 302.6      & $7.55\times10^{-6}$ & $4.01\times10^{7}$ \\
        15 & 4096 & 3177 & 77.6 & 504.9      & $1.20\times10^{-4}$ & $4.19\times10^{6}$ \\
        20 & 4096 & 3258 & 79.6 & 537.5      & $7.16\times10^{-6}$ & $7.50\times10^{7}$ \\
        25 & 4096 & 3260 & 79.6 & 618.7      & $9.48\times10^{-6}$ & $6.53\times10^{7}$ \\
        30 & 4096 & 2066 & 50.4 & 4922.3     & $3.85\times10^{-5}$ & $1.28\times10^{8}$ \\
        31 & 4096 & 1023 & 25.0 & 13003.3    & $3.05\times10^{-3}$ & $4.26\times10^{6}$ \\
        \bottomrule
    \end{tabular}
\end{table*}

This setup allows us to ask a precise question: when an edit induces a change in hidden representations, how large is that change relative to the native signal scale of each principal direction?

\begin{assumption}[Small Perturbation Hypothesis]
\label{hypo:2}
A single editing operation induces a localized parameter perturbation $\triangle W$ whose norm is sufficiently small for a first-order Taylor approximation of the induced representation change to be informative.
\end{assumption}

For single-edit regimes, the induced hidden-state perturbation norms remain relatively small in many edited layers across several parameter-based editing methods, supporting the use of local first-order analysis as a descriptive approximation. Full experimental results validating this hypothesis are provided in Appendix~\ref{sec:Hypothesis 2} (Tables~\ref{tab:perturbation_norms}--\ref{tab:perturbation_t5_mend}).

% For single-edit regimes, the induced hidden-state perturbation norms remain relatively small in many edited layers across several parameter-based editing methods. This supports the validity of Hypothesis~\ref{hypo:2}, which posits that the small-perturbation condition enables the first-order approximation in Eq.~(\ref{eq:taylor_approx}) to effectively describe the underlying behavior. Full experimental results validating this hypothesis are provided in Appendix~\ref{sec:Hypothesis 2} (Tables~\ref{tab:perturbation_norms}--\ref{tab:perturbation_t5_mend}).

\begin{assumption}[Local Representation Stability Hypothesis]
\label{hypo:3}
Within a short local editing regime, the principal directions $u_k$ of the original representation space remain sufficiently stable to serve as an approximate common basis for analyzing perturbations. This should be interpreted as a local analytical idealization rather than a global claim, especially because low-variance directions may drift rapidly under sequential editing.
\end{assumption}

% Empirically, dominant directions often remain comparatively stable in early edits, whereas low-variance components deteriorate much faster; see Appendix~\ref{sec:Hypothesis 3} (Tables~\ref{tab:pc_stability}--\ref{tab:pc_stability_t5_output}).

Empirical observations (Appendix~\ref{sec:Hypothesis 3}, Tables~\ref{tab:pc_stability}--\ref{tab:pc_stability_t5_output}) suggest that dominant directions show higher stability under early sequential editing, while low-scale components deteriorate rapidly.

These hypotheses are not limited to a single model scale or family. A broader set of empirical measurements across different model sizes and architectures is provided in Appendix \ref{sec:Generalization Across Model Scales and Architectures}, confirming the recurrence of the same spectral signatures and vulnerability mechanisms.
% ===============================================================
\subsection{Relative Perturbation Amplification in Collapsed Directions}
\label{sec:Perturbation-Analysis}
% 提出坍缩方向扰动放大定理，证明微小编辑扰动在低方差主成分方向会被极端放大，导致语义扭曲和推理崩溃。
% ===============================================================

For a given input x, the representation change induced by editing is approximated by a first-order Taylor expansion (per Hypothesis~\ref{hypo:2}):
\begin{align}
\triangle h \stackrel{\triangle}{=} h_{\text{edited}} - h &= \phi ((W+\triangle W)x) - \phi(Wx) \\
% \triangle h 
&\approx J_{\phi}(a) \cdot \triangle W \cdot x,
\label{eq:taylor_approx}
\end{align}

where $a=Wx$ and $J_{\phi}(a)=\operatorname{diag}(\phi'(a_1),\dots,\phi'(a_d))\in \mathbb{R}^{d\times d}$ is the Jacobian matrix of activation function $\phi$ at $a$. 

Under Hypothesis~\ref{hypo:3}, we project \(\Delta h\) onto the original principal basis \(\{u_k\}_{k=1}^d\):

\begin{equation}
\Delta h = \sum_{k=1}^{d} c_k u_k, 
\label{eq:h_delta_sum}
\end{equation}

where
$c_k = u_k^T \Delta h$
is the perturbation component along direction \(u_k\).

To quantify the native signal scale of the unedited model in direction $u_k$, we consider the root mean square (RMS) magnitude across the sample set:
\begin{equation}
\mathrm{RMS}(h_k)\stackrel{\triangle}{=} \sqrt{\frac{1}{n}\sum_{i=1}^n(u_k^Th^{(i)})^2}=\frac{\sigma_k}{\sqrt{n}}.
\end{equation}
% Thus, for directions with non-zero singular value, $\sigma_k$ directly determines the typical pre-edit scale of the original representation in direction $u_k$.
For directions with non-zero $\sigma_k$, it determines the typical pre-edit scale of $u_k$.

% Projecting $\triangle h$ onto the original principal component basis $\{u_k\}$ (per Hypothesis~\ref{hypo:3}) yields:
% \begin{equation}
% \triangle h=\sum_{k=1}^dc_ku_k,
% \end{equation}
% where $c_k=u_k^T\triangle h$ quantifies the projection strength of the editing perturbation onto direction $u_k$.

\begin{definition}[Directional Relative Change Rate]
% We define the \textbf{directional relative change rate} induced by editing along principal direction $u_k$ as
The relative change rate induced along direction $u_k$ is defined as
\begin{equation}
    R_k\stackrel{\triangle}{=} \frac{c_k}{\mathrm{RMS}(h_k)}=\frac{\sqrt{n}|c_k|}{\sigma_k}.
\end{equation}
\end{definition}

% The quantity $R_k$ measures the size of the editing-induced perturbation relative to the native signal scale of direction $u_k$. It is therefore a relative sensitivity metric: a direction can exhibit large $R_k$ either because the perturbation component \(|c_k|\) is large or because the native signal scale \(\sigma_k\) is small.
$R_k$ quantifies the perturbation size relative to the native signal scale of direction $u_k$. A large $R_k$ arises from either a large $c_k$ or a small $\sigma_k$.
When \(R_k \ll 1\), the edit remains small compared to the typical pre-edit magnitude in that direction. In contrast, when \(R_k \gtrsim 1\), the perturbation becomes comparable to or larger than the direction’s native signal scale, indicating substantial distortion.

\begin{theorem}[Relative Amplification in Low-Scale Directions]
\label{theo:Perturbation Amplification Theorem}
Let $u_{\min}$ denote a principal direction associated with the smallest non-zero singular value $\sigma_{\min}$. If the projection of the editing-induced perturbation onto this direction has magnitude $|c_{\min}|=\varepsilon$, then its directional relative change rate is
\begin{equation}
R_{\min}=\frac{\sqrt{n}\,\varepsilon}{\sigma_{\min}}.
\label{eq:R_min}
\end{equation}

More generally, for any principal direction with $\sigma_k>0$,
\begin{equation}
R_k=\frac{\sqrt{n}|c_k|}{\sigma_k}.
\label{eq:R_k}
\end{equation}

Hence, for perturbation components of comparable absolute magnitude across directions, the induced relative distortion is larger in directions with smaller singular values. In particular, when $\sigma_{\min}$ is extremely small, even a modest local perturbation can produce a very large relative distortion.
\end{theorem}

This theorem highlights a critical vulnerability: representation distortions are disproportionately amplified in directions with small $\sigma_k$ irrespective of input perturbation energy.
This concerns relative fragility, not semantic importance. We do not assume low-singular-value directions are meaningful or that editing concentrates perturbation energy in them. The claim is simply that any perturbation component landing there becomes disproportionately large relative to the native signal.

% This pattern is borne out by our measurements. The largest relative change rates concentrate in directions with small singular values. Moreover, even a single edit often produces $R_k>1$ in a large fraction of directions, indicating that local updates can already exceed the native signal scale of the original representation (see also the full distributions in Appendix
% ~\ref{sec:Direct Empirical Validation of the Amplification Mechanism}
% , 
% Tables~\ref{tab:relative_change_rates}--\ref{tab:rk_distribution_t5}).

% This pattern is corroborated by our empirical measurements: the largest relative change rates concentrate in directions with small singular values, and a single edit can often yields $R_k > 1$ in many directions, exceeding the native signal scale of the original representation. Full empirical validations, including visualizations and statistics of $R_k$, are provided in Appendix~\ref{sec:Direct Empirical Validation of the Amplification Mechanism} (Figure~\ref{fig:R_k_5method_3model}, Tables~\ref{tab:relative_change_rates}--\ref{tab:rk_distribution_t5}).

\begin{figure}[htbp]
    % \centering
    % \fbox{\parbox{0.7\columnwidth}{\centering\vspace{2em} Figure to be inserted: the amplification effect of $R_k$ values along small-singular-value directions.\vspace{2em}}}
    \vskip 0.2in
    \centering
    \includegraphics[width=0.94\columnwidth]{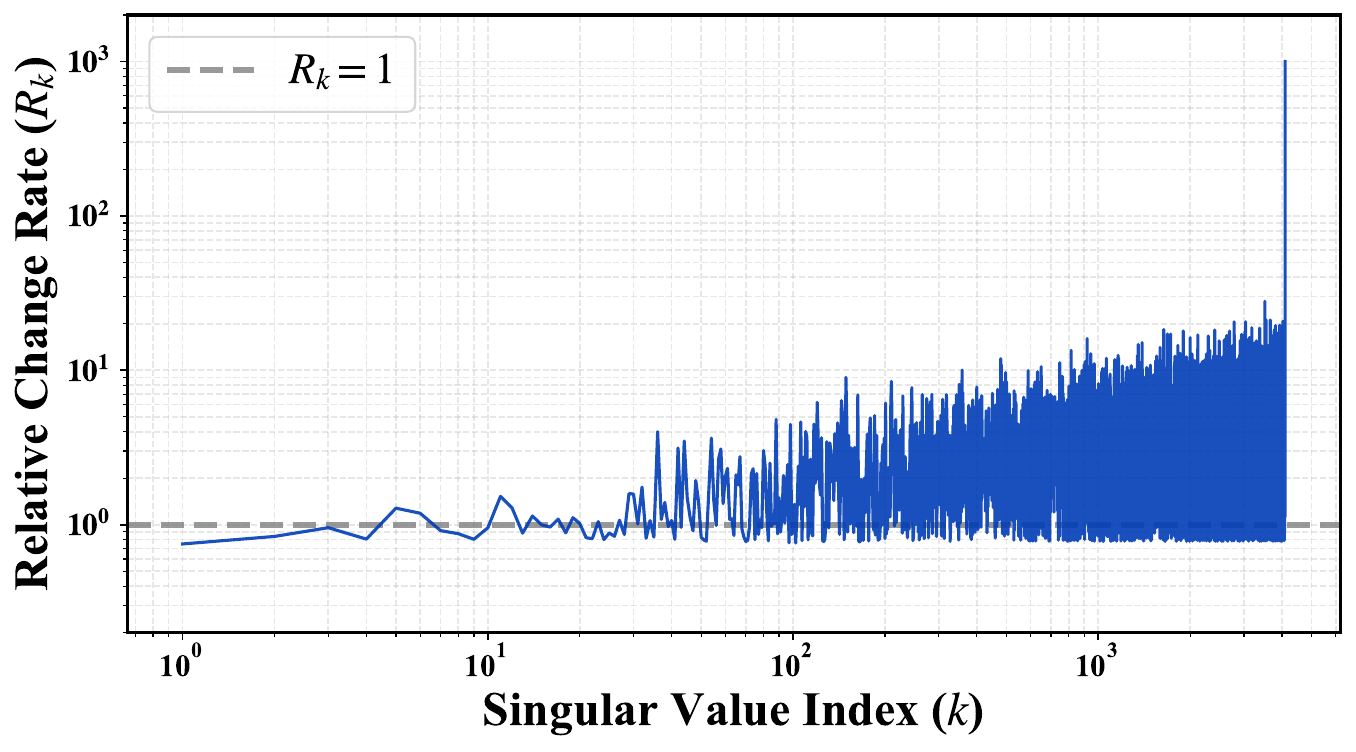}
    \caption{
    % The amplification effect of $R_k$ values is most pronounced along directions with small singular values, which corroborates the theoretical prediction of Theorem~\ref{theo:Perturbation Amplification Theorem}.
    Distribution of $R_k$ values in layer 30 of Llama-3.1-8B-Instruct  after performing 1000 sequential edits using MEMIT.
    }
    \label{fig:rk_amplification}
\end{figure}

This amplification effect, where $R_k$ is most pronounced along directions with small singular values, is empirically supported by Figure~\ref{fig:rk_amplification}, which shows the distribution of $R_k$ in layer 30 of Llama-3.1-8B-Instruct after performing 1000 sequential edits using MEMIT. Specifically, this observation aligns with our empirical measurements: the largest relative change rates concentrate in directions with small singular values, and a single edit can often yield $R_k > 1$ in many directions, exceeding the native signal scale of the original representation. Full empirical validations, including visualizations and statistics of $R_k$, are provided in Appendix~\ref{sec:Direct Empirical Validation of the Amplification Mechanism} (Figure~\ref{fig:R_k_5method_3model}, Tables~\ref{tab:rk_distribution_llama3}--\ref{tab:rk_distribution_t5}).

% ===============================================================
% \subsection{From Representation Distortion to Capability Degradation}
% \label{From Representation Distortion to Capability Degradation}
% 

\begin{table*}[t]
\centering
\caption{
Spearman correlation between step-level amplification and editing performance, using the \textbf{median} of the layer-wise $R_k$ distribution and \textbf{mean} aggregation across edited layers. 
% Performance values equal to zero are retained. 
Entries report Spearman's $\rho$ with significance markers, followed by the corresponding $p$-value in parentheses. 
\textit{Note.} $^{*}p<0.05$, $^{**}p<0.01$, $^{***}p<0.001$.
}
\label{tab:functional_relevance_main}
\small
\setlength{\tabcolsep}{7pt}
\begin{tabular}{l|ccccc}
\toprule
Metric & AlphaEdit & MEMIT & 
% MEND & 
ROME & WISE \\
\midrule
% Reliability    & $-0.2034$ $(0.5996)$ & $-0.9619^{***}$ $(<0.001)$ & $-0.8660$ $(0.3333)$ & $-0.8645^{**}$ $(0.0026)$ & $-0.7679^{*}$ $(0.0157)$ \\
Rel.    & $-0.2034$ $(0.5996)$ & $-0.9619^{***}$ $(<0.001)$ & $-0.8645^{**}$ $(0.0026)$ & $-0.7679^{*}$ $(0.0157)$ \\
% Generalization & $ 0.6360$ $(0.0656)$ & $-0.4238$ $(0.2557)$      & $-0.8660$ $(0.3333)$ & $-0.8645^{**}$ $(0.0026)$ & $-0.0840$ $(0.8298)$ \\
Gen. & $ 0.6360$ $(0.0656)$ & $-0.4238$ $(0.2557)$      & $-0.8645^{**}$ $(0.0026)$ & $-0.0840$ $(0.8298)$ \\
% Locality       & $ 0.0339$ $(0.9310)$ & $-0.7289^{*}$ $(0.0259)$ & $-0.8660$ $(0.3333)$ & $-0.9129^{***}$ $(<0.001)$ & $-0.4851$ $(0.1857)$ \\
Loc.       & $ 0.0339$ $(0.9310)$ & $-0.7289^{*}$ $(0.0259)$ & $-0.9129^{***}$ $(<0.001)$ & $-0.4851$ $(0.1857)$ \\
% Portability    & $ 0.3500$ $(0.3558)$ & $-0.8087^{**}$ $(0.0083)$ & $-0.8660$ $(0.3333)$ & $-0.8617^{**}$ $(0.0028)$ & $-0.6639$ $(0.0512)$ \\
Port.    & $ 0.3500$ $(0.3558)$ & $-0.8087^{**}$ $(0.0083)$ & $-0.8617^{**}$ $(0.0028)$ & $-0.6639$ $(0.0512)$ \\
\bottomrule
\end{tabular}

% \vspace{0.5em}
% \raggedright
% \footnotesize
% \textit{Note.} $^{*}p<0.05$, $^{**}p<0.01$, $^{***}p<0.001$.
\end{table*}

\subsection{Cumulative Effects in Sequential Editing}
\label{sec:Cumulative Effects in Sequential Editing}
% 连续编辑场景下，坍缩方向的微小扰动会线性累积放大，加剧语义扭曲风险，导致泛化能力随编辑次数增加而加速恶化。
% ===============================================================

We now extend the analysis to sequential editing, where the model undergoes a series of updates:
$W_{(0)}\to W_{(1)}\to \dots \to W_{(T)}$. 

Let \(h^{(t)}\) denote the hidden representation after the \(t\)-th edit. The total representation change after \(T\) edits relative to the original model satisfies the telescoping identity
\begin{equation}
\triangle h^{(T)}=h^{(T)}-h^{(0)}= \sum_{t=1}^T\triangle h_{\text{instant}}^{(t)},
\end{equation}

where $\triangle h_{\text{instant}}^{(t)}=h^{(t)}-h^{(t-1)}$ is the instantaneous change induced by the $t$-th edit.

Under the local stability approximation of Hypothesis~\ref{hypo:3}, we use the original principal basis as an approximate common coordinate system across edits. Projecting onto a principal direction \(u_{\min}\) associated with the smallest non-zero singular value \(\sigma_{\min}\), and defining $c_{\min}^{(t)} \triangleq u_{\min}^T \Delta h_{\text{inst}}^{(t)}$,

we obtain
\begin{equation}
c_{\min}^{(T)}
=
u_{\min}^T \Delta h^{(T)}
=
\sum_{t=1}^{T} c_{\min}^{(t)}.
\label{eq:c_cumulative}
\end{equation}

% In a coherent worst-case scenario where the perturbations in that direction remain similarly oriented across edits, we obtain

In a worst-case scenario, where edits produce coherent accumulations of perturbations, we obtain:
\begin{equation}
|c_{\min}^{(T)}|
\approx
T\cdot \bar{\varepsilon},
\label{eq:c_worstcase}
\end{equation}

where \(\bar{\varepsilon}\) is the average per-edit perturbation magnitude along \(u_{\min}\). Substituting into Eq.~(\ref{eq:h_delta_sum}) yields

\begin{equation}
R_{\min}^{(T)}
=
\frac{\sqrt{n}\,|c_{\min}^{(T)}|}{\sigma_{\min}}
\approx
T \cdot \frac{\sqrt{n}\,\bar{\varepsilon}}{\sigma_{\min}}
=
T\cdot R_{\min}^{(1)}.
\label{eq:R_min_sequential}
\end{equation}

% Thus, under this idealized coherent local regime, the relative distortion along the weakest-scale directions grows approximately linearly with the number of edits.

Under this idealized coherent local regime, distortion in weak directions grows linearly with the number of edits.

% However, this linear accumulation law should be interpreted carefully. It describes an idealized local-stability regime in which the principal directions remain approximately fixed over the editing sequence. In practice, this approximation degrades as edits accumulate. 
This linear law assumes stable principal directions across edits, but in practice, this approximation breaks down as edits accumulate.
Our empirical measurements show that principal directions, especially in the low-variance tail, drift rapidly under sequential editing across multiple models and architectures (Appendix~\ref{sec:Hypothesis 3}, Tables~\ref{tab:pc_stability}--\ref{tab:pc_stability_t5_output}). Therefore, practical sequential editing does not merely accumulate perturbations in a fixed vulnerable basis; it can also reorganize the basis itself, inducing broader representational instability.
This distinction strengthens, the main message of the theory. The fixed-basis analysis shows that cumulative fragility already arises even in a favorable local regime. Once basis drift and representation shattering are taken into account, practical degradation can be even more severe than this idealized prediction suggests.

Taken together, Sections~\ref{sec:Perturbation-Analysis} and~\ref{sec:Cumulative Effects in Sequential Editing} suggest the following picture: dimensional collapse creates directions with extremely small native signal scales; small local edits are therefore disproportionately amplified in relative terms; and repeated editing compounds this vulnerability, eventually destabilizing the broader representation geometry. 
% A broader set of empirical measurements across further model scales and architectures is provided in Appendix~\ref{sec:Generalization Across Model Scales and Architectures}, confirming the recurring presence of the same spectral signatures.

\subsection{From Representation Distortion to Capability Degradation}
\label{sec:From Representation Distortion to Capability Degradation}

The analysis above is a representation-level statement. 
Large $R_k$ does not directly imply behavioral failure but signifies structural risk. When $R_k$ is large, the post-edit representation in $u_k$ exceeds its native pre-edit scale.

To connect this to behavior, note that downstream computations depend not only on absolute activation magnitudes, but also on representation geometry, including relative positioning, subspace structure, and routing through subsequent nonlinear layers. Large relative distortions can therefore alter intermediate geometry in ways that propagate forward and potentially affect task behavior.

% This interpretation is empirically supported by the significant correlations between amplification and editing performance. As shown in Table~\ref{tab:functional_relevance_main}, larger amplification factors are associated with worse reliability, locality, and portability for several parameter-based editing methods. The experiment setting and significance table are reported in Appendix~\ref{sec:Functional Relevance} (Table~\ref{tab:correlation_pvalues}).

This interpretation is further supported by the correlation analysis between step-level $R_k$-based statistics and sequential editing performance. 
Table~\ref{tab:functional_relevance_main} reports the main Spearman correlation results, showing that larger amplification is generally associated with worse reliability, locality, and portability for several parameter-based editing methods. 
Full experimental details and complete results across summary and aggregation choices are provided in Appendix~\ref{sec:functional_relevance}.
(Table~\ref{tab:fr_median_full}--\ref{tab:fr_trimmed_mean_full}).
% ==================================================================
% \section{Experiments and Results}
\section{Empirical Study}
\label{sec:experiments}
% 5.1 实验设置
    % 模型选择（Llama系列、Mistral、DeepSeek-R1推理模型）
    % 12种知识编辑方法分类（参数修改类 vs 非参数类）
    % 引入SCR（选择性上下文推理）基线方法
    % 数据集（事实型、事件型、通用推理任务）
    % 评估指标与设置（自回归推理、单次/连续编辑）
% 5.2 实验结果分析
    % RQ1 实用场景性能：参数编辑法在连续编辑中崩溃，SCR全面占优
    % RQ2 推理能力影响：多数参数编辑法导致推理崩溃，SCR表现稳定
    % RQ3 事件知识泛化：参数编辑法在复杂事件型任务中表现极差
    % RQ4 时间效率对比：参数法编辑慢但推理快，非参数法编辑快但推理有延迟
% 核心发现：
% 参数编辑法在现实条件下表现普遍不佳
% SCR作为非参数基线方法在各项测试中全面领先
% 揭示参数编辑法的理论缺陷在实际应用中的具体表现
% ==================================================================
\begin{figure*}[!t]
  \vskip 0.2in
  \begin{center}
    % \centerline{\includegraphics[width=0.95\textwidth]{fig_scale_llama3.pdf}}
    \centerline{\includegraphics[width=1\textwidth]{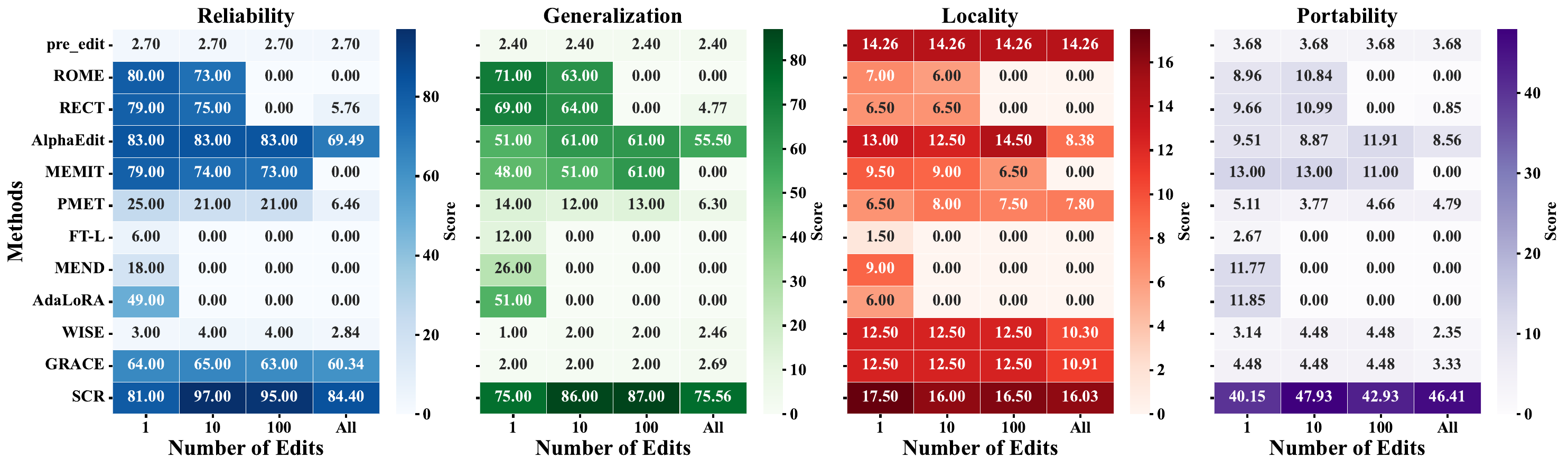}}
    \caption{
      Performance changes of knowledge editing methods during sequential editing of Llama-3.1-8B-Instruct on the ZsRE dataset. The x-axis represents the number of edits: 1, 10, 100, and `All' for the full dataset. 
      % Note that MEMIT and PMET are excluded, as they fail to run on Llama-3.1-8B-Instruct.
    }
    \label{fig:scale_llama3}
  \end{center}
  \vspace{-26pt}
\end{figure*}
% ==============================================================
\subsection{Experimental Settings}
\label{sec:experimental_settings}
% ===============================================================

We evaluate on Llama-2-7B-Chat~\citep{touvron2023llama}, Llama-3.1-8B-Instruct~\citep{llama3modelcard}, Mistral-7B-Instruct~\citep{jiang2023mistral}, and DeepSeek-R1-Distill-LLaMA-8B~\citep{guo2025deepseek}, with larger-scale verification on Llama-2-13B and Qwen3-14B. The studied parameter-modifying methods span locate-then-edit (ROME~\citep{meng2022locating}, MEMIT~\citep{mengmass}, PMET~\citep{li2024pmet}, RECT~\citep{gu2024model}, AlphaEdit~\citep{fang2024alphaedit}, FT-L), meta-learning (MEND~\citep{de2021editing}), and additional-parameter methods (AdaLoRA~\citep{zhang2023adalora}, WISE~\citep{wang2024wise}); external memory baselines include GRACE~\citep{hartvigsen2024aging} and SCR~\citep{he2025knowledge}. 
% We use ZsRE~\citep{levy2017zero}, WikiData$_\text{counterfact}$~\citep{cohen2024evaluating}, and the event-level ELKEN~\citep{peng2024event} datasets. 
We use the ZsRE~\citep{levy2017zero}, WikiData$_\text{counterfact}$~\citep{cohen2024evaluating}, and ELKEN~\citep{peng2024event} datasets for editing, and mathematical reasoning datasets for general tasks.
Evaluation adopts autoregressive decoding with semantic consistency judged by Qwen2.5-72B-Instruct across four dimensions~\citep{zhang2024comprehensive}. 
% We test 1, 10, and 100 sequential edits; 
A token-level locality check confirms the semantic results (Appendix~\ref{sec:Token-Level Locality Evaluation}). Full implementation details, hardware specifications, and metric calculation procedures are provided in Appendix~\ref{sec:Full Experimental Settings}.

% \begin{figure*}[!t]
%   \vskip 0.2in
%   \begin{center}
%     \centerline{\includegraphics[width=0.95\textwidth]{fig_scale_llama3.pdf}}
%     \caption{
%       Performance changes of knowledge editing methods during sequential editing of Llama-3.1-8B-Instruct on the ZsRE dataset. The x-axis represents the number of edits: 1, 10, 100, and `All' for the full dataset. Note that MEMIT and PMET are excluded, as they fail to run on Llama-3.1-8B-Instruct.
%     }
%     \label{fig:scale_llama3}
%   \end{center}
%   \vspace{-26pt}
% \end{figure*}

% ===============================================================
\subsection{Experimental Results} %  on General LLMs
% RQ1 实用场景性能：参数编辑法在连续编辑中崩溃，SCR全面占优
% RQ2 推理能力影响：多数参数编辑法导致推理崩溃，SCR表现稳定
% RQ3 事件知识泛化：参数编辑法在复杂事件型任务中表现极差
% RQ4 时间效率对比：参数法编辑慢但推理快，非参数法编辑快但推理有延迟
% ===============================================================

\textbf{RQ1: How do knowledge editing methods perform in practice-oriented settings?} 

% \begin{figure*}[htb]
%   \vskip 0.2in
%   \begin{center}
%     \centerline{\includegraphics[width=0.95\textwidth]{fig_scale_llama3.pdf}}
%     \caption{
%       Performance changes of knowledge editing methods during sequential editing of Llama-3.1-8B-Instruct on the ZsRE dataset. The x-axis represents the number of edits: 1, 10, 100, and `All' for the full dataset. Note that MEMIT and PMET are excluded, as they fail to run on Llama-3.1-8B-Instruct.
%     }
%     \label{fig:scale_llama3}
%   \end{center}
%   \vspace{-26pt}
% \end{figure*}

% ZsRE数据集上连续编辑场景中各类知识编辑方法的对比
Figure~\ref{fig:scale_llama3} illustrates the intermediate performance of Llama-3.1-8B-Instruct on the ZsRE dataset as the number of edits increases from 1 to 10, 100, and finally to the full dataset. For the corresponding results of Llama-2-7B-Chat and Mistral-7B-Instruct, refer to Figure~\ref{fig:scale_llama2} and Figure~\ref{fig:scale_mistral} in the Appendix \ref{sec:AdditionResult}, respectively. 
% Tables~\ref{tab:sequential_performance_zsre} and~\ref{tab:sequential_performance_counterfact} in Appendix \ref{sec:AdditionResult} present a comprehensive comparison of knowledge editing methods in sequential editing scenarios on the ZsRE and WikiData\(_\text{counterfact}\) datasets, respectively. 
Appendix \ref{sec:AdditionResult} (Tables~\ref{tab:sequential_performance_zsre}–\ref{tab:add_results_largerLLM}) presents a comprehensive comparison of knowledge editing methods across ZsRE, WikiData\(_\text{counterfact}\), and larger LLMs such as Llama-2-13B and Qwen3-14B.
% 更大规模模型上的适用性
% We further extend our evaluation to larger LLM, Llama-2-13B and Qwen3-14B(Appendix~\ref{sec:AdditionResult}, Table~\ref{tab:add_results_largerLLM}). 

% As shown in Table~\ref{tab:add_results_largerLLM} in Appendix~\ref{sec:AdditionResult}, the results are highly consistent with the insights observed on 7B and 8B LLMs.

\begin{table*}[t]
\centering
\caption{Performance comparison of knowledge editing methods applied to DeepSeek-R1-Distill-Llama-8B on the ZsRE dataset, reporting results across four evaluation dimensions in both single and sequential editing settings, along with their average scores.}

% \tiny
\small

\resizebox{0.8\linewidth}{!}{  % 改为 0.9 而不是 1.0
\begin{tabular}{l|*{5}{r}|*{5}{r}}
\toprule
\multirow{2}{*}{\textbf{Method}} & \multicolumn{5}{c|}{\textbf{Single Editing}} & \multicolumn{5}{c}{\textbf{Sequential Editing}} \\
\cmidrule(lr){2-6} \cmidrule(lr){7-11}
& \makecell[c]{Rel.} & \makecell[c]{Gen.} & \makecell[c]{Loc.} & \makecell[c]{Port.} & \makecell[c]{Avg.} 
& \makecell[c]{Rel.} & \makecell[c]{Gen.} & \makecell[c]{Loc.} & \makecell[c]{Port.} & \makecell[c]{Avg.} \\
\midrule
\textit{Pre-edit} & 3.00 & 3.00 & \underline{15.50} & 4.36 & 6.47 & 3.00 & 3.00 & \underline{15.50} & 4.36 & 6.47 \\
\midrule
ROME     & 36.00 & \underline{42.00} & 3.00  & \underline{17.99} & 24.75 & 1.00  & 0.00  & 0.00  & 0.00  & 0.25 \\
RECT     & \underline{37.00} & 35.00 & 6.00  & 16.02 & 23.51 & 0.00  & 0.00  & 0.00  & 0.00  & 0.00 \\
AlphaEdit& 43.00 & 24.00 & 13.50 & 8.88  & 22.35 & \underline{46.00} & \underline{35.00} & 8.00  & \underline{7.62}  & \underline{24.16} \\
FT-L       & 2.00  & 2.00  & 2.00  & 3.93  & 2.48  & 0.00  & 0.00  & 0.00  & 0.00  & 0.00 \\
MEND     & 36.00 & \underline{42.00} & 10.50 & 15.47 & \underline{25.99} & 0.00  & 0.00  & 0.00  & 0.00  & 0.00 \\
AdaLoRA  & 18.00 & 15.00 & 0.50  & 8.03  & 10.38 & 0.00  & 0.00  & 0.00  & 0.00  & 0.00 \\
WISE     & 8.00  & 5.00  & 3.00  & 2.59  & 4.65  & 2.00  & 2.00  & 7.50  & 2.52  & 3.50 \\
GRACE    & 31.00 & 3.00  & \underline{15.50} & 4.03  & 13.38 & 38.00 & 3.00  & \underline{15.50} & 4.03  & 15.13 \\
\midrule
SCR      & \textbf{85.00} & \textbf{84.00} & \textbf{15.50} & \textbf{41.87} & \textbf{56.59} & \textbf{90.00} & \textbf{90.00} & \textbf{15.50} & \textbf{45.26} & \textbf{60.19} \\
\bottomrule
\end{tabular}
}
\label{tab:llama_distill_zsre_comparison}
\vspace{-5pt}
\end{table*}

The experimental results reveal the following: 
% (1)所有基于参数修改的方法都远未达到先前研究中报道的近乎完美的单次编辑性能。
(1) Under autoregressive inference and evaluation based on semantic consistency, all parameter modification-based methods fall significantly short of the near-perfect single-edit performance reported in prior work.
% (2)基于参数修改的方法（如ROME和MEND）在连续编辑场景中失效
(2) Most parameter modification-based methods, such as ROME and MEND, collapse under sequential editing scenarios: as the number of edits increases, all metrics quickly drop to near zero. This indicates a complete failure to retain knowledge across multiple updates.
% (3)AlphaEdit的短板
(3) AlphaEdit maintains a stable performance rate across continuous edits, its Reliability and Generalization scores on LLaMA-3.1-8B-Instruct are only 69.49 and 55.50, respectively. 
More importantly, it underperforms on Locality and Portability, with scores of just 8.38 and 8.58, highlighting its limited flexibility in applying edited knowledge.
It is worth noting that AlphaEdit demonstrates sensitivity to hyperparameters and model architecture. Further analysis can be found in Appendix~\ref{sec:AlphaEdit Hyperparameter Sensitivity Analysis}.
% AlphaEdit对超参数和模型架构高度敏感——即使在共享MEMIT设置的情况下，其在Llama-2-7B-Chat和Mistral-7B-Instruct模型上的连续编辑性能接近零值
% It is important to note that AlphaEdit demonstrates high sensitivity to hyperparameters and model architecture; its performance on Llama-2-7B-Chat and Mistral-7B-Instruct was near zero under sequential editing, even with shared settings from MEMIT. See Appendix~\ref{sec:AlphaEdit Hyperparameter Sensitivity Analysis} for further details.
% (4)所有参数编辑方法的性能均被SCR这一简单直观的基线方案超越。
% (4) All parameter-based editing methods are outperformed by SCR, a simple and intuitive retrieval-based reference baseline
% % SCR employs an extensible external textual memory without modifying model parameters, and achieves the best performance across all dimensions in our evaluation. 
% without modifying model parameters.
% This highlights the practical potential of external-knowledge-based approaches in real-world knowledge updating tasks.
% % SCR的卓越表现为评估未来参数编辑方法的实用价值设立了重要参照基准，表明未来方法需超越该水平才能证明其更广泛的适用性。
(4) All parameter-based editing methods are outperformed by SCR, a simple and intuitive retrieval-based reference baseline without modifying model parameters. This highlights the practical potential of external-knowledge-based approaches in real-world knowledge updating tasks. Rather than serving as a direct competitor to parameter-based editing methods, SCR provides a distinct empirical perspective by operating outside the parametric editing paradigm, facilitating a principled study of the efficiency–reliability trade-off.
% The strong performance of SCR provides a useful practical reference point for evaluating future parameter-based editing methods. 
% In particular, methods that directly modify model parameters should ideally demonstrate clear advantages over such a simple retrieval-based alternative in order to justify broader applicability.
Supplemental comparisons between SCR and parameter-based editing methods
% , including a Pareto analysis of edit-time versus inference-time trade-offs and experiments combining SCR retrieval with parameter edits,
are reported in Appendix~\ref{sec:Fairness of SCR vs. Parameter-Based Editing Methods} (Tables~\ref{tab:scr_vs_alphaedit} and~\ref{tab:scr_vs_parameter_based}).

A fine-grained analysis over a wider range of edit counts  (Table~\ref{tab:fine_grained_sequential_editing} in Appendix~\ref{sec:AdditionResult}) further confirms that while the timing of degradation varies by method, all parameter-based editing methods exhibit performance decline over longer edit sequences. Additionally, 
% a breakdown of edits by model confidence (Table~\ref{tab:stubborn_vs_simple_errors} in Appendix~\ref{sec:AdditionResult}) shows 
there is consistent trends across both high- and low-confidence fact sets (Appendix~\ref{sec:AdditionResult}, Table~\ref{tab:stubborn_vs_simple_errors}).

\noindent\textbf{RQ2: Can knowledge editing help reasoning LLMs integrate new facts without degrading reasoning ability?}

% 通过一系列数学推理基准测试检测编辑后模型的性能
% As LLMs are increasingly applied to complex tasks involving multi-step reasoning, mathematical problem solving, and logical consistency, their strong reasoning capabilities have emerged as a central strength. However, the effectiveness of knowledge editing methods in reasoning-oriented LLMs remains largely underexplored. 
As LLMs are increasingly applied to complex reasoning tasks, reasoning ability has become a key strength. Yet the effectiveness of knowledge editing methods for reasoning-oriented LLMs remains largely underexplored.

Tables \ref{tab:llama_distill_zsre_comparison} and \ref{tab:llama_disitll_wiki_counterfact_comparison} (Appendix \ref{sec:AdditionResult}) present results on ZsRE dataset and WikiData\(_\text{counterfact}\), respectively. 
To evaluate the retention of reasoning abilities after editing, we assess edited LLMs on a suite of mathematical reasoning benchmarks and broader downstream tasks (including GPQA-Diamond, ARC\(_\text{c}\), and MMLU-Pro), reporting accuracy after 10 and 100 edits (Tables~\ref{tab:reasoning_ability_post_editing} and ~\ref{tab:Broader-Downstream-Tasks} in Appendix~\ref{sec:AdditionResult}). 
% We evaluate retention of reasoning abilities using mathematical reasoning benchmarks and broader downstream tasks, such as GPQA-Diamond, ARC\(_\text{c}\), and MMLU-Pro. Accuracy results after 10 and 100 edits are provided in Appendix~\ref{sec:AdditionResult}.
The results consistently show that reasoning performance collapses after extensive sequential edits for most parameter-based editing methods, while AlphaEdit exhibits relative resilience.

% 依赖参数修改的知识编辑方法在面向推理的大语言模型上表现逊色于通用大语言模型。
Based on the comparison between Table \ref{tab:llama_distill_zsre_comparison} and Table \ref{tab:sequential_performance_zsre}, knowledge editing methods that rely on parameter modification exhibit inferior performance on reasoning-oriented LLMs compared to general-purpose LLMs. For instance, AlphaEdit's average performance in sequential editing scenarios drops from 35.48 to 24.16. This degradation may be due to the more implicit and distributed nature of knowledge representation in reasoning LLMs, as well as their greater dependence on long-range context reasoning. In contrast, SCR remains a strong practical reference baseline by integrating internal and external knowledge through selective context construction.

% 连续多次编辑会导致模型推理能力崩溃
As shown in Table~\ref{tab:reasoning_ability_post_editing}
% (Appendix~\ref{sec:AdditionResult}), after 10 editing steps
, ROME and RECT retain some performance, with average accuracy drops of 2.74 and 8.37, respectively. However, after 100 edits, their accuracy nearly drops to zero, indicating a complete collapse in reasoning ability. For MEND, accuracy remains at 0.00 after both 10 and 100 edits, suggesting that the method leads to total LLM failure. In contrast, AlphaEdit exhibits minimal performance degradation across all datasets, even after 100 sequential edits, and in some cases, exhibits slight improvements. Its parameter modifications can maintain reasoning accuracy close to that of the original LLM. Nevertheless, due to its limitations in locality and portability for knowledge editing, AlphaEdit may be better suited for general-purpose Parameter-Efficient Fine-Tuning rather than targeted knowledge editing tasks.

% 编辑推理型大语言模型存在容易出现幻觉的局限
% Through case analysis (see details in Appendix~\ref{case_study}), we observe that editing reasoning-oriented LLMs poses distinct challenges. 
A detailed case analysis (see Appendix~\ref{case_study}) reveals that editing reasoning-oriented LLMs presents unique challenges.
While editing methods can successfully guide the LLM to produce the correct next token, the edited LLM's internal reasoning process often leads it to reflect outdated knowledge, thereby undermining the effectiveness of the edit. Moreover, in striving to preserve logical coherence within its reasoning trajectory, the edited LLM may generate explanations that are plausible-sounding but entirely fabricated. In some instances, the edited LLM may even disregard the original question in its attempt to round out the answer. These behaviors underscore key limitations of current knowledge editing methods when applied to reasoning-oriented LLMs.

\textbf{RQ3: Can knowledge editing methods generalize from factual knowledge to event knowledge?}

% \begin{figure}[t]
%   \vskip 0.2in
%   \begin{center}
%     \centerline{\includegraphics[width=\columnwidth]{fig_elken_alphaedit_vs_scr_comparison.pdf}}
%     \caption{
%       Performance comparison between AlphaEdit and SCR on sequential editing tasks using the ELKEN dataset.
%     }
%     \label{fig:elken_sample}
%   \end{center}
%   \vspace{-26pt}
% \end{figure}

% 进行基于事实的知识编辑实验
While most existing research focuses on fact-based knowledge editing in the form of triples, real-world knowledge is often organized in a more complex manner, such as event. 

\begin{figure}[t]
  \vskip 0.2in
  \begin{center}
    \centerline{\includegraphics[width=\columnwidth]{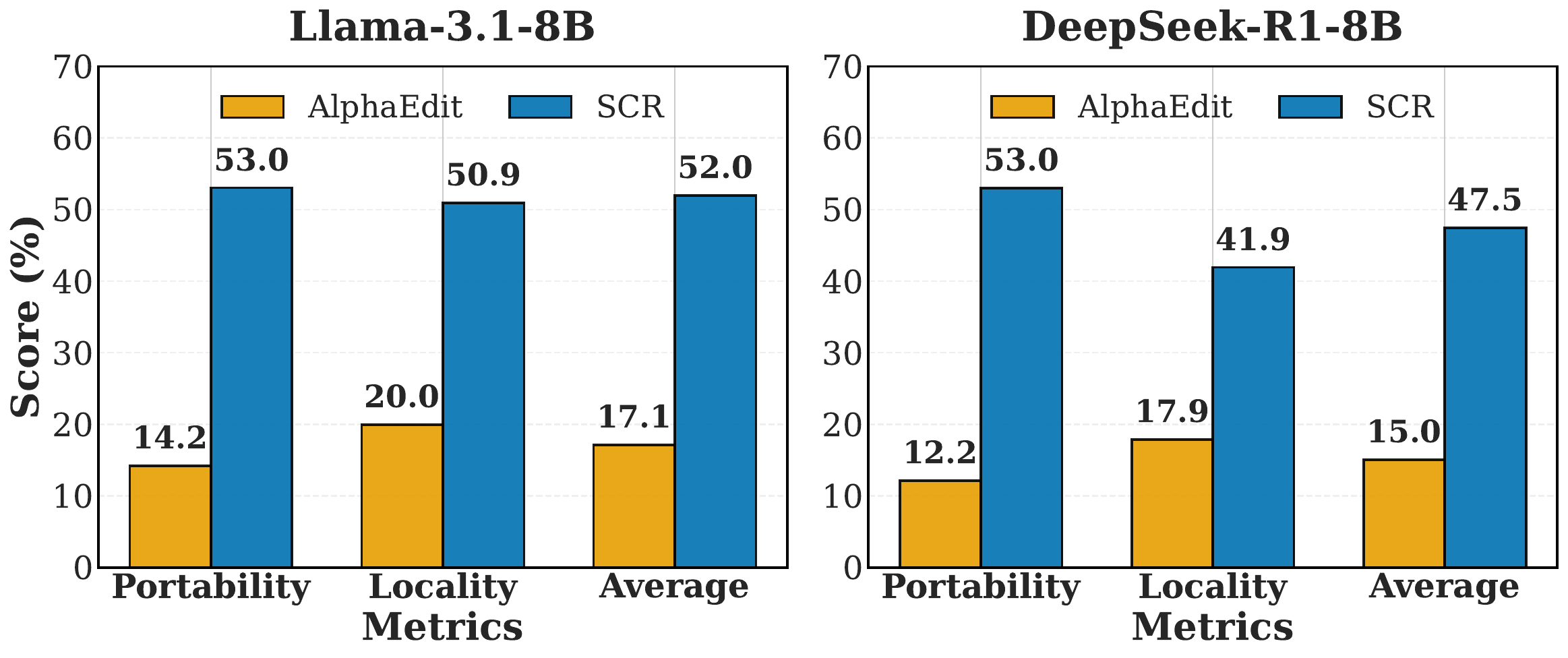}}
    \caption{
      Performance comparison between AlphaEdit and SCR on sequential editing tasks using the ELKEN dataset.
    }
    \label{fig:elken_sample}
  \end{center}
  \vspace{-26pt}
\end{figure}

However, current studies on knowledge editing rarely evaluate methods on more complex, event-level datasets. 
To address this limitation, we conduct knowledge editing experiments on Llama-2-7B-Chat, Llama-3.1-8B-Instruct, and DeepSeek-R1-Distill-Llama-8B using the ELKEN dataset~\citep{peng2024event}, 
the dataset processing code using GPT-4o is fully described in Appendix \ref{sec:ADataset}.
Performance comparison between AlphaEdit and SCR on sequential editing tasks are shown in Figure \ref{fig:elken_sample}.
Detailed results for sigle editing and sequential editing are presented in Table~\ref{tab:single_elken}  and Table~\ref{tab:sequential_elken} 
(Appendix \ref{sec:AdditionResult}). 

% 事实编辑中基于参数修改的知识编辑方法在单次和连续编辑场景中全面溃败，而SCR硕果仅存
The results indicate that most parameter-based knowledge editing methods struggle to achieve satisfactory performance even in single-edit scenarios, and they almost completely break down under sequential editing. For instance, although AlphaEdit performs well on fact-based datasets, its performance deteriorates significantly when applied to event-level editing tasks involving multiple entities and attributes occurring concurrently. This degradation is primarily due to the limited ability of these methods to capture complex semantic relationships between entities and to integrate contextual information across interconnected elements via parameter-level adjustments. In contrast, SCR consistently achieves the best results across nearly all settings we evaluate. These findings underscore the practical limitations of parameter updates for editing small amounts of knowledge, while highlighting the substantial potential of in-context learning in LLMs for real-world applications.

\noindent\textbf{RQ4: How do different knowledge editing methods compare in terms of time efficiency?}

% 知识编辑方法的延迟和效率评估
% While the correctness and robustness of knowledge edits are critical, latency and efficiency are equally important for real-world deployment. 
% An ideal knowledge editing method should introduce minimal overhead, both during the edit process and at inference time. 
Beyond correctness and robustness, latency and efficiency are equally crucial for practical deployment. Effective editing methods should minimize both editing overhead and inference latency.
To systematically evaluate these aspects, we consider two key metrics: (i) \textit{edit time}, the wall-clock time required to apply an individual edit, and (ii) \textit{inference time}, defined as the average wall-clock latency per input query normalized by the base LLM's latency, measured under greedy decoding with a fixed output length of 50 tokens post-edit.

% \begin{figure}[t]
%   \vskip 0.2in
%   \begin{center}
%     \centerline{\includegraphics[width=0.8\columnwidth]{fig_tradeoff_pareto.pdf}}
%     \caption{
%       Trade-off between editing time and inference latency relative to the base model across different editing paradigms.
%     }
%     \label{fig:tradeoff_pareto}
%   \end{center}
%   \vspace{-30pt}
% \end{figure}

\begin{figure}[t]
  \vskip 0.2in
  \begin{center}
    \centerline{\includegraphics[width=0.8\columnwidth]{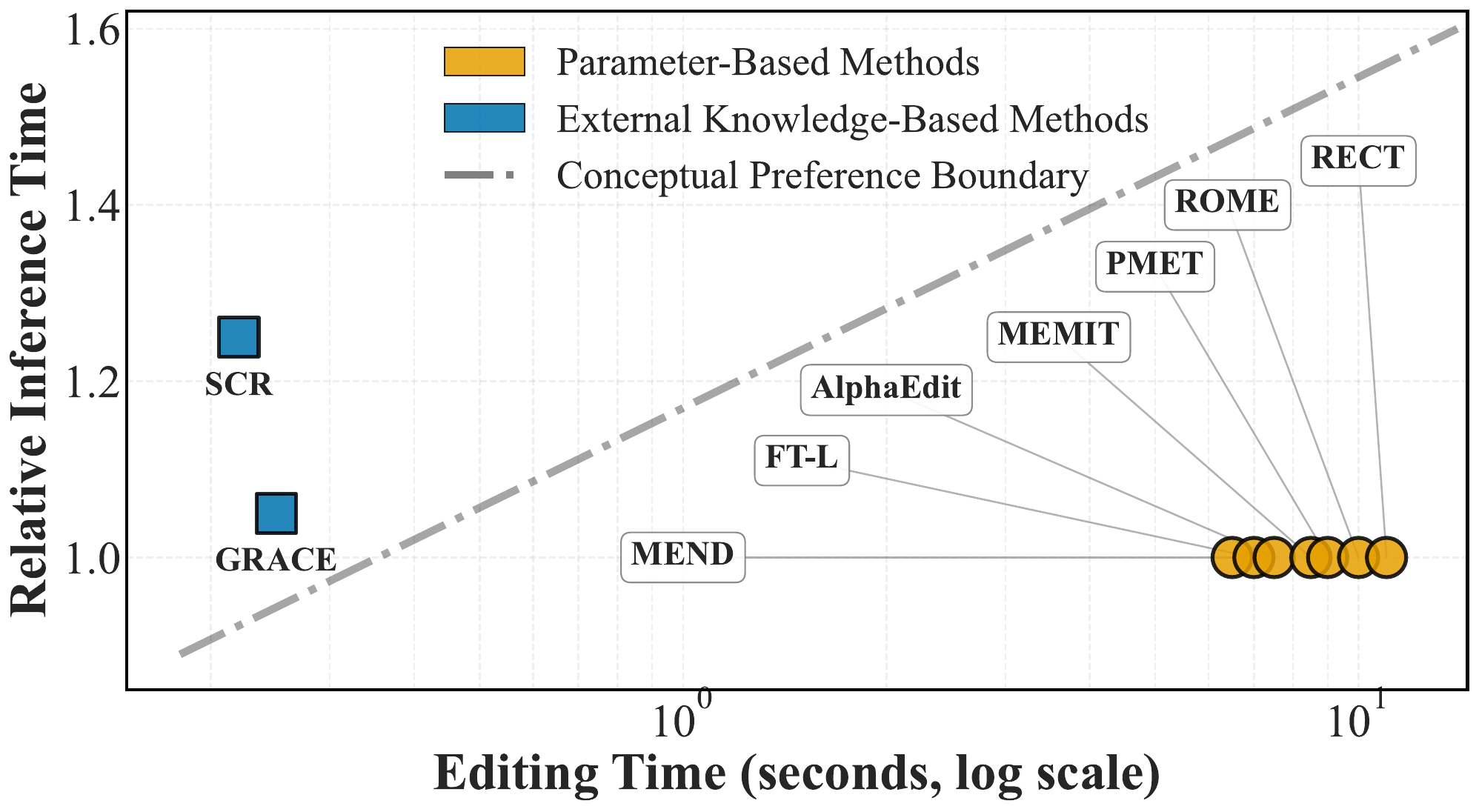}}
    \caption{
      Trade-off between editing time and inference latency relative to the base model across different editing paradigms.
    }
    \label{fig:tradeoff_pareto}
  \end{center}
  \vspace{-30pt}
\end{figure}

% 直接修改原始模型参数的方法不会引入额外的推理开销，SCR这样的基于检索的方法往往会产生额外的推理延迟。
Figure~\ref{fig:tradeoff_pareto} illustrates the trade-off between editing time and inference latency for parameter-based and external knowledge–based methods, with detailed results for all methods reported in Figure~\ref{fig:twocomparison} in the Appendix \ref{sec:AdditionResult}. Regarding editing time, methods that integrate knowledge through progressively more abstract representations, from textual edits to embeddings, hidden states, and ultimately parameter updates, exhibit increasing editing costs. This trend reflects the growing computational overhead associated with deeper and more indirect knowledge integration. In contrast, parameter-based editing methods do not introduce additional inference latency, as they directly modify the original model parameters. 
% External knowledge-based approaches, such as SCR, typically incur higher inference costs due to retrieval operations or longer input sequences. 
External knowledge-based methods, such as SCR, generally incur higher inference costs, primarily because of retrieval operations or the need to handle extended input sequences.
Overall, no method achieves strong performance on both axes simultaneously. This reveals a fundamental limitation of current model editing paradigms: intrusive parameter updates trade update efficiency for inference efficiency, while non-intrusive methods preserve fast updates at the expense of increased inference latency. 
A more detailed analysis of SCR's VRAM consumption, inference time, and performance stability across memory sizes and top-k settings is provided in Appendix~\ref{sec:Efficiency-Performance Analysis of SCR} (Tables~\ref{tab:vram_usage}--\ref{tab:scr_performance}).

% ==================================================
\section{Conclusion}
\label{sec:diss}
% 大多数基于参数修改的知识编辑方法表现显著低于预期，简单的SCR方法很棒。
% 更可靠的解决方案需要在上下文学习和模型训练之间寻找平衡
% ==================================================

This paper shows that the limitations of parameter-based knowledge editing are rooted in the geometry of LLM representation spaces. Building on dimensional collapse, we develop a theoretical framework that models editing as a first-order perturbation of hidden representations and proves that bounded weight updates can induce disproportionately large distortions along low-variance principal directions, with such effects compounding under sequential editing. This identifies a concrete mechanism by which localized parameter modifications lead to global failures in reasoning and generalization.
Extensive experiments under realistic settings corroborate these theoretical predictions: across varying knowledge complexity, numbers of edits, and evaluation dimensions, parameter-based editing methods consistently degrade core LLM capabilities, with performance deteriorating as edits accumulate, largely independent of the specific editing algorithm.
Taken together, our results reveal a fundamental stability--efficiency trade-off in knowledge updating: methods that enable efficient, localized modifications can compromise the global stability of model representations, while mechanisms that better preserve representational integrity are often less efficient. Understanding, formalizing, and managing this trade-off is therefore a central challenge for future knowledge update methods.

\section*{Acknowledgements}

% We sincerely thank ECNU LINK Lab for providing essential computational resources and a supportive environment for this study. We are also grateful to the anonymous reviewers for their insightful comments and suggestions, which helped enhance the quality of this paper. Additionally, we acknowledge the significant contributions of prior works in the field, as well as the open-source methods and datasets utilized in our research, which laid the foundation for this study.

This work was supported by the National Natural Science Foundation of China (Grant No. 72204087), the Chenguang Program of Shanghai Education Development Foundation and Shanghai Municipal Education Commission (Grant No. 23CGA28), the Shanghai Pujiang Program (Grant No. 23PJC030), and the Young Elite Scientists Sponsorship Program by CAST (Grant No. YESS20240562). We thank the ECNU LINK Lab for providing essential computational resources and a supportive research environment, which were instrumental in carrying out this study. We are also deeply grateful to the anonymous reviewers for their thoughtful comments and suggestions, which greatly improved the quality of this paper. Finally, we acknowledge the inspiring contributions of prior works in the field, as well as the open-source methods and datasets utilized in our study, which laid a critical foundation for this research.

\section*{Impact Statement}

This paper establishes a rigorous, practice-oriented evaluation framework for knowledge editing methods, 
% grounded in a geometric theory of why parameter-based edits can destabilize LLM representations. 
grounded in a geometric framework that explains why parameter-based knowledge editing methods may destabilize LLM representations.
By revealing that even carefully targeted weight modifications systematically degrade core model capabilities, especially under realistic sequential editing conditions, 
our work encourages the community to shift focus from isolated factual accuracy metrics to adopting holistic evaluation standards that account for reliability, generalization, locality, portability, and reasoning preservation. 
We believe this shift in evaluation practice can steer future research toward more robust and trustworthy knowledge updating paradigms.

At the same time, we acknowledge two limitations of our analysis. First, our theoretical framework relies on specific assumptions (dimensional collapse, small perturbations, and local representation stability) and should be regarded as a falsifiable hypothesis rather than a conclusive theory. Second, although our findings highlight consistent shortcomings of current parameter-based editing methods, they should not be considered a wholesale rejection of parameter-based editing methods, which may still prove effective under carefully constrained conditions or with improved regularization. We hope this work inspires principled progress rather than deterring further exploration.

Finally, as with any technique enabling targeted modification of LLM behavior, advanced knowledge editing tools present inherent dual-use risks. However, the evaluation standards and diagnostic tools proposed in this work can help detect and mitigate potential misuse.

% \bibliography{icml2026_conference}
\bibliographystyle{icml2026}

\newpage
\appendix
\onecolumn

% % ==============================================================
% \section{Use of LLMs}
% % ==============================================================
% In writing this paper, LLMs were used only for text polishing; no ideas or research content were generated by them. The authors bear full responsibility for all content and claims presented herein.

%================================
\section{Related Work}
\label{sec:Arelated}
%================================

% This paper is dedicated to revisiting knowledge updating in LLMs. We begin by reviewing existing knowledge editing methods, followed by a critical analysis of concerns associated with these methods. Lastly, we review the emergence and applications of contextual reasoning capabilities within LLMs.

Recent studies on updating factual knowledge in large language models have introduced diverse editing methods and evaluation protocols. In this section, we review representative knowledge editing approaches and summarize the experimental settings commonly used in prior work.

%=================================
\subsection{Knowledge Editing Methods}
\label{sec:Knowledge Editing Methods}
%=================================

LLMs are often regarded as knowledge bases, as they encapsulate vast amounts of world knowledge within their extensive parameters, which are derived from large-scale datasets during the pre-training phase~\citep{petroni2019language, geva2021transformer, geva2023dissecting, dai2022knowledge}. 
To cope with knowledge updates, the knowledge editing approach~\citep{meng2022locating, mengmass} encodes target knowledge into specific parameters, which are then replaced or supplemented in the LLM to update its factual knowledge.

\textbf{Locate-then-edit.} 
methods~\citep{mengmass, zhang2024knowledge, li2024pmet, li2024consecutive,hu2024wilke}
% yan-etal-2025-keys
are ground in the interpretability theory of Transformer architecture~\citep{geva2021transformer,lv2024interpreting}. It posits that knowledge is distributed across feed-forward networks (FFNs), while attention modules play a role in information copying and transmission. For instance, ROME~\citep{meng2022locating} utilizes causal tracing to first identify the crucial neurons associated with specific knowledge before performing targeted edits. 
Furthermore, RECT~\citep{gu2024model} and AlphaEdit~\citep{fang2024alphaedit} introduce additional constraints based on ROME to prevent excessive parameter shifts during the editing process.
% Recent extensions further improve this paradigm in practical settings: EAMET~\citep{dai2025eamet} enhances scalability in massive editing by aligning key and residual embedding spaces; CoRE~\citep{park2025context} improves robustness under preceding contexts by reducing context-sensitive hidden-state variance; and AnyEdit~\citep{jiang2025anyedit} extends editing beyond single-token facts through autoregressive chunk-wise updates for long-form knowledge.
Note that, the assumption of the localized storage of factual knowledge remains controversial~\citep{wei2024does}. 
An alternative hypothesis suggests that the relationship between neurons and knowledge is characterized by a many-to-many dynamic rather than a simplistic one-to-one association~\citep{allen2024physics}. Any modification to the parameters will inevitably affect other knowledge stored in the LLM, including both the original knowledge, and previously edited knowledge.

\textbf{Meta-learning.} methods, such as MEND~\citep{de2021editing} and MALMEN~\citep{tanmassive}, employ hyper-networks that are designed to forecast tailored weight updates for each knowledge data instance associated with an LLM. However, the hyper-network for a particular LLM limits their scalability in sequential editing scenarios. Furthermore, the additional training process incurs significant time and computational costs. Additionally, the necessity of modifying parameters for a limited amount of knowledge encapsulated in textual form is a matter of ongoing debate.

\textbf{Additional parameter-based.} methods aim to efficiently integrate target knowledge by isolating the parameters that require adjustment, such as WISE~\citep{wang2024wise}, T-Patcher~\citep{huangtransformer}, MELO~\citep{yu2024melo} and many others~\citep{dong2022calibrating, zhang2024dafnet, wang2024memoe, wang2024lemoe, wang2024roselora}. 
These methods typically introduce additional parameters or utilize mixture of experts (MoE) architectures~\citep{chen2022towards}, either at the head of the LLM or within its structure.
However, as the number of additional parameters increases, the likelihood of overfitting escalates. This phenomenon can result in the post-edit LLM neglecting prior edits, and compromising its original knowledge. Besides, the continual expansion of neurons may further exacerbate the post-edit LLM's inference burden.

\textbf{External Memory-based.} methods~\citep{zhong2023mquake,hartvigsen2024aging,mitchell2022memory,jiang2024learning, chen2024lifelong, daslarimar, zheng2023can, markowitz2025k} maintain a memory store for updated knowledge, which can be represented as plain text, hidden states, token embeddings, or knowledge graphs. SERAC~\citep{mitchell2022memory}, a classical method, simulates the editing scope by training a discriminator, whose results distinguish between the original LLM and the counterfactual model. GRACE~\citep{hartvigsen2024aging} maintains a dynamically updated codebook that alters the hidden states during the forward propagation.

In short, most existing methods update LLM knowledge by modifying their parameters or structures. However, in real-world scenarios, the amount of new knowledge available is limited, compared to the vast pre-training data used by LLMs. Encoding new knowledge into the model’s parameters can lead to the loss of original knowledge, risking both an incomplete understanding of updates and potential conflicts with prior knowledge.

\subsection{Summary of Experimental Settings}
\label{sec:sum_exp_setting}
% ===============================================================

A comprehensive summary of the experimental settings adopted by recent methods is provided in Table \ref{tab:model_setting}.

\begin{table*}[htbp]
  \centering
  \caption{Summary of experimental settings for recent methods. We specify whether \textbf{Fact}-, \textbf{Event}-based, or \textbf{Gene}ral purpose \textbf{Datasets} are used. Target \textbf{LLM} specifies whether an \textbf{Instruct} LLM or an \textbf{Reason}ing LLM is employed. \textbf{Infer}ence settings show whether \textbf{Auto}regressive setting is applied instead of teacher-forcing. Under \textbf{Edits}, we clarify if \textbf{Single} or \textbf{Seq}uential editing is tested. For evaluation \textbf{Dimensions}, we specify if \textbf{Rel}iability, \textbf{Gen}eralization, \textbf{Loc}ality, and \textbf{Port}ability are considered.} % \futing{the notation of different dimensions are the same, for example, there are three `Gen`. it's hard for readers.}
  \scriptsize
   \resizebox{\linewidth}{!}{  
    \begin{tabular}{@{}ll|ccc|cc|c|cc|cccc@{}}

    \toprule
    \multicolumn{2}{c|}{} & \multicolumn{3}{c|}{\textbf{Datasets}} & \multicolumn{2}{c|}{\textbf{LLM}} & \textbf{Infer} & \multicolumn{2}{c|}{\textbf{Edits}} & \multicolumn{4}{c}{\textbf{Dimensions}} \\
    \textbf{Category} & \textbf{Method} & \textbf{Fact} & \textbf{Event} & \textbf{Gene.} & \textbf{Instruct} & \textbf{Reason} & \textbf{Auto} & \textbf{Single} & \textbf{Seq.} & \textbf{Rel.} & \textbf{Gen.} & \textbf{Loc.} & \textbf{Port.} \\
    \midrule
    \multirow{6}[2]{*}{Locate-then-edit} & ROME \citep{meng2022locating}  & \cmark & \textcolor{red}{\xmark} & \textcolor{red}{\xmark} & \cmark & \textcolor{red}{\xmark} & \textcolor{red}{\xmark} & \cmark & \textcolor{red}{\xmark} & \cmark & \cmark & \cmark & \textcolor{red}{\xmark} \\
          & MEMIT \citep{mengmass} & \cmark & \textcolor{red}{\xmark} & \textcolor{red}{\xmark} & \cmark & \textcolor{red}{\xmark} & \textcolor{red}{\xmark} & \cmark & \cmark & \cmark & \cmark & \cmark & \textcolor{red}{\xmark} \\
          & PMET \citep{li2024pmet}  & \cmark & \textcolor{red}{\xmark} & \textcolor{red}{\xmark} & \cmark & \textcolor{red}{\xmark} & \textcolor{red}{\xmark} & \cmark & \cmark & \cmark & \cmark & \cmark & \textcolor{red}{\xmark} \\
          & RECT \citep{gu2024model} & \cmark & \textcolor{red}{\xmark} & \textcolor{red}{\xmark} & \cmark & \textcolor{red}{\xmark} & \textcolor{red}{\xmark} & \cmark & \cmark & \cmark & \cmark & \cmark & \textcolor{red}{\xmark} \\
          & AlphaEdit \citep{fang2024alphaedit} & \cmark & \textcolor{red}{\xmark} & \cmark & \cmark & \textcolor{red}{\xmark} & \textcolor{red}{\xmark} & \cmark & \cmark & \cmark & \cmark & \cmark & \cmark \\
          & FT-L \citep{meng2022locating}  & \cmark & \textcolor{red}{\xmark} & \textcolor{red}{\xmark} & \cmark & \textcolor{red}{\xmark} & \textcolor{red}{\xmark} & \cmark & \textcolor{red}{\xmark} & \cmark & \cmark & \cmark & \textcolor{red}{\xmark} \\
          % & SAKE \citep{scialanga2025sake}  & \cmark & \textcolor{red}{\xmark} & \textcolor{red}{\xmark} & \cmark & \textcolor{red}{\xmark} & \cmark & \cmark & \cmark & \cmark & \cmark & \cmark & \cmark \\
          & CoRE \citep{park2025context}  & \cmark & \textcolor{red}{\xmark} & \cmark & \cmark & \textcolor{red}{\xmark} & \cmark & \cmark & \textcolor{red}{\xmark} & \cmark & \cmark & \cmark & \textcolor{red}{\xmark} \\
          & EAMET \citep{dai2025eamet}  & \cmark & \textcolor{red}{\xmark} & \cmark & \cmark & \textcolor{red}{\xmark} & \textcolor{red}{\xmark} & \cmark & \cmark & \cmark & \cmark & \cmark & \cmark \\
          & AnyEdit \citep{jiang2025anyedit}  & \cmark & \cmark & \textcolor{red}{\xmark} & \cmark & \textcolor{red}{\xmark} & \cmark & \cmark & \textcolor{red}{\xmark} & \cmark & \cmark & \cmark & \textcolor{red}{\xmark} \\
    \midrule
    Meta-learning & MEND \citep{de2021editing}  & \cmark & \textcolor{red}{\xmark} & \textcolor{red}{\xmark} & \cmark & \textcolor{red}{\xmark} & \textcolor{red}{\xmark} & \cmark & \textcolor{red}{\xmark} & \cmark & \cmark & \cmark & \textcolor{red}{\xmark} \\
    \midrule
    \multirow{3}[2]{*}{Additional-Parameter} & AdaLoRA \citep{zhang2023adalora} & \cmark & \textcolor{red}{\xmark} & \textcolor{red}{\xmark} & \cmark & \textcolor{red}{\xmark} & \textcolor{red}{\xmark} & \cmark & \textcolor{red}{\xmark} & \cmark & \cmark & \cmark & \textcolor{red}{\xmark} \\
          & WISE \citep{wang2024wise} & \cmark & \textcolor{red}{\xmark} & \textcolor{red}{\xmark} & \cmark & \textcolor{red}{\xmark} & \textcolor{red}{\xmark} & \cmark & \cmark & \cmark & \cmark & \cmark & \textcolor{red}{\xmark} \\
          & AdaEdit \citep{li2025adaedit}  & \cmark & \textcolor{red}{\xmark} & \cmark & \cmark & \textcolor{red}{\xmark} & \textcolor{red}{\xmark} & \cmark & \cmark & \cmark & \cmark & \cmark & \textcolor{red}{\xmark} \\
    \midrule
    External Memory & GRACE \citep{hartvigsen2024aging} & \cmark & \textcolor{red}{\xmark} & \textcolor{red}{\xmark} & \cmark & \textcolor{red}{\xmark} & \textcolor{red}{\xmark} & \cmark & \cmark & \cmark & \cmark & \cmark & \textcolor{red}{\xmark} \\
    \bottomrule
    \end{tabular}%
    }
  \label{tab:model_setting}%
\end{table*}%

% ===============================================================
\section{Justifications for Working Hypotheses and Empirical Support}
\label{sec:Detailed Justifications}
% ===============================================================

\subsection{Hypothesis~\ref{hypo:1}: Dimensional Collapse}
\label{sec:Hypothesis 1}
This subsection evaluates whether the hidden representations in our studied models exhibit the spectral characteristics required by Hypothesis~\ref{hypo:1}.

\textbf{Literature Motivation}
% \paragraph{Literature motivation}

% Prior work on representation geometry has documented anisotropy, rank collapse, and low-dimensional concentration in Transformer hidden states. These findings motivate modeling the hidden representation matrix as having a compressed singular-value spectrum.
Prior research on the geometry of representations in neural networks has consistently documented phenomena such as anisotropy, rank collapse, and low-dimensional concentration in Transformer hidden states. These findings suggest that hidden representation matrices can be effectively modeled as having a compressed singular value spectrum. 
Recent empirical studies \cite{NEURIPS2023_a0e66093} specifically indicate that hidden representations in intermediate layers of large Transformer models experience a significant contraction in intrinsic dimensionality. This contraction implies that the representations lie near a low-dimensional manifold with a small effective rank and a rapidly decaying singular value spectrum, supporting the assumption of a near-zero minimum singular value ($\sigma_{\text{min}}$) in our hypothesis.

% Empirical studies on the geometry of large transformer representations indicate that hidden representations in intermediate layers exhibit a significant contraction in intrinsic dimensionality, suggesting a concentration near a low-dimensional manifold \cite{NEURIPS2023_a0e66093}. This geometric property implies a representation space with a low effective rank and a rapidly decaying singular value spectrum, which aligns with the assumption of an extremely small minimum singular value $\sigma_{\min}$ in our hypothesis of dimensional collapse. 

% The observation that editing triggers a pervasive ``ripple effect in the hidden space" \cite{wang2024missing} further indicates a structural fragility consistent with such a collapsed geometry. Furthermore, mechanistic analyses trace specific editing failures to low-variance substructures within attention modules \cite{xie2025revealingdeceptiveness}, implying that critical information is encoded in directions highly susceptible to perturbation.

Further supporting evidence comes from \citet{pinter2023emptying}, who observe that editing operations in large language models (LLMs) cause a pervasive "ripple effect" in the hidden representation space. This suggests structural fragility, which aligns with the geometric characteristics of dimensional collapse. Additionally, mechanistic analyses \cite{xie2025revealingdeceptiveness} link specific editing failures to low-variance substructures in attention modules, highlighting the vulnerability of critical information encoded in directions with low variance and sensitivity to perturbation.

% \subsection{Hidden State Extraction}

\textbf{Hidden State Extraction} 

For each evaluated model (Llama-3.1-8B-Instruct, DeepSeek-R1-Distill-Llama-8B, Llama-2-13B), we sample $n=100000$ input prompts from the C4-en dataset. 
% We perform a forward pass before and after each editing operation, collecting hidden states $h\in\mathbb{R}^d$ after theC down-projection layer of each Transformer block's feed-forward network using PyTorch forward hooks. For the T5-3B encoder-decoder model, hidden states are collected at the FFN input and output projections of the specific layers targeted by MEND (encoder layers 22–23, decoder layers 22–23), following the default hyperparameter settings for MEND.
Hidden states $h \in \mathbb{R}^d$ are then extracted using forward hooks implemented in PyTorch during forward passes, capturing layer-specific hidden representations before and after each editing operation. Specifically, hidden states are collected from the down-projection layer of each Transformer block's feed-forward network. For the T5-3B encoder-decoder model, hidden representations are captured at the input and output projections of feed-forward networks in the layers targeted by MEND (i.e., encoder layers 22–23, decoder layers 22–23), following the default hyperparameter settings for MEND.

% \textbf{Output.} Per-layer activations are stored as \texttt{activations.pt} files containing the original and edited hidden-state tensors of shape $(B, T, d)$. All states are stored in FP32 precision.

% \subsection{Singular Value Spectrum Analysis}

\textbf{Singular Value Spectrum Analysis} 

% For each layer's hidden representation matrix $H \in \mathbb{R}^{d \times n}$ ($d$: feature dimension, $n=1024$: number of samples), we perform a GPU-accelerated full singular value decomposition $H = U\Sigma V^T$ using PyTorch's \texttt{torch.linalg.svd}. From the resulting singular value spectrum $\sigma_1 \geq \cdots \geq \sigma_d$, we compute:
For each layer, we construct the hidden representation matrix $H \in \mathbb{R}^{d \times n}$, where $d$ represents the feature dimension and $n = 100000$ is the number of samples.
A full singular value decomposition (SVD) $H = U \Sigma V^\top$ is performed using PyTorch's GPU-accelerated torch.linalg.svd implementation. This yields the ordered singular value spectrum $\{\sigma_1 \geq \sigma_2 \geq \dots \geq \sigma_d\}$, which provides insights into the representation space's intrinsic dimensionality and capacity.

\textbf{Empirical Validation}

We empirically analyze the singular value spectra of hidden representations across models and layers:

\begin{itemize}
    \item Table~\ref{tab:dimensional_collapse_appendix_llama31} reports dimensional collapse statistics for Llama-3.1-8B-Instruct.
    \item Table~\ref{tab:dimensional_collapse_deepseek} reports the same analysis for DeepSeek-R1-Distill-LLaMA-8B.
    \item Table~\ref{tab:dimensional_collapse_llama2} reports the same analysis for LlaMA-2-13B.
    \item Table~\ref{tab:dimensional_collapse_t5_mend} reports the same analysis for selected encoder and decoder projections in T5-3B.
\end{itemize}

Across all models, architectures, and configurations, we consistently observe the following trends: (1) low effective rank, (2) vanishingly small singular values in the spectral tail, and (3) large condition numbers. Although the degree of collapse varies by layer and architecture, these results strongly align with the hypothesis of dimensional collapse. Consequently, they provide empirical support for interpreting dimensional collapse as the spectral precondition underlying the relative amplification mechanisms explored in Section~\ref{sec:Perturbation-Analysis}.

% \begin{table*}[htbp]
%     \centering
%     \caption{Empirical evidence of dimensional collapse in the hidden representations of Llama-3.1-8B-Instruct. 
%     % The hidden representation space exhibits clear characteristics of dimensional collapse: the effective rank $r$ is significantly smaller than the feature dimension, the smallest singular value $\sigma_{\min}$ is extremely small ($\sim 10^{-5}$), and the condition number is very large ($>10^6$). These empirical patterns support the general validity of the dimensional collapse assumption.
%     }
%     \label{tab:dimensional_collapse_appendix_llama31}
%     \begin{tabular}{c|cccrcc}
%         \toprule
%         layer & $d$ & $r_{\text{eff}}$ & $r/d$ (\%) & $\sigma_{1}$ & $\sigma_{\min}$ & $\text{cond}(\sigma_{1}/\sigma_{\min})$ \\
%         \midrule
%         5  & 4096 & 3249 & 79.3 & 194.1      & $3.38\times10^{-6}$ & $5.74\times10^{7}$ \\
%         10 & 4096 & 3099 & 75.7 & 302.6      & $7.55\times10^{-6}$ & $4.01\times10^{7}$ \\
%         15 & 4096 & 3177 & 77.6 & 504.9      & $1.20\times10^{-4}$ & $4.19\times10^{6}$ \\
%         20 & 4096 & 3258 & 79.6 & 537.5      & $7.16\times10^{-6}$ & $7.50\times10^{7}$ \\
%         25 & 4096 & 3260 & 79.6 & 618.7      & $9.48\times10^{-6}$ & $6.53\times10^{7}$ \\
%         30 & 4096 & 2066 & 50.4 & 4922.3     & $3.85\times10^{-5}$ & $1.28\times10^{8}$ \\
%         31 & 4096 & 1023 & 25.0 & 13003.3    & $3.05\times10^{-3}$ & $4.26\times10^{6}$ \\
%         \bottomrule
%     \end{tabular}
% \end{table*}

\begin{table*}[htbp]
    \centering
    \caption{Empirical evidence of dimensional collapse in the hidden representations of Llama-3.1-8B-Instruct.}
    \label{tab:dimensional_collapse_appendix_llama31}
    \begin{tabular}{c|cccrcc}
    \toprule
    Layer & $d$ & $r_{\text{eff}}$ & $r/d$ (\%) & $\sigma_{1}$ & $\sigma_{\min}$ & $\text{cond}(\sigma_{1}/\sigma_{\min})$ \\
    \midrule
    5  & 4096 & 3249 & 79.3 & $1.9\times10^{2}$ & $3.4\times10^{-6}$ & $5.7\times10^{7}$ \\
    10 & 4096 & 3099 & 75.7 & $3.0\times10^{2}$ & $7.5\times10^{-6}$ & $4.0\times10^{7}$ \\
    15 & 4096 & 3177 & 77.6 & $5.0\times10^{2}$ & $1.2\times10^{-4}$ & $4.2\times10^{6}$ \\
    20 & 4096 & 3258 & 79.6 & $5.4\times10^{2}$ & $7.2\times10^{-6}$ & $7.5\times10^{7}$ \\
    25 & 4096 & 3260 & 79.6 & $6.2\times10^{2}$ & $9.5\times10^{-6}$ & $6.5\times10^{7}$ \\
    30 & 4096 & 2066 & 50.4 & $4.9\times10^{3}$ & $3.9\times10^{-5}$ & $1.3\times10^{8}$ \\
    31 & 4096 & 1023 & 25.0 & $1.3\times10^{4}$ & $3.0\times10^{-3}$ & $4.3\times10^{6}$ \\
    \bottomrule
    \end{tabular}
\end{table*}

% \begin{table*}[htbp]
%     \centering
%     \caption{Empirical evidence of dimensional collapse in the hidden representations of DeepSeek-R1-Distill-Llama-8B. 
%     % The hidden representation space exhibits clear characteristics of dimensional collapse: the effective rank $r$ is significantly smaller than the feature dimension, the smallest singular value $\sigma_{\min}$ is extremely small ($\sim 10^{-5}$), and the condition number is very large ($>10^7$).
%     }
%     \label{tab:dimensional_collapse_deepseek}
%     \begin{tabular}{c|cccrcc}
%         \toprule
%         Layer & $d$ & $r_{\text{eff}}$ & $r/d$ (\%) & $\sigma_{1}$ & $\sigma_{\min}$ & $\text{cond}(\sigma_{1}/\sigma_{\min})$ \\
%         \midrule
%         5  & 4096 & 3269 & 79.8 & 216.2 & $3.23\times10^{-6}$ & $6.70\times10^{7}$ \\
%         10 & 4096 & 3087 & 75.4 & 363.3 & $4.84\times10^{-6}$ & $7.51\times10^{7}$ \\
%         15 & 4096 & 3168 & 77.4 & 594.2 & $6.97\times10^{-6}$ & $8.52\times10^{7}$ \\
%         20 & 4096 & 3371 & 82.3 & 432.3 & $7.10\times10^{-6}$ & $6.09\times10^{7}$ \\
%         25 & 4096 & 3403 & 83.1 & 558.1 & $8.48\times10^{-6}$ & $6.58\times10^{7}$ \\
%         30 & 4096 & 2450 & 59.8 & 3101.8 & $3.07\times10^{-5}$ & $1.01\times10^{8}$ \\
%         31 & 4096 & 1395 & 34.1 & 7089.9 & $6.50\times10^{-5}$ & $1.09\times10^{8}$ \\
%         \bottomrule
%     \end{tabular}
% \end{table*}

\begin{table*}[htbp]
    \centering
    \caption{Empirical evidence of dimensional collapse in the hidden representations of DeepSeek-R1-Distill-Llama-8B.}
    \label{tab:dimensional_collapse_deepseek}
    \begin{tabular}{c|cccrcc}
        \toprule
        Layer & $d$ & $r_{\text{eff}}$ & $r/d$ (\%) & $\sigma_{1}$ & $\sigma_{\min}$ & $\text{cond}(\sigma_{1}/\sigma_{\min})$ \\
        \midrule
        5  & 4096 & 3269 & 79.8 & $2.2\times10^{2}$ & $3.2\times10^{-6}$ & $6.7\times10^{7}$ \\
        10 & 4096 & 3087 & 75.4 & $3.6\times10^{2}$ & $4.8\times10^{-6}$ & $7.5\times10^{7}$ \\
        15 & 4096 & 3168 & 77.4 & $5.9\times10^{2}$ & $7.0\times10^{-6}$ & $8.5\times10^{7}$ \\
        20 & 4096 & 3371 & 82.3 & $4.3\times10^{2}$ & $7.1\times10^{-6}$ & $6.1\times10^{7}$ \\
        25 & 4096 & 3403 & 83.1 & $5.6\times10^{2}$ & $8.5\times10^{-6}$ & $6.6\times10^{7}$ \\
        30 & 4096 & 2450 & 59.8 & $3.1\times10^{3}$ & $3.1\times10^{-5}$ & $1.0\times10^{8}$ \\
        31 & 4096 & 1395 & 34.1 & $7.1\times10^{3}$ & $6.5\times10^{-5}$ & $1.1\times10^{8}$ \\
        \bottomrule
    \end{tabular}
\end{table*}

% \begin{table*}[htbp]
%     \centering
%     \caption{Empirical evidence of dimensional collapse in the hidden representations of Llama-2-13b-hf. 
%     % The hidden representation space exhibits clear characteristics of dimensional collapse: the effective rank $r$ is significantly smaller than the feature dimension, the smallest singular value $\sigma_{\min}$ is extremely small ($\sim 10^{-5}$), and the condition number is very large ($>10^7$).
%     }
%     \label{tab:dimensional_collapse_llama2}
%     \begin{tabular}{c|cccrcc}
%         \toprule
%         Layer & $d$ & $r_{\text{eff}}$ & $r/d$ (\%) & $\sigma_{1}$ & $\sigma_{\min}$ & $\text{cond}(\sigma_{1}/\sigma_{\min})$ \\
%         \midrule
%         5  & 5120 & 3941 & 77.0 & 283.0 & $5.15\times10^{-6}$ & $5.50\times10^{7}$ \\
%         10 & 5120 & 3956 & 77.3 & 403.1 & $1.00\times10^{-5}$ & $4.02\times10^{7}$ \\
%         15 & 5120 & 4089 & 79.9 & 555.5 & $1.42\times10^{-5}$ & $3.91\times10^{7}$ \\
%         20 & 5120 & 4177 & 81.6 & 1022.7 & $1.92\times10^{-5}$ & $5.33\times10^{7}$ \\
%         25 & 5120 & 4249 & 83.0 & 955.5 & $2.33\times10^{-5}$ & $4.10\times10^{7}$ \\
%         30 & 5120 & 4224 & 82.5 & 1569.8 & $2.43\times10^{-5}$ & $6.46\times10^{7}$ \\
%         31 & 5120 & 4238 & 82.8 & 1376.0 & $2.29\times10^{-5}$ & $6.00\times10^{7}$ \\
%         35 & 5120 & 4087 & 79.8 & 2071.0 & $2.77\times10^{-5}$ & $7.47\times10^{7}$ \\
%         39 & 5120 & 1448 & 28.3 & 25763.7 & $1.77\times10^{-4}$ & $1.46\times10^{8}$ \\
%         \bottomrule
%     \end{tabular}
% \end{table*}

\begin{table*}[htbp]
    \centering
    \caption{Empirical evidence of dimensional collapse in the hidden representations of Llama-2-13B.}
    \label{tab:dimensional_collapse_llama2}
    \begin{tabular}{c|cccrcc}
        \toprule
        Layer & $d$ & $r_{\text{eff}}$ & $r/d$ (\%) & $\sigma_{1}$ & $\sigma_{\min}$ & $\text{cond}(\sigma_{1}/\sigma_{\min})$ \\
        \midrule
        5  & 5120 & 3941 & 77.0 & $2.8\times10^{2}$ & $5.2\times10^{-6}$ & $5.5\times10^{7}$ \\
        10 & 5120 & 3956 & 77.3 & $4.0\times10^{2}$ & $1.0\times10^{-5}$ & $4.0\times10^{7}$ \\
        15 & 5120 & 4089 & 79.9 & $5.6\times10^{2}$ & $1.4\times10^{-5}$ & $4.0\times10^{7}$ \\
        20 & 5120 & 4177 & 81.6 & $1.0\times10^{3}$ & $1.9\times10^{-5}$ & $5.3\times10^{7}$ \\
        25 & 5120 & 4249 & 83.0 & $9.6\times10^{2}$ & $2.3\times10^{-5}$ & $4.1\times10^{7}$ \\
        30 & 5120 & 4224 & 82.5 & $1.6\times10^{3}$ & $2.4\times10^{-5}$ & $6.5\times10^{7}$ \\
        31 & 5120 & 4238 & 82.8 & $1.4\times10^{3}$ & $2.3\times10^{-5}$ & $6.0\times10^{7}$ \\
        35 & 5120 & 4087 & 79.8 & $2.1\times10^{3}$ & $2.8\times10^{-5}$ & $7.5\times10^{7}$ \\
        39 & 5120 & 1448 & 28.3 & $2.6\times10^{4}$ & $1.8\times10^{-4}$ & $1.5\times10^{8}$ \\
        \bottomrule
    \end{tabular}
\end{table*}

\begin{table*}[htbp]
    \centering
    \caption{Empirical evidence of dimensional collapse in T5-3B hidden representations at MEND editing layers.}
    \label{tab:dimensional_collapse_t5_mend}
    \begin{tabular}{r|rrrccc}
        \toprule
        Layer & $d$ & $r_{\text{eff}}$ & $r/d$ (\%) & $\sigma_{1}$ & $\sigma_{\min}$ & $\text{cond}(\sigma_{1}/\sigma_{\min})$ \\
        \midrule
        Enc22-FFN-wi & 16384 & 839 & 5.1 & $1.9\times10^{4}$ & $3.3\times10^{-9}$ & $5.7\times10^{12}$ \\
        Enc22-FFN-wo & 1024 & 31 & 3.1 & $1.7\times10^{6}$ & $6.0\times10^{1}$ & $2.8\times10^{4}$ \\
        Enc23-FFN-wi & 16384 & 1194 & 7.3 & $1.5\times10^{4}$ & $4.0\times10^{-9}$ & $3.6\times10^{12}$ \\
        Enc23-FFN-wo & 1024 & 129 & 12.6 & $6.8\times10^{5}$ & $5.6\times10^{1}$ & $1.2\times10^{4}$ \\
        Dec22-FFN-wi & 16384 & 3334 & 20.3 & $6.5\times10^{3}$ & $2.6\times10^{-9}$ & $2.5\times10^{12}$ \\
        Dec22-FFN-wo & 1024 & 447 & 43.6 & $1.1\times10^{5}$ & $1.4\times10^{2}$ & $8.3\times10^{2}$ \\
        Dec23-FFN-wi & 16384 & 5997 & 36.6 & $1.5\times10^{5}$ & $2.9\times10^{-6}$ & $5.1\times10^{10}$ \\
        Dec23-FFN-wo & 1024 & 94 & 9.2 & $8.3\times10^{6}$ & $1.2\times10^{3}$ & $6.7\times10^{3}$ \\
        \bottomrule
    \end{tabular}
\end{table*}

% \textbf{Implication for the theory}
% These results support the use of dimensional collapse as the spectral precondition behind the relative amplification mechanism in Section~\ref{sec:Perturbation-Analysis}.
% ===============================================================
\subsection{Hypothesis~\ref{hypo:2}: Small-Perturbation Regime}
\label{sec:Hypothesis 2}
% ===============================================================
% This subsection evaluates whether single-edit perturbations are sufficiently small for the first-order approximation in Eq.~(\ref{eq:taylor_approx}) to be informative.

This subsection examines whether single-edit perturbations remain sufficiently small to render the first-order approximation in Eq.~(\ref{eq:taylor_approx}) an informative analytical framework.

\textbf{Literature Motivation}

Empirical research on the unintended side effects of editing models highlights a strong connection between the magnitude of weight updates and the degradation of model performance. Large weight updates, characterized by high Frobenius norms $\|\Delta W\|_F$, have been shown to induce overfitting and diminish generalization capabilities \cite{gu-etal-2024-model}. In contrast, editing techniques such as RECT, which explicitly constrain the magnitude of weight changes, have demonstrated effectiveness in mitigating these adverse effects \cite{gu-etal-2024-model}. This body of evidence underscores the importance of small, localized parameter perturbations for successful editing. Further validation comes from failure analyses, which link catastrophic model collapse directly to the generation of excessively large parameter updates \cite{yang2024fall}. 
Collectively, these findings justify the adoption of a first-order Taylor expansion as a locally linear approximation to analyze the effects of small edits.

% \subsection{Perturbation Norm Measurement}

\textbf{Perturbation Norm Measurement}

 % For each editing method (AlphaEdit, MEMIT, MEND, ROME, WISE), we apply a single edit to each model and compute the induced hidden-state perturbation $\Delta h = h_{\text{edit}} - h_{\text{orig}}$ at each edited layer. The L2 norm $\|\Delta h\|$ is recorded for each of $n=1024$ input samples. We report the mean, standard deviation, minimum, and maximum of these norms. This directly tests Assumption~\ref{hypo:2}---whether single-edit perturbations remain within a regime where a first-order Taylor approximation is informative.

 We evaluate five editing methods (AlphaEdit, MEMIT, MEND, ROME, WISE) by applying a single edit to each model and measuring the resulting hidden state changes across layers. Specifically, we compute the hidden-state perturbation $\Delta h = h_{\text{edit}} - h_{\text{orig}}$ for each edited layer and record its $\ell_2$ norm $\|\Delta h\|$ over $n = 100000$ input samples. To test Assumption~\ref{hypo:2}, which posits that single-edit perturbations remain in the small regime where a first-order Taylor approximation is appropriate, we report the mean, standard deviation, minimum, and maximum of these norms.

\textbf{Empirical Validation}

% We measure the norm of editing-induced hidden-state perturbations across several methods and models.

We quantify the norm of hidden-state perturbations induced by single edits across a variety of methods and models:

\begin{itemize}
    \item Table~\ref{tab:perturbation_norms} reports single-edit perturbation norms in Llama-3.1-8B-Instruct.
    \item Table~\ref{tab:perturbation_deepseek} reports the same analysis for DeepSeek-R1-Distill-LLaMA-8B.
    \item Table~\ref{tab:perturbation_llama2} reports the same analysis for LlaMA-2-13B.
    \item Table~\ref{tab:perturbation_t5_mend} reports the same analysis for T5-3B under MEND.
\end{itemize}

In decoder-only models, perturbation norms are generally small in most edited layers, supporting the validity of local first-order analyses for single edits. However, the results for T5 reveal a greater degree of layerwise heterogeneity, suggesting that the small-perturbation regime is influenced by both architectural and layer-specific factors. These findings support the use of Eq.~(\ref{eq:taylor_approx}) as a local analytical approximation for understanding single-edit effects, while also clarifying its limitations in more heterogeneous settings.

\begin{table*}[htbp]
    \centering
    \caption{Perturbation norms induced by a single edit in selected layers of Llama-3.1-8B-Instruct.}
    \label{tab:perturbation_norms}
    \begin{tabular}{c|crrrr}
        \toprule
        Method & Layer & mean($\|\Delta h\|$) & std & min & max \\
        \midrule
        AlphaEdit & 5  & 1.24 & 0.12 & 0.00 & 2.18 \\
                  & 10 & 1.79 & 0.14 & 0.01 & 3.92 \\
                  & 15 & 2.64 & 0.28 & 0.01 & 5.12 \\
                  & 20 & 2.71 & 0.33 & 0.01 & 6.01 \\
                  & 25 & 3.27 & 0.48 & 0.01 & 8.05 \\
                  & 30 & 9.00 & 1.66 & 0.02 & 23.10 \\
                  & 31 & 0.05 & 0.03 & 0.01 & 1.17 \\
        \midrule
        MEMIT & 5  & 1.24 & 0.12 & 0.00 & 2.18 \\
              & 10 & 1.79 & 0.14 & 0.00 & 3.92 \\
              & 15 & 2.64 & 0.28 & 0.00 & 5.12 \\
              & 20 & 2.71 & 0.33 & 0.00 & 6.01 \\
              & 25 & 3.27 & 0.48 & 0.00 & 8.05 \\
              & 30 & 9.00 & 1.66 & 0.01 & 23.10 \\
              & 31 & 0.03 & 0.01 & 0.01 & 0.53 \\
        \midrule
        MEND & 29 & 0.46 & 0.16 & 0.08 & 2.11 \\
             & 30 & 0.36 & 0.16 & 0.07 & 3.06 \\
             & 31 & 0.75 & 0.31 & 0.19 & 5.89 \\
        \midrule
        ROME & 5  & 1.28 & 0.12 & 0.27 & 2.21 \\
             & 10 & 1.79 & 0.14 & 0.10 & 3.92 \\
             & 15 & 2.64 & 0.28 & 0.11 & 5.12 \\
             & 20 & 2.71 & 0.33 & 0.09 & 6.04 \\
             & 25 & 3.27 & 0.48 & 0.12 & 8.11 \\
             & 30 & 9.00 & 1.66 & 0.38 & 23.10 \\
             & 31 & 0.53 & 0.27 & 0.12 & 13.76 \\
        \midrule
        WISE & 29 & 0.13 & 1.71 & 0.01 & 45.38 \\
             & 30 & 0.07 & 0.77 & 0.01 & 19.99 \\
             & 31 & 0.20 & 2.28 & 0.02 & 70.90 \\
        \bottomrule
    \end{tabular}
\end{table*}

\begin{table*}[htbp]
    \centering
    \caption{Perturbation norms induced by a single edit in selected layers of DeepSeek-R1-Distill-Llama-8B.}
    \label{tab:perturbation_deepseek}
    \begin{tabular}{c|crrrr}
        \toprule
        Method & Layer & mean($\|\Delta h\|$) & std & min & max \\
        \midrule
        AlphaEdit & 5  & 1.15 & 0.11 & 0.00 & 2.06 \\
                  & 10 & 1.67 & 0.13 & 0.02 & 2.75 \\
                  & 15 & 2.44 & 0.24 & 0.01 & 4.23 \\
                  & 20 & 2.58 & 0.29 & 0.01 & 5.76 \\
                  & 25 & 3.22 & 0.43 & 0.02 & 6.88 \\
                  & 30 & 7.19 & 1.01 & 0.04 & 15.88 \\
                  & 31 & 0.08 & 0.03 & 0.02 & 0.80 \\
        \midrule
        MEMIT & 5  & 1.15 & 0.11 & 0.00 & 2.06 \\
              & 10 & 1.67 & 0.13 & 0.01 & 2.76 \\
              & 15 & 2.44 & 0.24 & 0.01 & 4.23 \\
              & 20 & 2.58 & 0.29 & 0.01 & 5.76 \\
              & 25 & 3.22 & 0.43 & 0.01 & 6.88 \\
              & 30 & 7.20 & 1.01 & 0.02 & 15.89 \\
              & 31 & 0.03 & 0.01 & 0.01 & 0.22 \\
        \midrule
        MEND & 29 & 0.46 & 0.20 & 0.03 & 1.91 \\
             & 30 & 7.19 & 1.00 & 0.26 & 15.81 \\
             & 31 & 0.70 & 0.27 & 0.11 & 2.60 \\
        \midrule
        ROME & 5  & 1.19 & 0.11 & 0.31 & 2.07 \\
             & 10 & 1.67 & 0.13 & 0.11 & 2.80 \\
             & 15 & 2.44 & 0.24 & 0.10 & 4.25 \\
             & 20 & 2.58 & 0.29 & 0.09 & 5.76 \\
             & 25 & 3.22 & 0.44 & 0.14 & 6.88 \\
             & 30 & 7.20 & 1.01 & 0.30 & 15.97 \\
             & 31 & 0.42 & 0.16 & 0.13 & 4.59 \\
        \midrule
        WISE & 29 & 15.28 & 1.79 & 0.00 & 24.00 \\
             & 30 & 8.11 & 1.01 & 3.69 & 15.94 \\
             & 31 & 11.32 & 1.31 & 0.01 & 18.73 \\
        \bottomrule
    \end{tabular}
\end{table*}

\begin{table*}[htbp]
    \centering
    \caption{Perturbation norms induced by a single edit in selected layers of Llama-2-13B.}
    \label{tab:perturbation_llama2}
    \begin{tabular}{c|crrrr}
        \toprule
        Method & Layer & mean($\|\Delta h\|$) & std & min & max \\
        \midrule
        AlphaEdit & 5  & 2.07 & 0.21 & 1.17 & 3.69 \\
                  & 10 & 4.19 & 0.37 & 2.78 & 7.54 \\
                  & 15 & 6.06 & 0.55 & 3.65 & 10.97 \\
                  & 20 & 8.04 & 0.96 & 4.69 & 20.12 \\
                  & 25 & 9.83 & 1.45 & 4.95 & 21.32 \\
                  & 30 & 9.36 & 1.32 & 5.13 & 25.31 \\
                  & 35 & 0.02 & 0.02 & 0.01 & 0.60 \\
                  & 39 & 0.07 & 0.05 & 0.02 & 2.69 \\
        \midrule
        MEMIT & 5  & 2.07 & 0.21 & 1.17 & 3.70 \\
              & 10 & 4.19 & 0.36 & 2.78 & 7.54 \\
              & 15 & 6.06 & 0.55 & 3.65 & 10.90 \\
              & 20 & 8.04 & 0.96 & 4.69 & 20.12 \\
              & 25 & 9.83 & 1.45 & 4.95 & 21.37 \\
              & 30 & 9.36 & 1.32 & 5.15 & 25.29 \\
              & 35 & 0.06 & 0.04 & 0.02 & 1.23 \\
              & 39 & 0.17 & 0.18 & 0.03 & 11.38 \\
        \midrule
        MEND & 37 & 0.25 & 0.02 & 0.17 & 0.54 \\
             & 38 & 0.38 & 0.04 & 0.20 & 0.76 \\
             & 39 & 1.43 & 0.17 & 0.58 & 6.15 \\
        \midrule
        ROME & 5  & 2.09 & 0.21 & 1.18 & 3.71 \\
             & 10 & 4.19 & 0.36 & 2.78 & 7.55 \\
             & 15 & 6.06 & 0.55 & 3.65 & 10.98 \\
             & 20 & 8.04 & 0.96 & 4.69 & 20.12 \\
             & 25 & 9.83 & 1.45 & 4.95 & 21.34 \\
             & 30 & 9.37 & 1.32 & 5.12 & 25.37 \\
             & 35 & 0.12 & 0.09 & 0.02 & 3.23 \\
             & 39 & 0.32 & 0.30 & 0.06 & 29.98 \\
        \midrule
        WISE & 27 & 0.05 & 0.01 & 0.03 & 0.66 \\
             & 30 & 9.37 & 1.32 & 5.13 & 25.33 \\
             & 35 & 0.07 & 0.02 & 0.04 & 0.89 \\
             & 39 & 0.91 & 0.26 & 0.18 & 5.43 \\
        \bottomrule
    \end{tabular}
\end{table*}

% \begin{table*}[htbp]
%     \centering
%     \caption{Perturbation norms induced by a single MEND edit in selected layers of T5-3B. 
%     % Summary statistics (mean, standard deviation, minimum, maximum) for perturbation norms across methods, including AlphaEdit, MEMIT, MEND, ROME, and WISE. Small perturbations in most layers validate the use of first-order approximations for local effect analysis.
%     }
%     \label{tab:perturbation_t5_mend}
%     \begin{tabular}{c|rrrr}
%         \toprule
%         Layer & mean($\|\Delta h\|$) & std & min & max \\
%         \midrule
%         Enc22-FFN-wi & 0.158 & 0.012 & 0.114 & 0.281 \\
%         Enc22-FFN-wo & 42.205 & 9.938 & 10.971 & 134.675 \\
%         Enc23-FFN-wi & 0.468 & 0.039 & 0.313 & 0.707 \\
%         Enc23-FFN-wo & 822.339 & 107.670 & 359.913 & $1.57\times10^{3}$ \\
%         Dec22-FFN-wi & 21.683 & 2.583 & 14.048 & 49.767 \\
%         Dec22-FFN-wo & 125.574 & 24.623 & 63.187 & 394.574 \\
%         Dec23-FFN-wi & 133.784 & 68.820 & 37.000 & 736.755 \\
%         Dec23-FFN-wo & $3.76\times10^{3}$ & $3.77\times10^{3}$ & 865.088 & $4.89\times10^{4}$ \\
%         \bottomrule
%     \end{tabular}
% \end{table*}

\begin{table*}[htbp]
    \centering
    \caption{Perturbation norms induced by a single MEND edit in selected layers of T5-3B.}
    \label{tab:perturbation_t5_mend}
    \begin{tabular}{c|rrrr}
        \toprule
        Layer & mean($\|\Delta h\|$) & std & min & max \\
        \midrule
        Enc22-FFN-wi & 0.16 & 0.01 & 0.11 & 0.28 \\
        Enc22-FFN-wo & 42.21 & 9.94 & 10.97 & 134.68 \\
        Enc23-FFN-wi & 0.47 & 0.04 & 0.31 & 0.71 \\
        Enc23-FFN-wo & 822.34 & 107.67 & 359.91 & $1.57\times10^{3}$ \\
        Dec22-FFN-wi & 21.68 & 2.58 & 14.05 & 49.77 \\
        Dec22-FFN-wo & 125.57 & 24.62 & 63.19 & 394.57 \\
        Dec23-FFN-wi & 133.78 & 68.82 & 37.00 & 736.76 \\
        Dec23-FFN-wo & $3.76\times10^{3}$ & $3.77\times10^{3}$ & 865.09 & $4.89\times10^{4}$ \\
        \bottomrule
    \end{tabular}
\end{table*}

% \textbf{Implication for the theory}
% These results justify viewing Eq.~(\ref{eq:taylor_approx}) as a local analytical approximation, while also clarifying its limits.

% ===============================================================
\subsection{Hypothesis~\ref{hypo:3}: Local Representation Stability}
\label{sec:Hypothesis 3}
% ===============================================================

This subsection investigates whether principal directions in the representation space remain approximately stable under sequential editing.

\textbf{Literature Motivation}

Recent studies have shown that model editing can substantially distort the global geometry of hidden representations, a phenomenon termed "representation shattering," which is strongly correlated with performance degradation \cite{pmlr-v267-nishi25a}. These findings suggest that maintaining the structural stability of the representation space is essential for preserving model functionality after editing.
Furthermore, widespread observations of unintended geometric disruptions, such as the hidden space ripple effect \cite{wang2025missing}, reinforce concerns about the challenges of achieving stable, localized updates. This aligns with broader skeptical perspectives on the feasibility of treating large language models (LLMs) as reliably editable knowledge repositories \cite{pinter2023emptying}. 
We therefore adopt the approximate stability of principal component directions {$u_k$} as a working hypothesis to describe a stable editing regime.

\textbf{Principal Component Stability Analysis} 

% Under AlphaEdit sequential editing ($T = 1, 2, \ldots, 1301$ edits), we extract the principal component matrices $U_{\text{edit}}^{(T)}$ from the hidden states after each $T$-th edit and compare them against the original $U_{\text{orig}}$ from the unedited model. For a range of subspace dimensions $k \in \{10, 50, 100, 200, 500, 1000, 2000\}$, we compute the minimum principal angle $\theta_{\min}$ between the subspaces spanned by the top-$k$ principal components of $U_{\text{orig}}$ and $U_{\text{edit}}^{(T)}$. 
% Following \citet{pmlr-v267-nishi25a}, we use QR-stabilized SVD of $M = Q_1^T Q_2$ to obtain the cosine of the minimum angle: $\cos\theta_{\min} = \min_i \; \sigma_i(M)$, where $\sigma_i(M)$ are the singular values of $M$. 
% A $\cos\theta_{\min}$ close to 1 indicates near-alignment (stable), while values approaching 0 indicate near-orthogonality (unstable).

In the context of sequential editing ($T = 1, 2, \ldots, 
% 1301
all
$ edits), we extract the principal component matrices $U_{\text{edit}}^{(T)}$ from the hidden states after each $T$-th edit and compare them to the original principal component matrix $U_{\text{orig}}$ of the unedited model. To assess the geometric consistency between the top-$k$ principal components of $U_{\text{edit}}^{(T)}$ and $U_{\text{orig}}$, we calculate the minimum principal angle $\theta_{\text{min}}$ between their respective subspaces. To further quantify alignment, we compute the cosine of the minimum angle, $\cos \theta_{\text{min}}$. A value of $\cos \theta_{\text{min}}$ close to 1 indicates strong alignment and stability, while values approaching 0 suggest near-orthogonality and instability. This analysis is performed across a range of subspace dimensions $k \in \{10, 50, 100, 200, 500, 1000, 2000\}$.

\textbf{Empirical Validation}

% We compute the cosine similarity between the original principal directions and those obtained after sequential edits:

% We evaluate the cosine similarity between the original principal components and those obtained after sequential edits. Specifically, we assess the stability of principal component directions under continuous editing for both decoder-only and encoder-decoder architectures. For decoder-only models, including Llama-3.1-8B-Instruct, DeepSeek-R1-Distill-LLaMA-8B, and LlaMA-2-13B, we conduct experiments using AlphaEdit and ROME. For the encoder-decoder architecture, we examine the T5-3B model, focusing on input-projection and output-projection spaces under MEND editing. The experimental results are summarized in the following tables:

We evaluate the cosine similarity between the original principal directions and those obtained after sequential edits to quantify the stability of principal component directions under sequential editing. We perform this analysis for both decoder-only and encoder-decoder models. For decoder-only models, including Llama-3.1-8B-Instruct, DeepSeek-R1-Distill-LLaMA-8B, and LlaMA-2-13B, we conduct experiments with AlphaEdit and ROME. For the encoder-decoder model T5-3B, we examine the FFN input-projection ($W_i$) and output-projection ($W_o$) spaces under edits performed with MEND. The results are summarized in the following tables:

\begin{itemize}
    \item Tables~\ref{tab:pc_stability}, \ref{tab:pc_stability_deepseek}, and \ref{tab:pc_stability_llama2} report the cosine similarity between the original and edited principal directions at layer 30 of Llama-3.1-8B-Instruct, DeepSeek-R1-Distill-LLaMA-8B, and LLaMA-2-13B, respectively, under sequential edits with AlphaEdit.
    \item Tables~\ref{tab:pc_stability_rome}, \ref{tab:pc_stability_deepseek_rome}, and \ref{tab:pc_stability_llama2_rome} report the corresponding results for the same models under sequential edits with ROME.
    \item Tables~\ref{tab:pc_stability_t5_input} and \ref{tab:pc_stability_t5_output} report the cosine similarity between the original and edited principal directions in the FFN input-projection and output-projection spaces of T5-3B under sequential edits with MEND.
\end{itemize}

Across models, dominant PCs often remain relatively stable during the initial edits, while lower-variance PCs deteriorate rapidly as edits accumulate. In T5-3B, certain layers exhibit a loss of stability almost immediately after editing begins. These observations reinforce the interpretation of Hypothesis~\ref{hypo:3} as a short-horizon idealization and underscore that the practical degradation of representation geometry may exceed the predictions of this simplified scenario.

\begin{table*}[htbp]
    \centering
    \caption{Stability of principal component directions at layer 30 of Llama-3.1-8B-Instruct under sequential editing using AlphaEdit. 
    % The cosine similarity is computed between the original principal components (PCs) and those after a specified number of edits. As the number of editing steps increases, the stability of the PC directions, particularly for components associated with low variance, progressively degrades.
    }
    \label{tab:pc_stability}
    \begin{tabular}{ccccc}
        \toprule
        Edit Step & Top10 PCs Cos & Top100 PCs Cos & Top200 PCs Cos & Top2000 PCs Cos \\
        \midrule
        1   & 0.999 & 0.981 & 0.927 & 0.003 \\
        2   & 1.000 & 1.000 & 1.000 & 0.940 \\
        3   & 1.000 & 1.000 & 1.000 & 0.936 \\
        4   & 0.999 & 0.981 & 0.927 & 0.002 \\
        5   & 1.000 & 1.000 & 0.999 & 0.872 \\
        6   & 0.999 & 0.981 & 0.927 & 0.003 \\
        7   & 1.000 & 1.000 & 0.998 & 0.917 \\
        8   & 0.999 & 0.981 & 0.926 & 0.003 \\
        9   & 1.000 & 0.999 & 0.997 & 0.866 \\
        10  & 0.999 & 0.981 & 0.926 & 0.004 \\
        20  & 0.999 & 0.981 & 0.925 & 0.006 \\
        50  & 0.997 & 0.993 & 0.956 & 0.299 \\
        100 & 0.996 & 0.980 & 0.910 & 0.002 \\
        200 & 0.994 & 0.966 & 0.802 & 0.002 \\
        500 & 0.990 & 0.813 & 0.379 & 0.003 \\
        1000& 0.973 & 0.552 & 0.128 & 0.002 \\
        % 1301& 0.962 & 0.447 & 0.067 & 0.020 \\
        all & 0.962 & 0.447 & 0.067 & 0.020 \\
        \bottomrule
    \end{tabular}
\end{table*}

\begin{table*}[htbp]
    \centering
    \caption{Stability of principal component directions at layer 30 of DeepSeek-R1-Distill-Llama-8B under sequential editing using AlphaEdit. 
    % The cosine similarity is computed between the original principal components (PCs) and those after a specified number of edits. As the number of editing steps increases, the stability of the PC directions, particularly for components associated with low variance, progressively degrades.
    }
    \label{tab:pc_stability_deepseek}
    \begin{tabular}{cccccc}
        \toprule
        Edit Step & Top10 PCs Cos & Top100 PCs Cos & Top200 PCs Cos & Top2000 PCs Cos \\
        \midrule
        1   & 1.000 & 0.984 & 0.906 & 0.014 \\
        2   & 0.999 & 0.984 & 0.902 & 0.015 \\
        3   & 0.999 & 0.984 & 0.902 & 0.014 \\
        4   & 0.999 & 0.984 & 0.904 & 0.015 \\
        5   & 0.999 & 0.984 & 0.897 & 0.014 \\
        6   & 0.999 & 0.984 & 0.898 & 0.014 \\
        7   & 1.000 & 0.978 & 0.996 & 0.878 \\
        8   & 0.999 & 0.975 & 0.901 & 0.015 \\
        9   & 0.999 & 0.961 & 0.908 & 0.016 \\
        10  & 0.999 & 0.920 & 0.901 & 0.016 \\
        20  & 0.999 & 0.840 & 0.871 & 0.011 \\
        50  & 0.997 & 0.910 & 0.751 & 0.005 \\
        100 & 0.994 & 0.446 & 0.207 & 0.105 \\
        200 & 0.989 & 0.086 & 0.207 & 0.053 \\
        500 & 0.979 & 0.169 & 0.283 & 0.033 \\
        1000& 0.958 & 0.199 & 0.021 & 0.018 \\
        all & 0.940 & 0.139 & 0.015 & 0.000 \\
        % 1301& 0.940 & 0.139 & 0.015 & 0.000 \\
        \bottomrule
    \end{tabular}
\end{table*}

\begin{table*}[htbp]
    \centering
    \caption{Stability of principal component directions at layer 30 of Llama-2-13B under sequential editing using AlphaEdit. 
    % The cosine similarity is computed between the original principal components (PCs) and those after a specified number of edits. As the number of editing steps increases, the stability of the PC directions, particularly for components associated with low variance, progressively degrades.
    }
    \label{tab:pc_stability_llama2}
    \begin{tabular}{cccccc}
        \toprule
        Edit Step & Top10 PCs Cos & Top100 PCs Cos & Top200 PCs Cos & Top2000 PCs Cos \\
        \midrule
        1   & 0.999 & 0.981 & 0.946 & 0.040 \\
        2   & 0.999 & 0.981 & 0.946 & 0.039 \\
        3   & 0.999 & 0.981 & 0.946 & 0.039 \\
        4   & 0.999 & 0.980 & 0.946 & 0.040 \\
        5   & 0.999 & 0.981 & 0.946 & 0.040 \\
        6   & 0.999 & 0.980 & 0.946 & 0.039 \\
        7   & 0.999 & 0.980 & 0.946 & 0.040 \\
        8   & 0.999 & 0.980 & 0.946 & 0.039 \\
        9   & 0.999 & 0.980 & 0.946 & 0.039 \\
        10  & 0.999 & 0.980 & 0.946 & 0.038 \\
        20  & 0.999 & 0.978 & 0.946 & 0.038 \\
        50  & 0.999 & 0.971 & 0.942 & 0.029 \\
        100 & 0.999 & 0.929 & 0.932 & 0.528 \\
        200 & 0.998 & 0.904 & 0.372 & 0.039 \\
        500 & 0.997 & 0.159 & 0.268 & 0.001 \\
        1000& 0.986 & 0.230 & 0.441 & 0.005 \\
        all & 0.972 & 0.101 & 0.313 & 0.013 \\
        % 1301& 0.972 & 0.101 & 0.313 & 0.013 \\
        \bottomrule
    \end{tabular}
\end{table*}

\begin{table*}[htbp]
    \centering
    \caption{Stability of principal component directions at layer 30 of Llama-3.1-8B-Instruct under sequential editing using ROME. 
    % The cosine similarity is computed between the original principal components (PCs) and those after a specified number of edits. As the number of editing steps increases, the stability of the PC directions, particularly for components associated with low variance, progressively degrades.
    }
    \label{tab:pc_stability_rome}
    \begin{tabular}{ccccc}
        \toprule
        Edit Step & Top10 PCs Cos & Top100 PCs Cos & Top200 PCs Cos & Top2000 PCs Cos \\
        \midrule
        1   & 0.999 & 0.981 & 0.926 & 0.002 \\
        2   & 0.997 & 0.995 & 0.974 & 0.517 \\
        3   & 0.995 & 0.989 & 0.908 & 0.441 \\
        4   & 0.983 & 0.974 & 0.825 & 0.001 \\
        5   & 0.979 & 0.977 & 0.764 & 0.116 \\
        6   & 0.979 & 0.970 & 0.619 & 0.002 \\
        7   & 0.980 & 0.861 & 0.585 & 0.001 \\
        8   & 0.964 & 0.928 & 0.388 & 0.001 \\
        9   & 0.971 & 0.881 & 0.413 & 0.039 \\
        10  & 0.971 & 0.893 & 0.412 & 0.001 \\
        20  & 0.274 & 0.004 & 0.003 & 0.001 \\
        50  & 0.029 & 0.004 & 0.002 & 0.000 \\
        100 & 0.005 & 0.000 & 0.000 & 0.000 \\
        200 & 0.001 & 0.001 & 0.000 & 0.000 \\
        500 & 0.004 & 0.001 & 0.000 & 0.000 \\
        1000& 0.012 & 0.000 & 0.001 & 0.000 \\
        all & 0.012 & 0.001 & 0.002 & 0.000 \\
        % 1301& 0.012 & 0.001 & 0.002 & 0.000 \\
        \bottomrule
    \end{tabular}
\end{table*}

\begin{table*}[htbp]
    \centering
    \caption{Stability of principal component directions at layer 30 of DeepSeek-R1-Distill-Llama-8B under sequential editing using ROME. 
    % The cosine similarity is computed between the original principal components (PCs) and those after a specified number of edits. As the number of editing steps increases, the stability of the PC directions, particularly for components associated with low variance, progressively degrades.
    }
    \label{tab:pc_stability_deepseek_rome}
    \begin{tabular}{cccccc}
        \toprule
        Edit Step & Top10 PCs Cos & Top100 PCs Cos & Top200 PCs Cos & Top2000 PCs Cos \\
        \midrule
        1   & 0.999 & 0.984 & 0.901 & 0.010 \\
        2   & 0.998 & 0.964 & 0.878 & 0.007 \\
        3   & 0.995 & 0.893 & 0.749 & 0.002 \\
        4   & 0.986 & 0.686 & 0.658 & 0.000 \\
        5   & 0.987 & 0.667 & 0.657 & 0.001 \\
        6   & 0.985 & 0.633 & 0.513 & 0.007 \\
        7   & 0.972 & 0.417 & 0.241 & 0.037 \\
        8   & 0.959 & 0.610 & 0.237 & 0.003 \\
        9   & 0.947 & 0.599 & 0.203 & 0.003 \\
        10  & 0.936 & 0.518 & 0.151 & 0.002 \\
        20  & 0.125 & 0.008 & 0.002 & 0.001 \\
        50  & 0.003 & 0.001 & 0.002 & 0.000 \\
        100 & 0.000 & 0.002 & 0.002 & 0.000 \\
        200 & 0.007 & 0.000 & 0.001 & 0.000 \\
        500 & 0.003 & 0.000 & 0.001 & 0.000 \\
        1000& 0.008 & 0.003 & 0.002 & 0.000 \\
        all & 0.005 & 0.000 & 0.002 & 0.000 \\
        % 1301& 0.005 & 0.000 & 0.002 & 0.000 \\
        \bottomrule
    \end{tabular}
\end{table*}

\begin{table*}[htbp]
    \centering
    \caption{Stability of principal component directions at layer 30 of Llama-2-13B under sequential editing using ROME. 
    % The cosine similarity is computed between the original principal components (PCs) and those after a specified number of edits. As the number of editing steps increases, the stability of the PC directions, particularly for components associated with low variance, progressively degrades.
    }
    \label{tab:pc_stability_llama2_rome}
    \begin{tabular}{cccccc}
        \toprule
        Edit Step & Top10 PCs Cos & Top100 PCs Cos & Top200 PCs Cos & Top2000 PCs Cos \\
        \midrule
        1   & 1.000 & 0.991 & 0.996 & 0.963 \\
        2   & 0.999 & 0.981 & 0.982 & 0.870 \\
        3   & 1.000 & 0.990 & 0.983 & 0.858 \\
        4   & 1.000 & 0.994 & 0.986 & 0.199 \\
        5   & 0.999 & 0.987 & 0.976 & 0.632 \\
        6   & 0.999 & 0.982 & 0.973 & 0.510 \\
        7   & 0.999 & 0.979 & 0.965 & 0.526 \\
        8   & 0.008 & 0.008 & 0.002 & 0.001 \\
        9   & 0.004 & 0.007 & 0.002 & 0.001 \\
        10  & 0.992 & 0.942 & 0.793 & 0.331 \\
        20  & 0.621 & 0.138 & 0.207 & 0.044 \\
        50  & 0.294 & 0.008 & 0.027 & 0.002 \\
        100 & 0.007 & 0.000 & 0.000 & 0.000 \\
        200 & 0.001 & 0.002 & 0.002 & 0.001 \\
        500 & 0.005 & 0.002 & 0.002 & 0.000 \\
        1000& 0.012 & 0.001 & 0.002 & 0.000 \\
        all & 0.003 & 0.001 & 0.001 & 0.001 \\
        % 1301& 0.003 & 0.001 & 0.001 & 0.001 \\
        \bottomrule
    \end{tabular}
\end{table*}

\begin{table*}[htbp]
    \centering
    \caption{Stability of principal component directions in input projections of MEND editing layers in T5-3B under sequential editing. 
    % The cosine similarity is computed between the original principal components (PCs) and those after a specified number of edits. As the number of editing steps increases, the stability of the PC directions, particularly for components associated with low variance, progressively degrades.
    }
    \label{tab:pc_stability_t5_input}
    \resizebox{\linewidth}{!}{%
    \begin{tabular}{ccccccc}
        \toprule
        Layer & Edit Step & Top10 PCs Cos & Top50 PCs Cos & Top100 PCs Cos & Top1000 PCs Cos & Top5000 PCs Cos \\
        \midrule
        Enc22-FFN-wi & 1    & 0.999 & 0.999 & 0.999 & 0.999 & 0.993 \\
                     & 10   & 0.998 & 0.989 & 0.981 & 0.841 & 0.031 \\
                     & 100  & 0.989 & 0.923 & 0.893 & 0.235 & 0.007 \\
                     & 1000 & 0.116 & 0.182 & 0.035 & 0.005 & 0.000 \\
        \midrule
        Enc23-FFN-wi & 1    & 0.999 & 0.999 & 0.999 & 0.971 & 0.962 \\
                     & 10   & 0.993 & 0.972 & 0.954 & 0.124 & 0.035 \\
                     & 100  & 0.308 & 0.033 & 0.016 & 0.003 & 0.000 \\
                     & 1000 & 0.003 & 0.002 & 0.000 & 0.000 & 0.000 \\
        \midrule
        Dec22-FFN-wi & 1    & 0.889 & 0.106 & 0.014 & 0.000 & 0.000 \\
                     & 10   & 0.219 & 0.000 & 0.003 & 0.000 & 0.000 \\
                     & 100  & 0.002 & 0.000 & 0.000 & 0.000 & 0.000 \\
                     & 1000 & 0.000 & 0.000 & 0.000 & 0.000 & 0.000 \\
        \midrule
        Dec23-FFN-wi & 1    & 0.767 & 0.082 & 0.036 & 0.002 & 0.000 \\
                     & 10   & 0.012 & 0.001 & 0.001 & 0.000 & 0.000 \\
                     & 100  & 0.001 & 0.000 & 0.000 & 0.000 & 0.000 \\
                     & 1000 & 0.000 & 0.000 & 0.000 & 0.000 & 0.000 \\
        \bottomrule
    \end{tabular}%
    }
\end{table*}

\begin{table*}[htbp]
    \centering
    \caption{Stability of principal component directions in output projections of MEND editing layers in T5-3B under sequential editing. 
    % The cosine similarity is computed between the original principal components (PCs) and those after a specified number of edits. As the number of editing steps increases, the stability of the PC directions, particularly for components associated with low variance, progressively degrades.
    }
    \label{tab:pc_stability_t5_output}
    \resizebox{\linewidth}{!}{%
    \begin{tabular}{ccccccc}
        \toprule
        Layer & Edit Step & Top5 PCs Cos & Top10 PCs Cos & Top50 PCs Cos & Top200 PCs Cos & Top500 PCs Cos \\
        \midrule
        Enc22-FFN-wo & 1    & 0.999 & 0.999 & 0.982 & 0.002 & 0.008 \\
                     & 10   & 0.978 & 0.952 & 0.766 & 0.006 & 0.001 \\
                     & 100  & 0.072 & 0.038 & 0.097 & 0.004 & 0.006 \\
                     & 1000 & 0.031 & 0.010 & 0.019 & 0.004 & 0.003 \\
        \midrule
        Enc23-FFN-wo & 1    & 0.947 & 0.816 & 0.401 & 0.097 & 0.007 \\
                     & 10   & 0.128 & 0.121 & 0.094 & 0.137 & 0.039 \\
                     & 100  & 0.019 & 0.029 & 0.003 & 0.012 & 0.002 \\
                     & 1000 & 0.009 & 0.002 & 0.001 & 0.000 & 0.001 \\
        \midrule
        Dec22-FFN-wo & 1    & 0.968 & 0.747 & 0.257 & 0.035 & 0.005 \\
                     & 10   & 0.082 & 0.129 & 0.039 & 0.001 & 0.001 \\
                     & 100  & 0.021 & 0.008 & 0.002 & 0.000 & 0.001 \\
                     & 1000 & 0.006 & 0.001 & 0.003 & 0.002 & 0.003 \\
        \midrule
        Dec23-FFN-wo & 1    & 0.830 & 0.732 & 0.459 & 0.009 & 0.018 \\
                     & 10   & 0.069 & 0.164 & 0.021 & 0.003 & 0.001 \\
                     & 100  & 0.009 & 0.007 & 0.002 & 0.000 & 0.003 \\
                     & 1000 & 0.014 & 0.004 & 0.002 & 0.001 & 0.003 \\
        \bottomrule
    \end{tabular}%
    }
\end{table*}

\subsection{Generalization Across Model Scales and Architectures}
\label{sec:Generalization Across Model Scales and Architectures}

The main text focuses on 7B--14B decoder-only LLMs, but the geometric signatures underlying our analysis extend beyond a single model. The dimensional-collapse statistics (Tables~\ref{tab:dimensional_collapse_appendix_llama31}--\ref{tab:dimensional_collapse_t5_mend}), perturbation norms (Tables~\ref{tab:perturbation_norms}--\ref{tab:perturbation_t5_mend}), PC-stability analyses (Tables~\ref{tab:pc_stability}--\ref{tab:pc_stability_t5_output}), and \(R_k\) distributions (Tables~\ref{tab:rk_distribution_llama3}--\ref{tab:rk_distribution_t5}) consistently reveal evidence of the same vulnerability mechanism across DeepSeek-R1-Distill-LLaMA-8B, LlaMA-2-13B, and specific components of T5-3B.

However, the encoder-decoder architecture exhibits greater heterogeneity, with certain layers departing from the small-perturbation regime more rapidly. As a result, we do not assert identical quantitative patterns across all architectures. Instead, we posit that the amplification mechanism arising from dimensional collapse is broadly applicable. The severity of this mechanism depends on factors including architecture type, layer characteristics, editing method, and editing horizon.

\subsection{Direct Empirical Validation of the Amplification Mechanism}
\label{sec:Direct Empirical Validation of the Amplification Mechanism}

This subsection evaluates the core prediction from Section~\ref{sec:Perturbation-Analysis}: perturbation amplification is strongest in low-singular-value directions and can be significant even after a single edit.

\subsubsection{Directional amplification along collapsed directions}

\textbf{Relative Change Rate $R_k$ Computation} 

For each editing method and each edited layer, we calculate the directional relative change rate $R_k$ as:
\[
R_k = \frac{\sqrt{n}\,|c_k|}{\sigma_k}, \quad c_k = u_k^T \Delta h,
\]
where $u_k$ and $\sigma_k$ are derived from the SVD of the original hidden representation matrix $H_{\text{orig}}$, and $\Delta h = h_{\text{edit}} - h_{\text{orig}}$ represents the per-sample perturbation. For each layer, we compute $R_k$ across all $d$ principal directions, summarizing the mean, median, maximum, standard deviation, and the proportion of directions where $R_k > 1$.

\textbf{Visualization and Key Insights} 

\begin{figure}[htbp]
  \centering
  % \vspace{0.2in}
  \includegraphics[width=0.9\columnwidth]{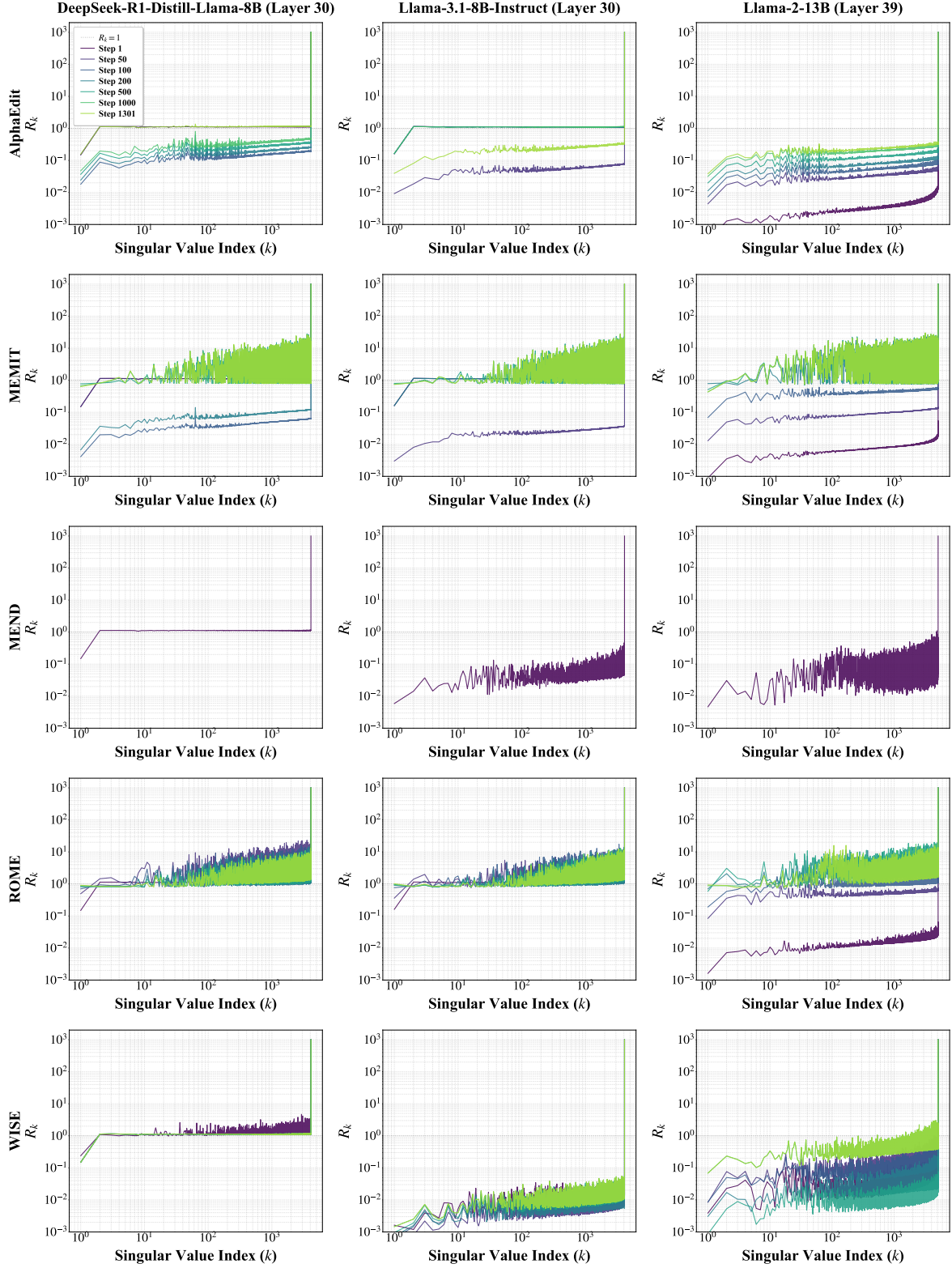}
  \caption{%
    Directional relative change rate \( R_k \) across editing methods, model architectures, and steps. 
    % Visualization of \( R_k = \frac{\sqrt{n} c_k}{\sigma_k} \) relative to the singular index \( k \), computed for various editing methods (AlphaEdit, MEMIT, MEND, ROME, WISE) across three architectures. The x-axis represents singular value indices \( k \), and the y-axis (log scale) reflects \( R_k \), where the dashed line (\( R_k = 1 \)) distinguishes mild perturbations (\( R_k < 1 \)) from severe distortions (\( R_k > 1 \)). Sequential editing accumulates perturbations, with low-singular-value directions (right) being disproportionately amplified, validating the mechanism predicted by Theorem~\ref{theo:Perturbation Amplification Theorem}.
  }
  \label{fig:R_k_5method_3model}
\end{figure}

In Figure~\ref{fig:R_k_5method_3model}, we visualize the empirical distributions of the directional relative change rate $R_k$ across different editing methods, language model architectures, and editing steps. The x-axis represents the singular value index $k$, while the y-axis (logarithmic scale) denotes $R_k$, reflecting the relative perturbation amplification. A critical threshold at $R_k = 1$ (dashed line) separates directions with tolerable perturbations ($R_k < 1$) from those exhibiting severe distortion ($R_k > 1$).

Several key patterns emerge from this analysis:

\begin{itemize}
    \item 
    \textbf{Cumulative Perturbation Amplification:} Across all methods and models, $R_k$ consistently increases with higher editing steps, confirming the cumulative nature of perturbation amplification.
    \item 
    \textbf{Sensitivity in Low-Singular-Value Directions:} Directions corresponding to smaller singular values are disproportionately affected. Even after a single edit, many of these directions exhibit $R_k \gg 1$, in alignment with the theoretical predictions in Theorem~\ref{theo:Perturbation Amplification Theorem}.
    \item 
    \textbf{Method-Specific Differences:} Editing methods vary in their impact on $R_k$. For example, AlphaEdit demonstrates greater stability over sequential edits, whereas methods like MEMIT and ROME exhibit sharp amplification, particularly during later editing steps.
    \item 
    \textbf{Model-Specific Sensitivity:} Architecture differences also influence sensitivity to perturbation. For instance, models like DeepSeek-R1-Distill-Llama-8B exhibit extreme $R_k$ values in low-singular-value directions after multiple editing iterations, suggesting heightened vulnerability to editing-induced distortions in these collapsed dimensions.
\end{itemize}

These findings provide direct empirical validation of the theoretical amplification mechanism outlined in Section~\ref{sec:Perturbation-Analysis}, particularly in low-singular-value dimensions. 

\subsubsection{Layerwise distributions of relative change rates}

We also report the layerwise distributions of $R_k$ for single edits across various language models and editing methods:

\begin{itemize}
    \item Table~\ref{tab:rk_distribution_llama3} reports detailed statistics of $R_k$ distributions for Llama-3.1-8B-Instruct
    \item Table~\ref{tab:rk_distribution_deepseek} reports the same analysis for DeepSeek-R1-Distill-LLaMA-8B
    \item Table~\ref{tab:rk_distribution_llama2} reports the same analysis for LlaMA-2-13B
    \item Table~\ref{tab:rk_distribution_t5} reports the same analysis for T5-3B
\end{itemize}

These tables show that even a single edit often yields \(R_k>1\) in a large fraction of directions and can produce extremely large mean and maximum values.

\begin{table}[t]
\centering
\caption{Layer-wise distribution of relative change rates ($R_k$) for a single edit on Llama-3.1-8B-Instruct.}
\label{tab:rk_distribution_llama3}
\small
\begin{tabular}{l c r r r r c c}
\toprule
Method & Layer & $R_k$ mean & $R_k$ min & $R_k$ max & Std & num($R_k>1$) & \%($R_k>1$) \\
\midrule
AlphaEdit & 5  & $3.39 \times 10^{2}$ & $7.43 \times 10^{-2}$ & $1.38 \times 10^{6}$ & $2.16 \times 10^{4}$ & 4095 & 99.98 \\
          & 10 & $2.23 \times 10^{2}$ & $1.57 \times 10^{-1}$ & $9.10 \times 10^{5}$ & $1.42 \times 10^{4}$ & 4095 & 99.98 \\
          & 15 & $2.09 \times 10^{1}$ & $1.97 \times 10^{-1}$ & $8.11 \times 10^{4}$ & $1.27 \times 10^{3}$ & 4095 & 99.98 \\
          & 20 & $3.43 \times 10^{2}$ & $2.08 \times 10^{-1}$ & $1.40 \times 10^{6}$ & $2.19 \times 10^{4}$ & 4095 & 99.98 \\
          & 25 & $3.42 \times 10^{2}$ & $2.17 \times 10^{-1}$ & $1.40 \times 10^{6}$ & $2.18 \times 10^{4}$ & 4093 & 99.93 \\
          & 30 & $2.17 \times 10^{2}$ & $1.61 \times 10^{-1}$ & $8.86 \times 10^{5}$ & $1.38 \times 10^{4}$ & 4095 & 99.98 \\
\midrule
MEMIT & 5  & $3.39 \times 10^{2}$ & $7.43 \times 10^{-2}$ & $1.38 \times 10^{6}$ & $2.16 \times 10^{4}$ & 4095 & 99.98 \\
      & 10 & $3.54 \times 10^{2}$ & $1.57 \times 10^{-1}$ & $1.45 \times 10^{6}$ & $2.26 \times 10^{4}$ & 4095 & 99.98 \\
      & 15 & $3.41 \times 10^{2}$ & $1.97 \times 10^{-1}$ & $1.39 \times 10^{6}$ & $2.17 \times 10^{4}$ & 4095 & 99.98 \\
      & 20 & $3.31 \times 10^{2}$ & $2.08 \times 10^{-1}$ & $1.35 \times 10^{6}$ & $2.11 \times 10^{4}$ & 4095 & 99.98 \\
      & 25 & $3.61 \times 10^{2}$ & $2.17 \times 10^{-1}$ & $1.48 \times 10^{6}$ & $2.31 \times 10^{4}$ & 4093 & 99.93 \\
      & 30 & $2.01 \times 10^{2}$ & $1.61 \times 10^{-1}$ & $8.20 \times 10^{5}$ & $1.28 \times 10^{4}$ & 4095 & 99.98 \\
\midrule
MEND & 29 & $6.09 \times 10^{2}$ & $7.20 \times 10^{-3}$ & $2.49 \times 10^{6}$ & $3.89 \times 10^{4}$ & 14  & 0.34 \\
     & 30 & $1.78 \times 10^{2}$ & $5.90 \times 10^{-3}$ & $7.28 \times 10^{5}$ & $1.14 \times 10^{4}$ & 1   & 0.02 \\
     & 31 & $2.70 \times 10^{1}$ & $5.60 \times 10^{-3}$ & $1.10 \times 10^{5}$ & $1.72 \times 10^{2}$ & 1   & 0.02 \\
\midrule
ROME & 5  & $3.39 \times 10^{2}$ & $7.54 \times 10^{-2}$ & $1.38 \times 10^{6}$ & $2.16 \times 10^{4}$ & 4095 & 99.98 \\
      & 10 & $3.55 \times 10^{2}$ & $1.57 \times 10^{-1}$ & $1.45 \times 10^{6}$ & $2.26 \times 10^{4}$ & 4095 & 99.98 \\
      & 15 & $3.41 \times 10^{2}$ & $1.98 \times 10^{-1}$ & $1.39 \times 10^{6}$ & $2.18 \times 10^{4}$ & 4095 & 99.98 \\
      & 20 & $3.31 \times 10^{2}$ & $2.08 \times 10^{-1}$ & $1.35 \times 10^{6}$ & $2.11 \times 10^{4}$ & 4095 & 99.98 \\
      & 25 & $3.61 \times 10^{2}$ & $2.17 \times 10^{-1}$ & $1.48 \times 10^{6}$ & $2.30 \times 10^{4}$ & 4093 & 99.93 \\
      & 30 & $2.01 \times 10^{2}$ & $1.61 \times 10^{-1}$ & $8.19 \times 10^{5}$ & $1.28 \times 10^{4}$ & 4095 & 99.98 \\
\midrule
WISE & 29 & $6.58 \times 10^{1}$ & $1.10 \times 10^{-3}$ & $2.69 \times 10^{5}$ & $4.21 \times 10^{3}$ & 1 & 0.02 \\
     & 30 & $1.25 \times 10^{0}$ & $1.20 \times 10^{-3}$ & $5.07 \times 10^{3}$ & $7.91 \times 10^{1}$ & 1 & 0.02 \\
     & 31 & $3.98 \times 10^{0}$ & $1.10 \times 10^{-3}$ & $1.63 \times 10^{4}$ & $2.54 \times 10^{2}$ & 1 & 0.02 \\
\bottomrule
\end{tabular}
\end{table}

\begin{table}[t]
\centering
\caption{Layer-wise distribution of relative change rates ($R_k$) for a single edit on DeepSeek-R1-Distill-Llama-8B.}
\label{tab:rk_distribution_deepseek}
\small
\begin{tabular}{l c r r r r c c}
\toprule
Method & Layer & $R_k$ mean & $R_k$ min & $R_k$ max & Std & num($R_k>1$) & \%($R_k>1$) \\
\midrule
AlphaEdit & 5  & $3.37 \times 10^{2}$ & $7.17 \times 10^{-2}$ & $1.38 \times 10^{6}$ & $2.15 \times 10^{4}$ & 4095 & 99.98 \\
          & 10 & $3.32 \times 10^{2}$ & $9.78 \times 10^{-2}$ & $1.35 \times 10^{6}$ & $2.12 \times 10^{4}$ & 4095 & 99.98 \\
          & 15 & $3.28 \times 10^{2}$ & $1.30 \times 10^{-1}$ & $1.34 \times 10^{6}$ & $2.09 \times 10^{4}$ & 4095 & 99.98 \\
          & 20 & $3.49 \times 10^{2}$ & $1.62 \times 10^{-1}$ & $1.43 \times 10^{6}$ & $2.23 \times 10^{4}$ & 4095 & 99.98 \\
          & 25 & $3.68 \times 10^{2}$ & $2.15 \times 10^{-1}$ & $1.50 \times 10^{6}$ & $2.35 \times 10^{4}$ & 4095 & 99.98 \\
          & 30 & $2.01 \times 10^{2}$ & $1.49 \times 10^{-1}$ & $8.18 \times 10^{5}$ & $1.28 \times 10^{4}$ & 4095 & 99.98 \\
          & 31 & $1.39 \times 10^{0}$ & $3.00 \times 10^{-4}$ & $5.66 \times 10^{3}$ & $8.84 \times 10^{1}$ & 1 & 0.02 \\
\midrule
MEMIT & 5  & $3.37 \times 10^{2}$ & $7.17 \times 10^{-2}$ & $1.38 \times 10^{6}$ & $2.15 \times 10^{4}$ & 4095 & 99.98 \\
      & 10 & $3.32 \times 10^{2}$ & $9.79 \times 10^{-2}$ & $1.35 \times 10^{6}$ & $2.12 \times 10^{4}$ & 4095 & 99.98 \\
      & 15 & $3.28 \times 10^{2}$ & $1.30 \times 10^{-1}$ & $1.34 \times 10^{6}$ & $2.09 \times 10^{4}$ & 4095 & 99.98 \\
      & 20 & $3.49 \times 10^{2}$ & $1.62 \times 10^{-1}$ & $1.43 \times 10^{6}$ & $2.23 \times 10^{4}$ & 4095 & 99.98 \\
      & 25 & $3.68 \times 10^{2}$ & $2.15 \times 10^{-1}$ & $1.50 \times 10^{6}$ & $2.35 \times 10^{4}$ & 4095 & 99.98 \\
      & 30 & $2.01 \times 10^{2}$ & $1.49 \times 10^{-1}$ & $8.18 \times 10^{5}$ & $1.28 \times 10^{4}$ & 4095 & 99.98 \\
      & 31 & $4.35 \times 10^{-1}$ & $1.00 \times 10^{-4}$ & $1.76 \times 10^{3}$ & $2.76 \times 10^{1}$ & 1 & 0.02 \\
\midrule
MEND & 29 & $8.32 \times 10^{2}$ & $9.20 \times 10^{-3}$ & $3.41 \times 10^{6}$ & $5.32 \times 10^{4}$ & 5 & 0.12 \\
     & 30 & $2.08 \times 10^{2}$ & $1.49 \times 10^{-1}$ & $8.46 \times 10^{5}$ & $1.32 \times 10^{4}$ & 4095 & 99.98 \\
     & 31 & $1.74 \times 10^{1}$ & $1.08 \times 10^{-2}$ & $7.11 \times 10^{4}$ & $1.11 \times 10^{3}$ & 1 & 0.02 \\
\midrule
ROME & 5  & $3.54 \times 10^{2}$ & $7.33 \times 10^{-2}$ & $1.44 \times 10^{6}$ & $2.26 \times 10^{4}$ & 4095 & 99.98 \\
      & 10 & $3.32 \times 10^{2}$ & $9.78 \times 10^{-2}$ & $1.36 \times 10^{6}$ & $2.12 \times 10^{4}$ & 4095 & 99.98 \\
      & 15 & $3.28 \times 10^{2}$ & $1.31 \times 10^{-1}$ & $1.34 \times 10^{6}$ & $2.09 \times 10^{4}$ & 4095 & 99.98 \\
      & 20 & $3.49 \times 10^{2}$ & $1.62 \times 10^{-1}$ & $1.43 \times 10^{6}$ & $2.23 \times 10^{4}$ & 4095 & 99.98 \\
      & 25 & $3.68 \times 10^{2}$ & $2.15 \times 10^{-1}$ & $1.50 \times 10^{6}$ & $2.35 \times 10^{4}$ & 4095 & 99.98 \\
      & 30 & $2.01 \times 10^{2}$ & $1.49 \times 10^{-1}$ & $8.18 \times 10^{5}$ & $1.28 \times 10^{4}$ & 4095 & 99.98 \\
      & 31 & $6.91 \times 10^{0}$ & $1.60 \times 10^{-3}$ & $2.81 \times 10^{4}$ & $4.39 \times 10^{2}$ & 1 & 0.02 \\
\midrule
WISE & 29 & $1.25 \times 10^{4}$ & $7.32 \times 10^{-2}$ & $5.10 \times 10^{7}$ & $7.97 \times 10^{5}$ & 3615 & 88.26 \\
     & 30 & $2.05 \times 10^{2}$ & $2.39 \times 10^{-1}$ & $8.33 \times 10^{5}$ & $1.30 \times 10^{4}$ & 4093 & 99.93 \\
     & 31 & $3.60 \times 10^{2}$ & $1.60 \times 10^{-1}$ & $1.47 \times 10^{6}$ & $2.30 \times 10^{4}$ & 1607 & 39.23 \\
\bottomrule
\end{tabular}
\end{table}

\begin{table}[t]
\centering
\caption{Layer-wise distribution of relative change rates ($R_k$) for a single edit on Llama-2-13B.}
\label{tab:rk_distribution_llama2}
\small
\begin{tabular}{l c r r r r c c}
\toprule
Method & Layer & $R_k$ mean & $R_k$ min & $R_k$ max & Std & num($R_k>1$) & \%($R_k>1$) \\
\midrule
AlphaEdit & 5  & $2.69 \times 10^{2}$ & $1.42 \times 10^{-1}$ & $1.37 \times 10^{6}$ & $1.92 \times 10^{4}$ & 5119 & 99.98 \\
          & 10 & $2.83 \times 10^{2}$ & $2.04 \times 10^{-1}$ & $1.44 \times 10^{6}$ & $2.01 \times 10^{4}$ & 5119 & 99.98 \\
          & 15 & $2.91 \times 10^{2}$ & $2.59 \times 10^{-1}$ & $1.48 \times 10^{6}$ & $2.07 \times 10^{4}$ & 5119 & 99.98 \\
          & 20 & $2.86 \times 10^{2}$ & $3.54 \times 10^{-1}$ & $1.46 \times 10^{6}$ & $2.04 \times 10^{4}$ & 5117 & 99.94 \\
          & 25 & $2.91 \times 10^{2}$ & $3.30 \times 10^{-1}$ & $1.48 \times 10^{6}$ & $2.07 \times 10^{4}$ & 5118 & 99.96 \\
          & 30 & $2.66 \times 10^{2}$ & $2.72 \times 10^{-1}$ & $1.36 \times 10^{6}$ & $1.90 \times 10^{4}$ & 5119 & 99.98 \\
          & 35 & $6.07 \times 10^{-1}$ & $3.00 \times 10^{-4}$ & $3.09 \times 10^{3}$ & $4.32 \times 10^{1}$ & 1 & 0.02 \\
          & 39 & $2.36 \times 10^{-1}$ & $3.00 \times 10^{-4}$ & $1.18 \times 10^{3}$ & $1.65 \times 10^{1}$ & 1 & 0.02 \\
\midrule
MEMIT & 5  & $2.69 \times 10^{2}$ & $1.42 \times 10^{-1}$ & $1.37 \times 10^{6}$ & $1.92 \times 10^{4}$ & 5119 & 99.98 \\
      & 10 & $2.83 \times 10^{2}$ & $2.04 \times 10^{-1}$ & $1.44 \times 10^{6}$ & $2.01 \times 10^{4}$ & 5119 & 99.98 \\
      & 15 & $2.91 \times 10^{2}$ & $2.59 \times 10^{-1}$ & $1.48 \times 10^{6}$ & $2.07 \times 10^{4}$ & 5119 & 99.98 \\
      & 20 & $2.86 \times 10^{2}$ & $3.54 \times 10^{-1}$ & $1.46 \times 10^{6}$ & $2.04 \times 10^{4}$ & 5117 & 99.94 \\
      & 25 & $2.91 \times 10^{2}$ & $3.30 \times 10^{-1}$ & $1.48 \times 10^{6}$ & $2.07 \times 10^{4}$ & 5118 & 99.96 \\
      & 30 & $2.66 \times 10^{2}$ & $2.72 \times 10^{-1}$ & $1.36 \times 10^{6}$ & $1.90 \times 10^{4}$ & 5119 & 99.98 \\
      & 35 & $1.65 \times 10^{0}$ & $7.00 \times 10^{-4}$ & $8.38 \times 10^{3}$ & $1.17 \times 10^{2}$ & 1 & 0.02 \\
      & 39 & $5.99 \times 10^{-1}$ & $9.00 \times 10^{-4}$ & $3.01 \times 10^{3}$ & $4.21 \times 10^{1}$ & 1 & 0.02 \\
\midrule
MEND & 37 & $4.24 \times 10^{0}$ & $3.00 \times 10^{-4}$ & $2.15 \times 10^{4}$ & $3.01 \times 10^{2}$ & 1 & 0.02 \\
     & 38 & $2.46 \times 10^{0}$ & $5.00 \times 10^{-4}$ & $1.23 \times 10^{4}$ & $1.72 \times 10^{2}$ & 1 & 0.02 \\
     & 39 & $1.67 \times 10^{0}$ & $4.70 \times 10^{-3}$ & $7.80 \times 10^{3}$ & $1.09 \times 10^{2}$ & 2 & 0.04 \\
\midrule
ROME & 5  & $2.70 \times 10^{2}$ & $1.42 \times 10^{-1}$ & $1.37 \times 10^{6}$ & $1.92 \times 10^{4}$ & 5119 & 99.98 \\
      & 10 & $2.83 \times 10^{2}$ & $2.03 \times 10^{-1}$ & $1.44 \times 10^{6}$ & $2.01 \times 10^{4}$ & 5119 & 99.98 \\
      & 15 & $2.91 \times 10^{2}$ & $2.59 \times 10^{-1}$ & $1.48 \times 10^{6}$ & $2.07 \times 10^{4}$ & 5119 & 99.98 \\
      & 20 & $2.86 \times 10^{2}$ & $3.53 \times 10^{-1}$ & $1.46 \times 10^{6}$ & $2.04 \times 10^{4}$ & 5117 & 99.94 \\
      & 25 & $2.91 \times 10^{2}$ & $3.30 \times 10^{-1}$ & $1.48 \times 10^{6}$ & $2.07 \times 10^{4}$ & 5118 & 99.96 \\
      & 30 & $2.66 \times 10^{2}$ & $2.72 \times 10^{-1}$ & $1.36 \times 10^{6}$ & $1.90 \times 10^{4}$ & 5119 & 99.98 \\
      & 35 & $3.04 \times 10^{0}$ & $1.40 \times 10^{-3}$ & $1.55 \times 10^{4}$ & $2.16 \times 10^{2}$ & 1 & 0.02 \\
      & 39 & $1.05 \times 10^{0}$ & $1.60 \times 10^{-3}$ & $5.26 \times 10^{3}$ & $7.35 \times 10^{1}$ & 1 & 0.02 \\
\midrule
WISE & 27 & $1.55 \times 10^{0}$ & $1.00 \times 10^{-3}$ & $7.87 \times 10^{3}$ & $1.10 \times 10^{2}$ & 1 & 0.02 \\
     & 30 & $2.66 \times 10^{2}$ & $2.72 \times 10^{-1}$ & $1.36 \times 10^{6}$ & $1.90 \times 10^{4}$ & 5119 & 99.98 \\
     & 35 & $1.92 \times 10^{0}$ & $9.00 \times 10^{-4}$ & $9.79 \times 10^{3}$ & $1.37 \times 10^{2}$ & 1 & 0.02 \\
     & 39 & $7.42 \times 10^{0}$ & $3.90 \times 10^{-3}$ & $3.77 \times 10^{4}$ & $5.27 \times 10^{2}$ & 1 & 0.02 \\
\bottomrule
\end{tabular}
\end{table}

% \begin{table*}[htbp]
%     \centering
%     \caption{Layer-wise distribution of relative change rates ($R_k$) for a single edit on T5-3B under MEND editing. 
%     % Even a single edit can produce $R_k > 1$ in the majority of directions in some layers, with extremely large mean and variance. (The table reports mean, minimum, maximum, standard deviation, and the proportion of directions with $R_k>1$ for each layer.)
%     }
%     \label{tab:rk_distribution_t5}
%     \begin{tabular}{c|ccccr}
%         \toprule
%         Layer & $R_k$ mean & $R_k$ min & $R_k$ max & Std & num($R_k>1$) \\
%         \midrule
%         Enc22-FFN-wi & $2.20\times10^{17}$ & 0 & $1.00\times10^{20}$ & $\infty$ & 47 (0.29\%) \\
%         Enc22-FFN-wo & 0.2160 & $1.41\times10^{-4}$ & 14.2335 & 0.5732 & 11 (1.07\%) \\
%         Enc23-FFN-wi & $4.03\times10^{17}$ & 0 & $1.00\times10^{20}$ & $\infty$ & 72 (0.44\%) \\
%         Enc23-FFN-wo & 5.3797 & 0.0076 & 1482.6935 & 46.3268 & 820 (80.08\%) \\
%         Dec22-FFN-wi & $1.86\times10^{19}$ & 0.0988 & $1.00\times10^{20}$ & $\infty$ & 5864 (35.79\%) \\
%         Dec22-FFN-wo & 0.4325 & 0.1657 & 19.7270 & 0.6070 & 2 (0.20\%) \\
%         Dec23-FFN-wi & $4.63\times10^{18}$ & 0.1755 & $1.00\times10^{20}$ & $\infty$ & 3243 (19.79\%) \\
%         Dec23-FFN-wo & 3.0178 & 0.1114 & 199.9235 & 7.4669 & 717 (70.02\%) \\
%         \bottomrule
%     \end{tabular}
% \end{table*}

\begin{table}[t]
    \centering
    \caption{Layer-wise distribution of relative change rates ($R_k$) for a single edit on T5-3B under MEND editing.
    Entries shown as $1.00\times10^{20}$ indicate numerically unbounded variance.}
    \label{tab:rk_distribution_t5}
    \small
    \begin{tabular}{l r r r r c c}
    \toprule
    Layer & $R_k$ mean & $R_k$ min & $R_k$ max & Std & num($R_k>1$) & \%($R_k>1$) \\
    \midrule
    Enc22-FFN-wi & $2.20 \times 10^{17}$ & $0.00 \times 10^{0}$ & $1.00 \times 10^{20}$ & $1.00 \times 10^{20}$ & 47  & 0.29 \\
    Enc22-FFN-wo & $2.16 \times 10^{-1}$ & $1.41 \times 10^{-4}$ & $1.42 \times 10^{1}$  & $5.73 \times 10^{-1}$  & 11  & 1.07 \\
    Enc23-FFN-wi & $4.03 \times 10^{17}$ & $0.00 \times 10^{0}$ & $1.00 \times 10^{20}$ & $1.00 \times 10^{20}$ & 72  & 0.44 \\
    Enc23-FFN-wo & $5.38 \times 10^{0}$  & $7.60 \times 10^{-3}$ & $1.48 \times 10^{3}$  & $4.63 \times 10^{1}$  & 820 & 80.08 \\
    Dec22-FFN-wi & $1.86 \times 10^{19}$ & $9.88 \times 10^{-2}$ & $1.00 \times 10^{20}$ & $1.00 \times 10^{20}$ & 5864 & 35.79 \\
    Dec22-FFN-wo & $4.33 \times 10^{-1}$ & $1.66 \times 10^{-1}$ & $1.97 \times 10^{1}$  & $6.07 \times 10^{-1}$  & 2   & 0.20 \\
    Dec23-FFN-wi & $4.63 \times 10^{18}$ & $1.76 \times 10^{-1}$ & $1.00 \times 10^{20}$ & $1.00 \times 10^{20}$ & 3243 & 19.79 \\
    Dec23-FFN-wo & $3.02 \times 10^{0}$  & $1.11 \times 10^{-1}$ & $2.00 \times 10^{2}$  & $7.47 \times 10^{0}$  & 717 & 70.02 \\
    \bottomrule
    \end{tabular}
\end{table}

\subsection{Functional Relevance: Correlation with Editing Performance}
\label{sec:functional_relevance}

This subsection examines whether perturbation amplification is functionally associated with practical editing degradation. 
Rather than treating amplification at the individual layer level, we evaluate whether \emph{step-level} summaries of the layer-wise $R_k$ distribution correlate with step-level editing performance metrics.

% \paragraph{Step-level amplification summaries.}
\textbf{Step-level $R_k$ Summaries.}
For each editing method and edit step, we first summarize the layer-wise $R_k$ distribution using several scalar features, including the median, upper quantiles (q90 and q95), and trimmed mean. 
Because the editing metrics are defined at the step level rather than the layer level, we then aggregate these layer-wise summaries across edited layers to obtain a single step-level amplification statistic. 
To test the robustness of the observed trends, we consider three cross-layer aggregation strategies: mean, max, and median.

% \paragraph{Correlation protocol.}
\textbf{Correlation Protocol.}
% We measure the association between step-level amplification summaries and editing performance using Spearman's rank correlation coefficient. 
% We use Spearman correlation because the number of valid edit steps is limited for several methods and the relationship between amplification and performance degradation is not necessarily linear. 
% Importantly, we do not filter out metric values equal to zero, since these correspond to real measured failure states rather than missing or corrupted observations. 
% Accordingly, the reported correlations should be interpreted as characterizing the relationship between amplification and the \emph{full degradation trajectory}, including complete editing failure.
We measure the association between step-level statistics derived from $R_k$ and editing performance using Spearman’s rank correlation coefficient. 
We use Spearman correlation because the number of valid edit steps is limited for several methods and the relationship between amplification and performance degradation is not necessarily linear. 
We exclude MEND from this analysis because its performance under sequential editing degrades to zero too quickly, leaving insufficient variation for a meaningful correlation analysis.

% \paragraph{Main findings.}
\textbf{Main Findings.}
Table~\ref{tab:functional_relevance_main} reports the main specification, using the median of the layer-wise $R_k$ distribution followed by mean aggregation across edited layers. 
Under this default setting, ROME and MEMIT exhibit consistently strong negative correlations between amplification and multiple editing metrics, especially reliability, locality, and portability. 
WISE shows weaker but still partially aligned negative trends, 
while AlphaEdit exhibits a qualitatively different pattern: its correlations are generally weak across metrics, possibly because its null-space-constrained update restricts parameter changes to directions designed to preserve existing knowledge and reduce interference. 
These observations indicate that larger amplification is systematically associated with worse editing outcomes for several parameter-based editing methods, although the strength and direction of this association remain method-dependent.

\textbf{Complete Spearman Correlation Results for Step-Level $R_k$ Statistics}

We report the complete Spearman correlation results between editing performance metrics and step-level statistics derived from $R_k$ for all summary definitions considered in this work. 
Across all settings, we compare three cross-layer aggregation rules over the resulting layer-wise statistics: \textbf{mean}, \textbf{max}, and \textbf{median}. The results are
summarized in the following tables:

\begin{itemize}
    \item Table~\ref{tab:fr_median_full} reports the full results for the \textbf{median} summary.
    \item Table~\ref{tab:fr_q90_full} reports the full results for the \textbf{q90} summary.
    \item Table~\ref{tab:fr_q95_full} reports the full results for the \textbf{q95} summary.
    \item Table~\ref{tab:fr_trimmed_mean_full} reports the full results for the \textbf{trimmed mean} summary.
\end{itemize}

Across summary choices and aggregation strategies, the qualitative pattern remains stable. 
ROME and MEMIT consistently show strong negative correlations between amplification and degradation-sensitive metrics, whereas AlphaEdit exhibits weaker associations overall and occasionally positive correlations with generalization-related metrics. 
This suggests that the observed correlation structure is robust to both the choice of $R_k$-based summary statistic and the cross-layer aggregation rule.

% We report the complete Spearman correlation results between editing performance metrics and step-level statistics derived from the layer-wise $R_k$ values. 

% \begin{itemize}
%     \item Table~\ref{tab:fr_median_full} reports the complete results when the within-layer statistic is defined as the \textbf{median} of the corresponding $R_k$ distribution.
%     \item Table~\ref{tab:fr_q90_full} reports the corresponding results when the within-layer statistic is defined as the \textbf{90th percentile (q90)} of the corresponding $R_k$ distribution.
%     \item Table~\ref{tab:fr_q95_full} reports the corresponding results when the within-layer statistic is defined as the \textbf{95th percentile (q95)} of the corresponding $R_k$ distribution.
%     \item Table~\ref{tab:fr_trimmed_mean_full} reports the corresponding results when the within-layer statistic is defined as the \textbf{trimmed mean} of the corresponding $R_k$ distribution.
% \end{itemize}
% Within each table, we additionally compare three cross-layer aggregation rules over $R_k$, namely \textbf{mean}, \textbf{max}, and \textbf{median}.

\begin{table*}[p]
\centering
\caption{
Complete Spearman correlation results for the \textbf{median} of the layer-wise $R_k$ distribution, with no filtering of zero-valued performance observations. 
For each metric and editing method, entries report Spearman's $\rho$ with significance markers, followed by the corresponding $p$-value in parentheses. 
Results are shown for three cross-layer aggregation strategies: mean, max, and median. 
% Sample sizes are $n=9$ for AlphaEdit, MEMIT, ROME, and WISE, and $n=3$ for MEND.
\textit{Note.} 
$^{*}p<0.05$, $^{**}p<0.01$, $^{***}p<0.001$.
}
\label{tab:fr_median_full}
\scriptsize
\setlength{\tabcolsep}{4pt}
\begin{tabular}{l|l|ccccc}
\toprule
Aggregation & Metric & AlphaEdit & MEMIT & 
% MEND & 
ROME & WISE \\
\midrule
\multirow{4}{*}{Mean}
& Rel.    
& $-0.2034$ $(0.5996)$ 
& $-0.9619^{***}$ $(<0.001)$ 
% & $-0.8660$ $(0.3333)$ 
& $-0.8645^{**}$ $(0.0026)$ 
& $-0.7679^{*}$ $(0.0157)$ \\
& Gen.
& $ 0.6360$ $(0.0656)$ 
& $-0.4238$ $(0.2557)$ 
% & $-0.8660$ $(0.3333)$ 
& $-0.8645^{**}$ $(0.0026)$ 
& $-0.0840$ $(0.8298)$ \\
& Loc.      
& $ 0.0339$ $(0.9310)$ 
& $-0.7289^{*}$ $(0.0259)$ 
% & $-0.8660$ $(0.3333)$ 
& $-0.9129^{***}$ $(<0.001)$ 
& $-0.4851$ $(0.1857)$ \\
& Port.   
& $ 0.3500$ $(0.3558)$ 
& $-0.8087^{**}$ $(0.0083)$ 
% & $-0.8660$ $(0.3333)$ 
& $-0.8617^{**}$ $(0.0028)$ 
& $-0.6639$ $(0.0512)$ \\
\midrule
\multirow{4}{*}{Max}
& Rel.   
& $-0.2034$ $(0.5996)$ 
& $-0.9279^{***}$ $(<0.001)$ 
% & $-0.8660$ $(0.3333)$ 
& $-0.8645^{**}$ $(0.0026)$ 
& $-0.7679^{*}$ $(0.0157)$ \\
& Gen.
& $ 0.5439$ $(0.1301)$ 
& $-0.4577$ $(0.2154)$ 
% & $-0.8660$ $(0.3333)$ 
& $-0.8645^{**}$ $(0.0026)$ 
& $-0.0336$ $(0.9316)$ \\
& Loc.    
& $ 0.0678$ $(0.8624)$ 
& $-0.7120^{*}$ $(0.0314)$ 
% & $-0.8660$ $(0.3333)$ 
& $-0.9129^{***}$ $(<0.001)$ 
& $-0.4331$ $(0.2442)$ \\
& Port.  
& $ 0.3667$ $(0.3317)$ 
& $-0.8087^{**}$ $(0.0083)$ 
% & $-0.8660$ $(0.3333)$ 
& $-0.8617^{**}$ $(0.0028)$ 
& $-0.6134$ $(0.0789)$ \\
\midrule
\multirow{4}{*}{Median}
& Rel.   
& $ 0.0000$ $(1.0000)$ 
& $-0.7746^{*}$ $(0.0142)$ 
% & $-0.8660$ $(0.3333)$ 
& $-0.8476^{**}$ $(0.0039)$ 
& $-0.6245$ $(0.0722)$ \\
& Gen.
& $ 0.7782^{*}$ $(0.0135)$ 
& $-0.5255$ $(0.1462)$ 
% & $-0.8660$ $(0.3333)$ 
& $-0.8476^{**}$ $(0.0039)$ 
& $-0.2605$ $(0.4984)$ \\
& Loc.   
& $ 0.2712$ $(0.4802)$ 
& $-0.9662^{***}$ $(<0.001)$ 
% & $-0.8660$ $(0.3333)$ 
& $-0.8946^{**}$ $(0.0011)$ 
& $-0.3811$ $(0.3115)$ \\
& Port. 
& $ 0.4167$ $(0.2646)$ 
& $-0.9789^{***}$ $(<0.001)$ 
% & $-0.8660$ $(0.3333)$ 
& $-0.8791^{**}$ $(0.0018)$ 
& $-0.7731^{*}$ $(0.0146)$ \\
\bottomrule
\end{tabular}

% \vspace{0.4em}
% \raggedright
% \footnotesize
% \textit{Note.} 
% $^{*}p<0.05$, $^{**}p<0.01$, $^{***}p<0.001$.
\end{table*}

\begin{table*}[p]
\centering
\caption{
Complete Spearman correlation results for the \textbf{q90} of the layer-wise $R_k$ distribution, with no filtering of zero-valued performance observations. 
For each metric and editing method, entries report Spearman's $\rho$ with significance markers, followed by the corresponding $p$-value in parentheses. 
Results are shown for three cross-layer aggregation strategies: mean, max, and median. 
% Sample sizes are $n=9$ for AlphaEdit, MEMIT, ROME, and WISE, and $n=3$ for MEND.
\textit{Note.} 
$^{*}p<0.05$, $^{**}p<0.01$, $^{***}p<0.001$.
}
\label{tab:fr_q90_full}
\scriptsize
\setlength{\tabcolsep}{4pt}
\begin{tabular}{l|l|ccccc}
\toprule
Aggregation & Metric & AlphaEdit & MEMIT & 
% MEND & 
ROME & WISE \\
\midrule
\multirow{4}{*}{Mean}
& Rel.   
& $-0.2034$ $(0.5996)$ 
& $-0.9619^{***}$ $(<0.001)$ 
% & $-0.8660$ $(0.3333)$ 
& $-0.8645^{**}$ $(0.0026)$ 
& $-0.7679^{*}$ $(0.0157)$ \\
& Gen.
& $ 0.6360$ $(0.0656)$ 
& $-0.4238$ $(0.2557)$ 
% & $-0.8660$ $(0.3333)$ 
& $-0.8645^{**}$ $(0.0026)$ 
& $-0.0840$ $(0.8298)$ \\
& Loc.       
& $ 0.0339$ $(0.9310)$ 
& $-0.7289^{*}$ $(0.0259)$ 
% & $-0.8660$ $(0.3333)$ 
& $-0.9129^{***}$ $(<0.001)$ 
& $-0.4851$ $(0.1857)$ \\
& Port.   
& $ 0.3500$ $(0.3558)$ 
& $-0.8087^{**}$ $(0.0083)$ 
% & $-0.8660$ $(0.3333)$ 
& $-0.8617^{**}$ $(0.0028)$ 
& $-0.6639$ $(0.0512)$ \\
\midrule
\multirow{4}{*}{Max}
& Rel.   
& $-0.2034$ $(0.5996)$ 
& $-0.9279^{***}$ $(<0.001)$ 
% & $-0.8660$ $(0.3333)$ 
& $-0.8645^{**}$ $(0.0026)$ 
& $-0.7679^{*}$ $(0.0157)$ \\
& Gen. 
& $ 0.5439$ $(0.1301)$ 
& $-0.4577$ $(0.2154)$ 
% & $-0.8660$ $(0.3333)$ 
& $-0.8645^{**}$ $(0.0026)$ 
& $-0.0336$ $(0.9316)$ \\
& Loc.     
& $ 0.0678$ $(0.8624)$ 
& $-0.7120^{*}$ $(0.0314)$ 
% & $-0.8660$ $(0.3333)$ 
& $-0.9129^{***}$ $(<0.001)$ 
& $-0.4331$ $(0.2442)$ \\
& Port.    
& $ 0.3667$ $(0.3317)$ 
& $-0.8087^{**}$ $(0.0083)$ 
% & $-0.8660$ $(0.3333)$ 
& $-0.8617^{**}$ $(0.0028)$ 
& $-0.6134$ $(0.0789)$ \\
\midrule
\multirow{4}{*}{Median}
& Rel.    
& $-0.2034$ $(0.5996)$ 
& $-0.7746^{*}$ $(0.0142)$ 
% & $-0.8660$ $(0.3333)$ 
& $-0.8645^{**}$ $(0.0026)$ 
& $-0.7679^{*}$ $(0.0157)$ \\
& Gen.
& $ 0.6360$ $(0.0656)$ 
& $-0.5255$ $(0.1462)$ 
% & $-0.8660$ $(0.3333)$ 
& $-0.8645^{**}$ $(0.0026)$ 
& $-0.0840$ $(0.8298)$ \\
& Loc.      
& $ 0.0339$ $(0.9310)$ 
& $-0.9662^{***}$ $(<0.001)$ 
% & $-0.8660$ $(0.3333)$ 
& $-0.9129^{***}$ $(<0.001)$ 
& $-0.4851$ $(0.1857)$ \\
& Port.  
& $ 0.3500$ $(0.3558)$ 
& $-0.9789^{***}$ $(<0.001)$ 
% & $-0.8660$ $(0.3333)$ 
& $-0.8617^{**}$ $(0.0028)$ 
& $-0.6639$ $(0.0512)$ \\
\bottomrule
\end{tabular}

% \vspace{0.4em}
% \raggedright
% \footnotesize
% \textit{Note.} 
% $^{*}p<0.05$, $^{**}p<0.01$, $^{***}p<0.001$.
\end{table*}

\begin{table*}[p]
\centering
\caption{
Complete Spearman correlation results for the \textbf{q95} of the layer-wise $R_k$ distribution, with no filtering of zero-valued performance observations. 
For each metric and editing method, entries report Spearman's $\rho$ with significance markers, followed by the corresponding $p$-value in parentheses. 
Results are shown for three cross-layer aggregation strategies: mean, max, and median. 
% Sample sizes are $n=9$ for AlphaEdit, MEMIT, ROME, and WISE, and $n=3$ for MEND.
\textit{Note.} 
$^{*}p<0.05$, $^{**}p<0.01$, $^{***}p<0.001$.
}
\label{tab:fr_q95_full}
\scriptsize
\setlength{\tabcolsep}{4pt}
\begin{tabular}{l|l|ccccc}
\toprule
Aggregation & Metric & AlphaEdit & MEMIT & 
% MEND & 
ROME & WISE \\
\midrule
\multirow{4}{*}{Mean}
& Rel.  
& $-0.2034$ $(0.5996)$ 
& $-0.9619^{***}$ $(<0.001)$ 
% & $-0.8660$ $(0.3333)$ 
& $-0.8645^{**}$ $(0.0026)$ 
& $-0.7679^{*}$ $(0.0157)$ \\
& Gen.
& $ 0.6360$ $(0.0656)$ 
& $-0.4238$ $(0.2557)$ 
% & $-0.8660$ $(0.3333)$ 
& $-0.8645^{**}$ $(0.0026)$ 
& $-0.0840$ $(0.8298)$ \\
& Loc.      
& $ 0.0339$ $(0.9310)$ 
& $-0.7289^{*}$ $(0.0259)$ 
% & $-0.8660$ $(0.3333)$ 
& $-0.9129^{***}$ $(<0.001)$ 
& $-0.4851$ $(0.1857)$ \\
& Port.   
& $ 0.3500$ $(0.3558)$ 
& $-0.8087^{**}$ $(0.0083)$ 
% & $-0.8660$ $(0.3333)$ 
& $-0.8617^{**}$ $(0.0028)$ 
& $-0.6639$ $(0.0512)$ \\
\midrule
\multirow{4}{*}{Max}
& Rel.   
& $-0.2034$ $(0.5996)$ 
& $-0.9279^{***}$ $(<0.001)$ 
% & $-0.8660$ $(0.3333)$ 
& $-0.8645^{**}$ $(0.0026)$ 
& $-0.7679^{*}$ $(0.0157)$ \\
& Gene.
& $ 0.6360$ $(0.0656)$ 
& $-0.4577$ $(0.2154)$ 
% & $-0.8660$ $(0.3333)$ 
& $-0.8645^{**}$ $(0.0026)$ 
& $-0.0336$ $(0.9316)$ \\
& Loc.      
& $ 0.0339$ $(0.9310)$ 
& $-0.7120^{*}$ $(0.0314)$ 
% & $-0.8660$ $(0.3333)$ 
& $-0.9129^{***}$ $(<0.001)$ 
& $-0.4331$ $(0.2442)$ \\
& Port.
& $ 0.3500$ $(0.3558)$ 
& $-0.8087^{**}$ $(0.0083)$ 
% & $-0.8660$ $(0.3333)$ 
& $-0.8617^{**}$ $(0.0028)$ 
& $-0.6134$ $(0.0789)$ \\
\midrule
\multirow{4}{*}{Median}
& Rel.    
& $-0.2034$ $(0.5996)$ 
& $-0.8598^{**}$ $(0.0030)$ 
% & $-0.8660$ $(0.3333)$ 
& $-0.8645^{**}$ $(0.0026)$ 
& $-0.7679^{*}$ $(0.0157)$ \\
& Gen. 
& $ 0.6360$ $(0.0656)$ 
& $-0.4577$ $(0.2154)$ 
% & $-0.8660$ $(0.3333)$ 
& $-0.8645^{**}$ $(0.0026)$ 
& $-0.0840$ $(0.8298)$ \\
& Loc.   
& $ 0.0339$ $(0.9310)$ 
& $-0.9323^{***}$ $(<0.001)$ 
% & $-0.8660$ $(0.3333)$ 
& $-0.9129^{***}$ $(<0.001)$ 
& $-0.4851$ $(0.1857)$ \\
& Port. 
& $ 0.3500$ $(0.3558)$ 
& $-0.9534^{***}$ $(<0.001)$ 
% & $-0.8660$ $(0.3333)$ 
& $-0.8617^{**}$ $(0.0028)$ 
& $-0.6639$ $(0.0512)$ \\
\bottomrule
\end{tabular}

% \vspace{0.4em}
% \raggedright
% \footnotesize
% \textit{Note.} 
% $^{*}p<0.05$, $^{**}p<0.01$, $^{***}p<0.001$.
\end{table*}

\begin{table*}[h]
\centering
\caption{
Complete Spearman correlation results for the \textbf{trimmed mean} of the layer-wise $R_k$ distribution, with no filtering of zero-valued performance observations. 
For each metric and editing method, entries report Spearman's $\rho$ with significance markers, followed by the corresponding $p$-value in parentheses. 
Results are shown for three cross-layer aggregation strategies: mean, max, and median. 
% Sample sizes are $n=9$ for AlphaEdit, MEMIT, ROME, and WISE, and $n=3$ for MEND.
\textit{Note.} 
$^{*}p<0.05$, $^{**}p<0.01$, $^{***}p<0.001$.
}
\label{tab:fr_trimmed_mean_full}
\scriptsize
\setlength{\tabcolsep}{4pt}
\begin{tabular}{l|l|ccccc}
\toprule
Aggregation & Metric & AlphaEdit & MEMIT & 
% MEND & 
ROME & WISE \\
\midrule
\multirow{4}{*}{Mean}
& Rel.  
& $-0.2034$ $(0.5996)$ 
& $-0.9619^{***}$ $(<0.001)$ 
% & $-0.8660$ $(0.3333)$ 
& $-0.8645^{**}$ $(0.0026)$ 
& $-0.7679^{*}$ $(0.0157)$ \\
& Gen.
& $ 0.6360$ $(0.0656)$ 
& $-0.4238$ $(0.2557)$ 
% & $-0.8660$ $(0.3333)$ 
& $-0.8645^{**}$ $(0.0026)$ 
& $-0.0840$ $(0.8298)$ \\
& Loc.      
& $ 0.0339$ $(0.9310)$ 
& $-0.7289^{*}$ $(0.0259)$ 
% & $-0.8660$ $(0.3333)$ 
& $-0.9129^{***}$ $(<0.001)$ 
& $-0.4851$ $(0.1857)$ \\
& Port.   
& $ 0.3500$ $(0.3558)$ 
& $-0.8087^{**}$ $(0.0083)$ 
% & $-0.8660$ $(0.3333)$ 
& $-0.8617^{**}$ $(0.0028)$ 
& $-0.6639$ $(0.0512)$ \\
\midrule
\multirow{4}{*}{Max}
& Rel.  
& $-0.2034$ $(0.5996)$ 
& $-0.9279^{***}$ $(<0.001)$ 
% & $-0.8660$ $(0.3333)$ 
& $-0.8645^{**}$ $(0.0026)$ 
& $-0.7679^{*}$ $(0.0157)$ \\
& Gen.
& $ 0.5439$ $(0.1301)$ 
& $-0.4577$ $(0.2154)$ 
% & $-0.8660$ $(0.3333)$ 
& $-0.8645^{**}$ $(0.0026)$ 
& $-0.0336$ $(0.9316)$ \\
& Loc.     
& $ 0.0678$ $(0.8624)$ 
& $-0.7120^{*}$ $(0.0314)$ 
% & $-0.8660$ $(0.3333)$ 
& $-0.9129^{***}$ $(<0.001)$ 
& $-0.4331$ $(0.2442)$ \\
& Port.  
& $ 0.3667$ $(0.3317)$ 
& $-0.8087^{**}$ $(0.0083)$ 
% & $-0.8660$ $(0.3333)$ 
& $-0.8617^{**}$ $(0.0028)$ 
& $-0.6134$ $(0.0789)$ \\
\midrule
\multirow{4}{*}{Median}
& Rel.   
& $ 0.1017$ $(0.7946)$ 
& $-0.7746^{*}$ $(0.0142)$ 
% & $-0.8660$ $(0.3333)$ 
& $-0.8476^{**}$ $(0.0039)$ 
& $-0.6245$ $(0.0722)$ \\
& Gen.
& $ 0.7699^{*}$ $(0.0152)$ 
& $-0.5255$ $(0.1462)$ 
% & $-0.8660$ $(0.3333)$ 
& $-0.8476^{**}$ $(0.0039)$ 
& $-0.2605$ $(0.4984)$ \\
& Loc.      
& $ 0.3729$ $(0.3229)$ 
& $-0.9662^{***}$ $(<0.001)$ 
% & $-0.8660$ $(0.3333)$ 
& $-0.8946^{**}$ $(0.0011)$ 
& $-0.3811$ $(0.3115)$ \\
& Port.
& $ 0.4167$ $(0.2646)$ 
& $-0.9789^{***}$ $(<0.001)$ 
% & $-0.8660$ $(0.3333)$ 
& $-0.8791^{**}$ $(0.0018)$ 
& $-0.7731^{*}$ $(0.0146)$ \\
\bottomrule
\end{tabular}

% \vspace{0.4em}
% \raggedright
% \footnotesize
% \textit{Note.} 
% $^{*}p<0.05$, $^{**}p<0.01$, $^{***}p<0.001$.
\end{table*}

% \subsection{Generalization Across Model Scales and Architectures}
% \label{sec:Generalization Across Model Scales and Architectures}

% The main text focuses on 7B--14B decoder-only LLMs, but the geometric signatures underlying our analysis extend beyond a single model. The dimensional-collapse statistics (Tables~\ref{tab:dimensional_collapse_appendix_llama31}--\ref{tab:dimensional_collapse_t5_mend}), perturbation norms (Tables~\ref{tab:perturbation_norms}--\ref{tab:perturbation_t5_mend}), PC-stability analyses (Tables~\ref{tab:pc_stability}--\ref{tab:pc_stability_t5_output}), and \(R_k\) distributions (Tables~\ref{tab:relative_change_rates}--\ref{tab:rk_distribution_t5}) consistently reveal evidence of the same vulnerability mechanism across DeepSeek-R1-Distill-LLaMA-8B, LLaMA-2-13B, and specific components of T5-3B.

% However, the encoder-decoder architecture exhibits greater heterogeneity, with certain layers departing from the small-perturbation regime more rapidly. As a result, we do not assert identical quantitative patterns across all architectures. Instead, we posit that the amplification mechanism arising from dimensional collapse is broadly applicable. The severity of this mechanism depends on factors including architecture type, layer characteristics, editing method, and editing horizon.

% ===============================================================
\section{Experiments and Results}
\label{sec:results}

\subsection{Full Experimental Settings for the Benchmark Study}
\label{sec:Full Experimental Settings}

This section provides the complete experimental configuration for the benchmark evaluation in Section~\ref{sec:experiments}, complementing the concise description in Section~\ref{sec:experimental_settings}.

\textbf{LLMs.} We primarily conduct experiments on decoder-only LLMs, including general-purpose LLMs like Llama-2-7B-Chat \citep{touvron2023llama}, Llama-3.1-8B-Instruct \citep{llama3modelcard}, and Mistral-7B-Instruct \citep{jiang2023mistral}. To examine the impact of knowledge editing methods on reasoning-oriented models, we additionally evaluate DeepSeek-R1-Distill-LLaMA-8B \citep{guo2025deepseek}. We also include larger LLMs, Llama-2-13B and Qwen3-14B, for scale comparison.

\textbf{Knowledge Editing Methods.} This study examines recent knowledge editing methods, which can be broadly categorized based on whether they modify a model's internal parameters. Parameter-based editing methods include locate-then-edit methods such as ROME \citep{meng2022locating}, MEMIT \citep{mengmass}, PMET \citep{li2024pmet}, RECT \citep{gu2024model}, AlphaEdit \citep{fang2024alphaedit}, and FT-L \citep{meng2022locating}, as well as meta-learning-based methods such as MEND \citep{de2021editing}. Additional-parameter methods include AdaLoRA \citep{zhang2023adalora} and WISE \citep{wang2024wise}. In contrast, external memory-based methods include GRACE \citep{hartvigsen2024aging} and a simple, RAG baseline, Selective Contextual Reasoning (SCR) \citep{he2025knowledge}.

\textbf{Knowledge Editing Datasets.} We use two widely adopted context-free question answering (QA) datasets in knowledge editing research: WikiData$_\text{counterfact}$ \citep{cohen2024evaluating} and Zero-Shot Relation Extraction (ZsRE) \citep{levy2017zero}, both of which serve as counterfactual benchmarks for modifying knowledge in LLMs. In addition, we incorporate an event-centric dataset, ELKEN \citep{peng2024event}, in which each instance contains multiple related statements, to evaluate the ability of knowledge editing methods to handle more abundant and complex knowledge.

\textbf{General Datasets.} To evaluate the ability of edited LLMs to handle general tasks, we select mathematical reasoning benchmarks (including AIME 2024 \citep{aime24}, AIME 2025 \citep{aime25}, AMC \citep{li2024numinamath}, OlympiadBench \citep{he2024olympiadbench}, and MATH-500 \citep{lightman2024let}) and broader downstream tasks (including GPQA-Diamond \citep{rein2023gpqa}, ARC$_\text{c}$ \citep{allenai:arc}, and MMLU-Pro \citep{wang2024mmlu}) to assess the extent to which their reasoning capabilities are preserved after editing.

\textbf{Inference Setting of Knowledge Editing.} Following methods such as MEND \citep{mitchell2021fast} and ROME \citep{melocang2022locating}, several follow-up studies \citep{zhang2024comprehensive,wang2024wise} have adopted teacher forcing \citep{williams1989learning} during inference and evaluated token-level changes. While this evaluation setting is widely used, it relies on ground-truth answer sequences at inference time, which introduces potential data leakage. Such setups may lead to overly optimistic results and fail to accurately reflect the method's effectiveness in real-world generative scenarios, as also highlighted by \citet{yang2025mirage}. To ensure a more fair and realistic evaluation, we adopt a unified autoregressive generation paradigm \citep{mccoy2023embers} for prediction. We directly evaluate whether the generated answer is correct with respect to the target answer.

\textbf{Sequential Editing Scenario.} Since single editing is less representative of real-world applications, we conduct experiments under a sequential editing scenario. For each dataset, we edit each knowledge item in sequence, treating the final LLM as the fully edited version.

\textbf{Event Knowledge Editing Settings.} For parameter-based knowledge editing methods, we use GPT-4o \citep{hurst2024gpt} to convert each event into a set of fact triples, which are then used for sequential editing. The prompt used is given in Appendix~\ref{sec:ADataset}. For external memory-based methods, the original event text is directly provided as context input or integrated into the editable memory.

\textbf{Implementation Environment.} The main experiments are executed on 8 NVIDIA A800 GPUs under a Linux system. All methods are implemented using EasyEdit.\footnote{\url{https://github.com/zjunlp/EasyEdit.git}} Notably, AlphaEdit fails to work with Llama2 and Mistral. We also tested with its original implementation. The results are consistent with those obtained using the EasyEdit version. See Appendix~\ref{sec:AlphaEdit Hyperparameter Sensitivity Analysis} for further details. Evaluation on reasoning benchmarks is conducted using LUFFY.\footnote{\url{https://github.com/ElliottYan/LUFFY.git}}

\textbf{Metrics Calculation.} For the instruct LLM, we limit the max output token length to 50, while for the reasoning LLM, it is set to 1024. For the four evaluation dimensions, we use Qwen2.5-72B-Instruct to assess whether the answers generated by the edited LLMs are semantically consistent with the ground-truth answers, and compute their accuracy (\%) accordingly.
We validate this evaluation protocol against human judgments on a subset of examples. Manual verification on 250 randomly sampled cases yields 96\% agreement with human annotations, supporting the reliability of the automatic evaluation.

To evaluate the impact of different editing loads on the performance of knowledge editing, we test scenarios involving 1, 10, and 100 edits. Specifically, we select the first 100 knowledge items from the dataset. For the 1-edit setting, each item is edited individually, and the average performance across all 100 edits is reported. For the 10-edit setting, the 100 items are evenly divided into 10 groups, with sequential editing applied within each group; the average performance after each 10-edit sequence is used. For the 100-edit setting, we sequentially edit all 100 items. Regarding locality, we additionally conduct a token-level validation, which yields results consistent with those at the semantic level (see Appendix~\ref{sec:Token-Level Locality Evaluation}).

% ===============================================================
\subsection{Dataset}
\label{sec:ADataset}
% ===============================================================

\begin{itemize}
\item \textbf{ZsRE}: Originally a question-answering dataset, ZsRE is extended by \cite{yao2023editing} to assess various dimensions of model editing methods. 

\item \textbf{WikiData\(_\text{counterfact}\)}: This dataset collects triplets about popular entities, ensuring that the subject corresponds to one of the most viewed Wikipedia pages. 

\item \textbf{ELKEN}: This dataset is designed for event-level knowledge editing, focusing on directly editing new events into LLMs. It includes a diverse set of events and corresponding questions about factual knowledge and future trends.  Due to the fact that tendency predictions are influenced by factors beyond the current edited event, we have filtered out questions related to future trends, focusing exclusively on factual knowledge. 
\end{itemize}

The first two datasets, \textbf{ZsRE} and \textbf{WikiData\(_\text{counterfact}\)}, are from KnowEdit\footnote{\url{https://huggingface.co/datasets/zjunlp/KnowEdit}} and released under the \textbf{MIT License}. The \textbf{ELKEN} dataset\footnote{\url{https://github.com/THU-KEG/Event-Level-Knowledge-Editing.git}} follows the \textbf{CC BY-NC-SA 4.0} license.

\clearpage

\newpage

\begin{tcolorbox}[colback=white, colframe=black, title=Prompt for Extracting Triples from Event, width=\textwidth]

Please extract one or more (subject, relation, object) triples from the following event.\\
Each triple must express a complete, standalone fact, with no redundancy or dependency on other triples.\\[6pt]

\textbf{Instructions:}\\
If the event contains multiple independent facts, extract \textbf{multiple triples}, one per fact.\\
The \textbf{Subject} must be the main entity or actor involved in the fact.\\
The \textbf{Relation} should be a clear \textbf{predicate phrase} (verb or action-based phrase) that uniquely points to the \textbf{Object}.\\
The \textbf{Object} should contain the new or important information not repeated in the subject or relation.\\
Avoid vague or generic relations like ``is involved in'', ``is associated with''.\\
Do not merge multiple facts into a single triple.\\
If there are multiple pieces of information (\textit{e.g.}, place, time, role), consider extracting \textbf{multiple triples}.\\[6pt]

\textbf{Format each triple as follows:}\\
\textbf{Subject}: \\
Prompt: (Concatenate subject and relation into a natural language phrase) \\
Target New: (Only the object, ideally as short as possible)\\[6pt]

\textbf{Example1:}\\
Event: Serena Williams announces her retirement from professional tennis.\\
Output: \\
Subject: Serena Williams \\
Prompt: Serena Williams retires from \\
Target New: professional tennis\\[6pt]

\textbf{Example2:}\\
Event: Pete Townshend is pursuing a degree in philosophy at the Royal College of Art. \\
Output: \\
Subject: Pete Townshend \\
Prompt: Pete Townshend is pursuing the degree of \\
Target New: philosophy \\
Subject: Pete Townshend \\
Prompt: Pete Townshend is studying at \\
Target New: the Royal College of Art\\[6pt]

\textbf{Example3:}\\
Event: Paul Wight founded NexGen Technologies in Bergen, appointing Martin Allen as CEO. \\
Output: \\
Subject: Paul Wight \\
Prompt: Paul Wight founded \\
Target New: NexGen Technologies \\
Subject: NexGen Technologies \\
Prompt: The CEO of NexGen Technologies is \\
Target New: Martin Allen\\[6pt]

Event: \{event\} \\
Output:
\end{tcolorbox}

% ===============================================================
\subsection{Additional Results}
\label{sec:AdditionResult}
% ===============================================================

\begin{table*}[!h]
  \centering
  \caption{Performance comparison of knowledge editing methods on general LLMs using the ZsRE dataset. The best results are shown in \textbf{bold}, and second best results are \underline{underlined}. `–' denotes that the experiment could not be completed.} % Please refer to Footnote 2 for further details.
    % \caption{Performance comparison of knowledge editing on general LLMs on ZsRE : \texttt{Llama-2-7B-Chat}, \texttt{Llama-3.1-8B-Instruct}, and \texttt{Mistral-7B-Instruct-v0.1}. Best results are highlighted in bold.} % `-' indicates the model fails to produce any results.

  \scriptsize
  \resizebox{\linewidth}{!}{  
  \begin{tabular}{c|*{5}{r}|*{5}{r}|*{5}{r}}
    \toprule
    Model & \multicolumn{5}{c|}{\textbf{Llama-2-7B-Chat}} & \multicolumn{5}{c|}{\textbf{Llama-3.1-8B-Instruct}} & \multicolumn{5}{c}{\textbf{Mistral-7B-Instruct-v0.1}} \\
    \cmidrule(r){2-6} \cmidrule(r){7-11} \cmidrule(r){12-16}
    Metric & Rel. & Gen. & Loc. & Port. & Avg. & Rel. & Gen. & Loc. & Port. & Avg. & Rel. & Gen. & Loc. & Port. & Avg. \\
    \midrule
    \textit{Pre-edit} & 2.54 & 2.31 & \underline{11.49} & \underline{4.50} & 5.21 & 2.70 & 2.40 & \underline{14.26} & 3.68 & 5.76 & 3.77 & 3.07 & \underline{13.30} & \underline{5.35} & 6.37 \\
    \midrule
    ROME & 0.61 & 0.46 & 0.00 & 0.00 & 0.27 & 0.00 & 0.00 & 0.00 & 0.00 & 0.00 & 0.23 & 0.08 & 0.04 & 0.00 & 0.09 \\
    MEMIT & 0.00 & 0.00 & 0.00 & 0.00 & 0.00 & 0.00 & 0.00 & 0.00 & 0.00 & 0.00 & 0.00 & 0.00 & 0.00 & 0.00 & 0.00 \\
    PMET & 0.00 & 0.00 & 0.00 & 0.00 & 0.00 & 6.46 & 6.30 & 7.80 & 4.79 & 6.34 & 0.00 & 0.00 & 0.00 & 0.00 & 0.00 \\
    RECT & 0.15 & 0.31 & 0.00 & 0.00 & 0.12 & 5.76 & 4.77 & 0.00 & 0.85 & 2.84 & 0.54 & 0.77 & 0.00 & 0.38 & 0.42 \\
    AlphaEdit & - & - & - & - & - & \underline{69.49} & \underline{55.50} & 8.38 & \underline{8.56} & \underline{35.48} & - & - & - & - & - \\
    FT-L & 0.23 & 0.23 & 0.15 & 0.00 & 0.15 & 0.00 & 0.00 & 0.00 & 0.00 & 0.00 & 5.15 & \underline{4.46} & 10.65 & 4.16 & 6.10 \\
    MEND & 0.00 & 0.00 & 0.00 & 0.00 & 0.00 & 0.00 & 0.00 & 0.00 & 0.00 & 0.00 & 0.00 & 0.00 & 0.00 & 0.00 & 0.00 \\
    AdaLoRA & 0.00 & 0.00 & 0.00 & 0.00 & 0.00 & 0.00 & 0.00 & 0.00 & 0.00 & 0.00 & 0.00 & 0.00 & 0.00 & 0.00 & 0.00 \\
    WISE & 8.22 & \underline{6.69} & 8.69 & 2.58 & 6.55 & 2.84 & 2.46 & 10.30 & 2.35 & 4.49 & 2.00 & 1.38 & 1.23 & 0.90 & 1.38 \\
    % IKE & \textbf{91.62} & \textbf{93.16} & 1.87 & \underline{45.44} & \textbf{58.02} & \textbf{95.85} & \textbf{97.69} & 4.73 & \underline{52.38} & \textbf{62.66} & 71.33 & 77.71 & 0.73 & 24.84 & 43.65 \\
    % ICE & 79.78 & 74.40 & \textbf{23.56} & \textbf{49.56} & \underline{56.83} & 74.79 & 72.79 & \textbf{27.94} & \textbf{53.78} & \underline{57.33} & \underline{86.86} & \textbf{84.09} & \textbf{23.06} & \textbf{52.46} & \textbf{61.62} \\
    GRACE & \underline{48.96} & 0.38 & 9.19 & 0.00 & \underline{14.63} & 60.34 & 2.69 & 10.91 & 3.33 & 19.32 & \underline{60.80} & 0.23 & 12.72 & 0.00 & \underline{18.44} \\
    \midrule
    % SCR & \underline{80.71} & \underline{73.25} & \underline{16.18} & 39.99 & 52.53 & \underline{84.40} & \underline{75.56} & \underline{16.03} & 46.41 & 55.60 & \textbf{88.39} & \underline{78.94} & \underline{16.14} & \underline{40.98} & \underline{56.11} \\
    SCR & \textbf{80.71} & \textbf{73.25} & \textbf{16.18} & \textbf{39.99} & \textbf{52.53} & \textbf{84.40} & \textbf{75.56} & \textbf{16.03} & \textbf{46.41} & \textbf{55.60} & \textbf{88.39} & \textbf{78.94} & \textbf{16.14} & \textbf{40.98} & \textbf{56.11} \\
    \bottomrule
  \end{tabular}
  }
  \label{tab:sequential_performance_zsre}
  \vskip -0.1in
\end{table*}

\begin{table*}[!h]
  \centering
  \caption{Performance comparison of knowledge editing methods on general LLMs using the WikiData$_\text{counterfact}$ dataset. The best results are shown in \textbf{bold}, and second best results are \underline{underlined}. `--' denotes that the experiment could not be completed.}

  \scriptsize
  \resizebox{\linewidth}{!}{%
  \begin{tabular}{c|*{5}{r}|*{5}{r}|*{5}{r}}
    \toprule
    Model & \multicolumn{5}{c|}{\textbf{Llama-2-7B-Chat}} & \multicolumn{5}{c|}{\textbf{Llama-3.1-8B-Instruct}} & \multicolumn{5}{c}{\textbf{Mistral-7B-Instruct-v0.1}} \\
    \cmidrule(r){2-6} \cmidrule(r){7-11} \cmidrule(r){12-16}
    Metric & Rel. & Gen. & Loc. & Port. & Avg. & Rel. & Gen. & Loc. & Port. & Avg. & Rel. & Gen. & Loc. & Port. & Avg. \\
    \midrule
    \textit{Pre-edit} & 0.24 & 0.12 & \textbf{29.68} & 2.04 & 8.02 & 0.24 & 0.36 & \textbf{30.19} & 3.86 & 8.66 & 0.36 & 0.36 & \textbf{31.87} & \underline{5.15} & 9.43 \\
    \midrule
    ROME        & 0.00 & 0.00 & 0.00 & 0.00 & 0.00 & 0.00 & 0.00 & 0.00 & 0.00 & 0.00 & 0.00 & 0.00 & 0.00 & 0.00 & 0.00 \\
    MEMIT       & 0.00 & 0.00 & 0.00 & 0.00 & 0.00 & 0.00   & 0.00   & 0.00   & 0.00   & 0.00   & 0.00 & 0.00 & 0.00 & 0.00 & 0.00 \\
    PMET        & 0.00 & 0.00 & 0.00 & 0.00 & 0.00 & 0.24   & 0.36   & 8.17   & 2.34   & 2.78   & 0.00 & 0.00 & 0.00 & 0.00 & 0.00 \\
    RECT        & 0.00 & 0.00 & 0.00 & 0.00 & 0.00 & 1.55 & 1.91 & 0.00 & 0.10 & 0.89 & 0.48 & 0.00 & 0.00 & 0.03 & 0.13 \\
    AlphaEdit   & --   & --   & --   & --   & --   & 30.51 & \underline{25.63} & 1.33 & \underline{8.09} & 16.39 & -- & -- & -- & -- & -- \\
    FT-L        & 0.00 & 0.00 & 0.08 & 0.01 & 0.02 & 0.00 & 0.12 & 0.00 & 0.00 & 0.03 & 0.95 & 0.83 & 1.37 & 0.78 & 0.98 \\
    MEND        & 0.00 & 0.00 & 0.00 & 0.00 & 0.00 & 0.00 & 0.00 & 0.00 & 0.00 & 0.00 & 0.00 & 0.00 & 0.00 & 0.00 & 0.00 \\
    AdaLoRA     & 0.00 & 0.00 & 0.00 & 0.00 & 0.00 & 0.00 & 0.00 & 0.00 & 0.00 & 0.00 & 0.00 & 0.00 & 0.00 & 0.00 & 0.00 \\
    WISE        & 21.33 & \underline{13.35} & 14.18 & \underline{5.02} & 13.47 & 0.36 & 0.47 & \underline{29.36} & 3.06 & 8.31 & 4.41 & \underline{5.96} & 5.21 & 1.94 & 4.38 \\
    GRACE       & \underline{48.39} & 0.00 & \underline{24.69} & 1.34 & \underline{18.60} & \underline{40.88} & 0.36 & 28.53 & 2.70 & \underline{18.12} & \underline{52.80} & 0.24 & \underline{31.65} & 1.98 & \underline{21.67} \\
    \midrule
    SCR         & \textbf{76.64} & \textbf{73.54} & 11.63 & \textbf{28.75} & \textbf{47.64} & \textbf{88.20} & \textbf{87.01} & 28.43 & \textbf{32.90} & \textbf{59.14} & \textbf{85.10} & \textbf{76.04} & 22.21 & \textbf{29.92} & \textbf{53.32} \\
    \bottomrule
  \end{tabular}%
  }
  \label{tab:sequential_performance_counterfact}
  \vskip -0.1in
\end{table*}

\begin{table*}
  \centering
  \caption{Performance comparison of single and sequential editing on Llama2-13B and Qwen3-14B using the ZsRE dataset.
  The results demonstrate consistency with the insights observed in smaller LLMs. In practice-oriented settings, all parameter modification-based methods fall significantly short of the near-perfect single-edit performance. However, 
  retrieval-based methods maintain robustness in both single and sequential editing scenarios.}
  % \tiny
  \small
  \resizebox{0.8\linewidth}{!}{
  \begin{tabular}{c|rrrrr|rrrrr}
    \toprule
    \multicolumn{11}{c}{\textbf{Llama2-13B}} \\
    \midrule
    \textbf{Method} & \multicolumn{5}{c|}{\textbf{Single Editing}} & \multicolumn{5}{c}{\textbf{Sequential Editing}} \\
    \midrule
    Metric & Rel. & Gen. & Loc. & Port. & Avg. & Rel. & Gen. & Loc. & Port. & Avg. \\
    \midrule
    \textit{Pre-edit} & 3.00 & 3.61 & \underline{12.64} & 4.73 & 6.00 & 3.00 & 3.61 & \underline{12.64} & 4.73 & 6.00 \\
    \midrule
    ROME      & 57.41 & 51.88 & 5.61  & 7.66  & 30.64 & 0.46  & 0.23  & 0.00  & 0.63  & 0.33 \\
    MEMIT     & 32.59 & 27.44 & 2.96  & 6.91  & 17.48 & 0.00  & 0.00  & 0.00  & 0.00  & 0.00 \\
    PMET      & 13.68 & 10.91 & 6.61  & 6.38  & 9.40  & 4.77  & 4.38  & 6.31  & \underline{5.41}  & 5.22 \\
    RECT      & \underline{59.42} & 50.12 & 6.03  & 7.84  & 30.85 & 1.15  & 0.92  & 0.04  & 1.77  & 0.97 \\
    AlphaEdit & 37.66 & 29.82 & 7.07  & 8.23  & 20.70 & 28.05 & \underline{22.91} & 2.42  & 4.20  & 14.40 \\
    FT-L      & 4.07  & 8.61  & 5.69  & 4.99  & 5.84  & 0.00  & 0.08  & 0.00  & 0.00  & 0.02 \\
    MEND      & 1.84  & 1.69  & 12.45 & 3.91  & 4.97  & 0.00  & 0.00  & 0.00  & 0.00  & 0.00 \\
    AdaLoRA   & 58.80 & \underline{52.50} & 11.45 & \underline{10.66} & \underline{33.35} & 0.00 & 0.00 & 0.00 & 0.00 & 0.00 \\
    WISE      & 31.44 & 30.28 & 8.15  & 8.85  & 19.68 & 15.07 & 11.07 & 11.80 & 3.60  & 10.39 \\
    GRACE     & 47.81 & 3.61  & \underline{12.64} & 4.73  & 17.20 & \underline{47.35} & 3.77 & \underline{12.64} & 4.73  & \underline{17.12} \\
    \midrule
    SCR       & \textbf{70.41} & \textbf{70.10} & \textbf{18.52} & \textbf{47.80} & \textbf{51.71} & \textbf{68.79} & \textbf{61.03} & \textbf{17.45} & \textbf{38.69} & \textbf{46.49} \\
    \midrule
    \multicolumn{11}{c}{\textbf{Qwen3-14B}} \\
    \midrule
    \textbf{Method} & \multicolumn{5}{c|}{\textbf{Single Editing}} & \multicolumn{5}{c}{\textbf{Sequential Editing}} \\
    \midrule
    Metric & Rel. & Gen. & Loc. & Port. & Avg. & Rel. & Gen. & Loc. & Port. & Avg. \\
    \midrule
    \textit{Pre-edit} & 2.15 & 1.69 & 11.72 & 1.79 & 4.34 & 2.15 & 1.69 & \underline{11.72} & 1.79 & 4.34 \\
    \midrule
    ROME      & \underline{68.49} & \underline{45.35} & 6.03  & 5.49  & \underline{31.34} & 0.08  & 0.00  & 0.00  & 0.00  & 0.02 \\
    RECT      & \textbf{68.72} & 44.81 & 6.11  & 5.55  & 31.30 & 0.00  & 0.00  & 0.00  & 0.00  & 0.00 \\
    FT-L      & 18.75 & 20.29 & 5.61  & \underline{7.18}  & 12.96 & 0.00  & 0.08  & 0.00  & 0.00  & 0.02 \\
    MEND      & 1.69  & 1.69  & 10.88 & 1.52  & 3.95  & 0.00  & 0.00  & 0.00  & 0.00  & 0.00 \\
    AdaLoRA   & 22.83 & 7.15  & \textbf{12.03} & 3.12  & 11.28 & 0.23  & 0.15  & 1.31  & 0.21  & 0.48 \\
    WISE      & 5.38  & 4.07  & 10.11 & 1.88  & 5.36  & 2.77  & \underline{2.46}  & 10.53 & \underline{2.17}  & 4.48 \\
    GRACE     & 58.34 & 1.69  & \underline{11.84} & 1.79  & 18.42 & \underline{58.80} & 2.38 & \textbf{11.84} & 1.79 & \underline{18.70} \\
    \midrule
    SCR       & 49.27 & \textbf{58.42} & 11.59 & \textbf{34.69} & \textbf{38.49} & \textbf{85.40} & \textbf{74.56} & 11.53 & \textbf{29.45} & \textbf{50.24} \\
    \bottomrule
  \end{tabular}%
  }
  \label{tab:add_results_largerLLM}%
\end{table*}

\begin{figure}[htb]%[!htbp]
\vskip 0.2in
    \centering
    \includegraphics[width=\textwidth]{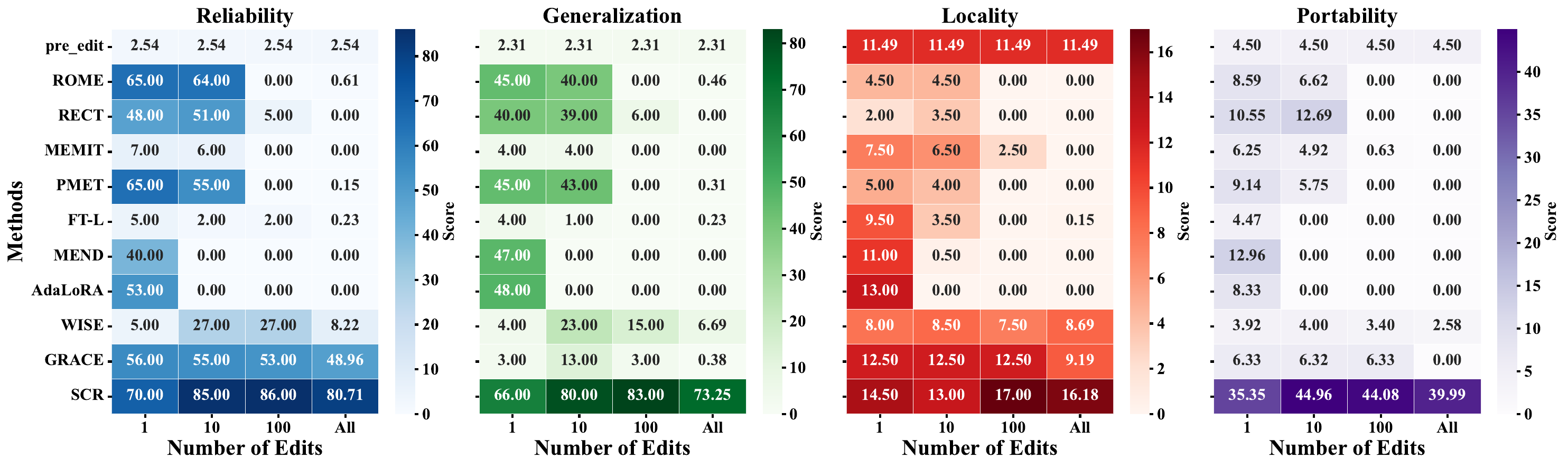}
    \caption{Performance changes of knowledge editing methods during sequential editing of Llama-2-7B-Chat on the ZsRE dataset. The x-axis represents the number of edits: 1, 10, 100, and the full dataset.}
    \label{fig:scale_llama2}
\end{figure}

\begin{figure}[!htbp]%[h]
\vskip 0.2in
    \centering
    \includegraphics[width=\textwidth]{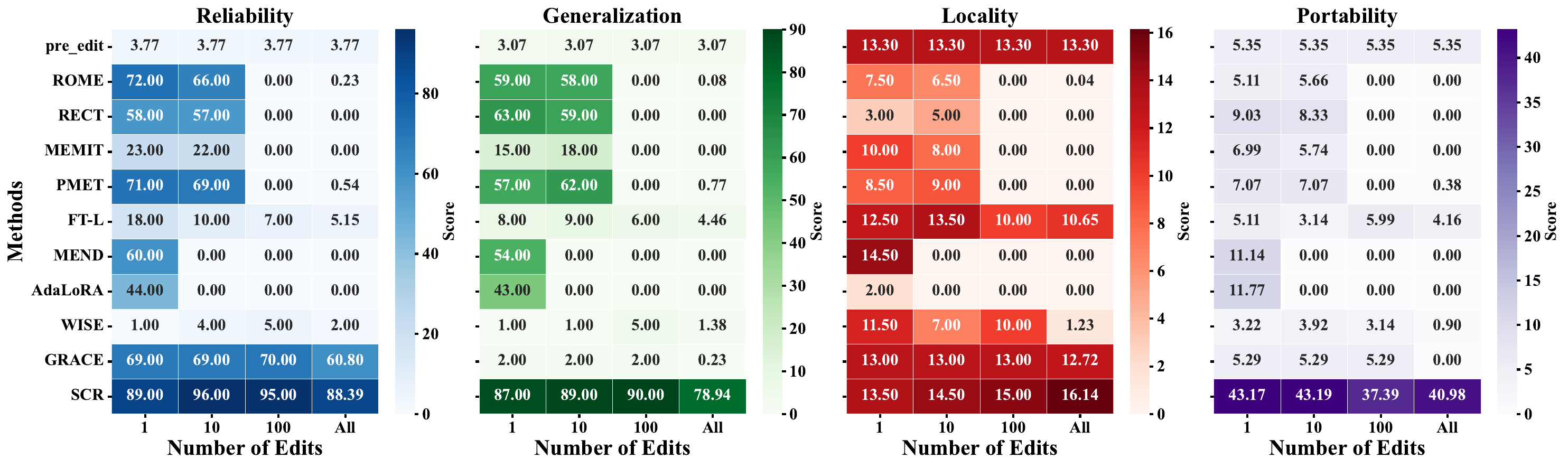}
    \caption{Performance changes of knowledge editing methods during sequential editing of Mistral-7B-Instruct-v0.1 on the ZsRE dataset. The x-axis represents the number of edits: 1, 10, 100, and the full dataset.}
    \label{fig:scale_mistral}
\end{figure}

\begin{table}[h]
\centering
\caption{Performance comparison on the WikiData\(_\text{counterfact}\) dataset using DeepSeek-R1-Distill-Llama-8B as the backbone. We report four metrics (Reliability, Generalization, Locality, Portability) and their average for both Single and Sequential Editing scenarios.}
\label{tab:llama_disitll_wiki_counterfact_comparison}
% \tiny  % 使用与表3相同的字体大小
\small  % 使用与表3相同的字体大小
\resizebox{0.8\linewidth}{!}{
\renewcommand{\arraystretch}{1.2}
\setlength{\tabcolsep}{5pt}
\begin{tabular}{l|rrrrr|rrrrr}
\toprule
\multirow{2}{*}{\textbf{Method}} & \multicolumn{5}{c|}{\textbf{Single Editing}} & \multicolumn{5}{c}{\textbf{Sequential Editing}} \\
\cmidrule(lr){2-6} \cmidrule(lr){7-11}
& \makecell[c]{Rel.} & \makecell[c]{Gen.} & \makecell[c]{Loc.} & \makecell[c]{Port.} & \makecell[c]{Avg.} 
& \makecell[c]{Rel.} & \makecell[c]{Gen.} & \makecell[c]{Loc.} & \makecell[c]{Port.} & \makecell[c]{Avg.} \\
\midrule
\textit{Pre-edit} & 0.00 & 0.00 & \textbf{23.55} & 3.85 & 6.85 & 0.00 & 0.00 & \textbf{23.55} & 3.85 & 6.85 \\
\midrule
ROME     & 30.00 & 18.00 & 6.22  & 9.23  & 15.86 & 0.00 & 0.00 & 0.00  & 0.00  & 0.00 \\
RECT     & 26.00 & 16.00 & 8.80  & 9.14  & 14.98 & 0.00 & 0.00 & 0.00  & 0.03  & 0.01 \\
AlphaEdit& 31.00 & 5.00  & 15.19 & 5.53  & 14.18 & \underline{33.00} & \underline{14.00} & 8.37  & \underline{9.71}  & \underline{16.27} \\
FT       & 2.00  & 1.00  & 0.82  & 4.45  & 2.07  & 0.00 & 0.00 & 0.00  & 0.00  & 0.00 \\
MEND     & \underline{32.00} & 14.00 & 19.88 & \underline{13.35} & \underline{19.81} & 0.00 & 0.00 & 0.00  & 0.00  & 0.00 \\
AdaLoRA  & 22.00 & \underline{19.00} & 19.88 & \underline{13.35} & 18.56 & 0.00 & 0.00 & 0.00  & 0.00  & 0.00 \\
WISE     & 8.00  & 7.00  & 3.00  & 2.59  & 5.15  & 0.00 & 0.00 & 6.00  & 2.86  & 2.22 \\
GRACE    & 12.00 & 0.00  & \underline{23.54} & 4.96  & 10.13 & 12.00 & 0.00  & \underline{23.49} & 4.96  & 10.11 \\
\midrule
SCR      & \textbf{72.00} & \textbf{67.00} & 23.50 & \textbf{34.83} & \textbf{49.33} & \textbf{84.00} & \textbf{76.00} & 15.70 & \textbf{35.67} & \textbf{52.84} \\
\bottomrule
\end{tabular}
}
\vskip -0.1in
\end{table}

\begin{table*}%[h]
  \centering
  \caption{Performance comparison of edited DeepSeek-R1-Distill-Llama-8B on mathematical benchmarks requiring reasoning, reporting accuracy after 10 and 100 edits.}
    % \small
    % \resizebox{\linewidth}{!}{ 
    % \begin{tabular}{c|c|ccccc|c} 
  \tiny  % 使用与表3相同的字体大小
  \resizebox{0.8\linewidth}{!}{ % 使用与表3相同的缩放比例
    \begin{tabular}{c|c|ccccc|c} 
    \toprule
    Method & \#Editing & AIME 2024 & AIME 2025 & AMC   & MATH-500 & Olympiad & Avg. \\
    \midrule
     \textit{Pre-edit} & 0     & 36.35  & 25.63  & 68.67  & 83.20  & 53.48  & 53.47  \\
    \midrule
    \multirow{2}[2]{*}{ROME} & 10    & 31.04  & 24.79  & 66.57  & 82.20  & 49.04  & 50.73  \\
          & 100   & 0.00  & 0.00  & 0.00  & 0.00  & 0.00  & 0.00  \\
    \midrule
    \multirow{2}[2]{*}{RECT} & 10    & 33.02  & 24.58  & 33.02  & 83.60  & 51.26  & 45.10  \\
          & 100   & 0.00  & 0.10  & 0.04  & 0.00  & 0.00  & 0.03  \\
    \midrule
    \multirow{2}[2]{*}{AlphaEdit} & 10    & 36.56  & 26.56  & 69.43  & 85.20  & 51.85  & 53.92  \\
          & 100   & 35.31  & 25.94  & 69.47  & 81.40  & 53.48  & 53.12  \\
    \midrule
    \multirow{2}[2]{*}{MEND} & 10    & 0.00  & 0.00  & 0.00  & 0.00  & 0.00  & 0.00  \\
          & 100   & 0.00  & 0.00  & 0.00  & 0.00  & 0.00  & 0.00  \\
    \bottomrule
    \end{tabular}%
   }
  \label{tab:reasoning_ability_post_editing}%
\end{table*}%

\begin{table}[htbp]
\centering
\caption{
% Impact of Sequential Knowledge Editing (10 and 100 edits) on Broader Downstream Tasks. 
% To assess the broader side effects of knowledge editing, we evaluate the accuracy of post-edit models (DeepSeek-R1-Distill-Llama-8B) on three challenging benchmarks: GPQA-Diamond (science graduate-level knowledge), ARC-Challenge (open-domain reasoning), and MMLU-Pro (reasoning-based academic questions). Results show that parameter-modification methods (ROME, RECT) and MEND suffer catastrophic performance collapse after extensive editing, while AlphaEdit demonstrates relative resilience, maintaining near-original performance even after 100 edits, albeit with limitations in targeted editing efficacy.
Performance comparison of edited DeepSeek-R1-Distill-Llama-8B on broader reasoning-based downstream tasks, reporting accuracy after 10 and 100 edits.
}
\label{tab:Broader-Downstream-Tasks}
\tiny
\resizebox{0.6\linewidth}{!}{  % 调整为更合适的宽度
\begin{tabular}{l|c|ccc|c}
\toprule
Method & \#Editing & GPQA-Diamond & ARC\(_\text{c}\) & MMLU-Pro & Avg. \\
\midrule
\textit{Pre-edit} & 0 & 28.79 & 87.80 & 46.13 & 54.24\\
\midrule  % 添加横线分隔 pre-edit 和其他方法
\multirow{2}{*}{ROME} & 10 & 18.69 & 85.15 & 43.30 & 49.05\\
     & 100 & 0.00 & 0.00 & 0.00 & 0.00\\
\midrule 
\multirow{2}{*}{AlphaEdit} & 10 & 25.76 & 87.97 & 45.51 & 53.08\\  
                           & 100 & 24.75 & 88.05 & 45.84 & 52.88\\
\midrule 
\multirow{2}{*}{RECT} & 10 & 20.20 & 86.26 & 44.66 & 50.37\\
     & 100 & 0.00 & 0.00 & 0.00 & 0.00\\
% \addlinespace[0.1em]
\midrule 
% \addlinespace[0.1em]
\multirow{2}{*}{MEND} & 10 & 0.00 & 0.00 & 0.00 & 0.00\\
     & 100 & 0.00 & 0.00 & 0.00 & 0.00\\
\bottomrule
\end{tabular}
}
\vskip -0.1in
\end{table}

\begin{table}
  \centering
  \caption{Performance comparison of \textbf{Single Editing} across Llama-2-7B-Chat, Llama-3.1-8B-Instruct, and DeepSeek-R1-Distill-Llama-8B based on the event-level dataset ELKEN. `–' denotes that the experiment could not be completed. Please refer to Footnote 1 for further details.}
    % \caption{Comparison of \textbf{Single Editing} performance on \textbf{Event-level Dataset ELKEN}. Backbone LLMs include \texttt{Llama-2-7B-Chat}, \texttt{Llama-3.1-8B-Instruct}, and \texttt{DeepSeek-R1-Distill-Llama-8B}.} 
  %   \small
  % \resizebox{\linewidth}{!}{  
    %\tiny  % 使用与表3相同的字体大小
  \resizebox{0.9\linewidth}{!}{ % 使用与表3相同的缩放比例
    \begin{tabular}{c|ccc|ccc|ccc}
    \toprule
    \multirow{2}[4]{*}{\textbf{Single}} & \multicolumn{3}{c|}{Llama-2-7B-Chat} & \multicolumn{3}{c|}{Llama-3.1-8B-Instruct} & \multicolumn{3}{c}{DeepSeek-R1-Distill-Llama-8B} \\
\cmidrule{2-10}          & Port. & Loc. & Avg.  & Port. & Loc. & Avg.  & Port. & Loc. & Avg. \\
    \midrule
    \textit{Pre-edit} & 5.92  & \textbf{43.75}  & \underline{24.84}  & 7.08  & \textbf{56.73}  & \underline{31.91}  & 9.01  & \textbf{48.73}  & \underline{28.87}  \\
    \midrule
    ROME  & 2.41  & 19.45  & 10.93  & 17.38  & 23.39  & 20.39  & 12.91 & 22.21  & 17.56  \\
    % MEMIT & 3.03  & 14.34  & 8.69  & -     & -     & -     & -     & -     & - \\
    % PMET  & 2.41  & 31.67  & 17.04  & -     & -     & -     & -     & -     & - \\
    RECT  & \underline{7.69}  & 19.83  & 13.76  & \underline{18.45}  & 26.55  & 22.50  & \underline{14.70}  & 23.29  & 19.00  \\
    AlphaEdit & -     & -     & -     & 14.60  & 40.61  & 27.61  & 14.25  & 39.76  & 27.01  \\
    FT-L  & 2.33  & 18.19  & 10.26  & 4.79  & 22.86  & 13.83  & 5.53  & 12.90  & 9.22  \\
    AdaLoRA & 2.07  & 15.21  & 8.64  & 10.18  & 17.18  & 13.68  & 8.32  & 13.62  & 10.97  \\
    WISE  & 2.49  & 25.77  & 14.13  & 4.70  & 33.54  & 19.12  & 8.79  & 30.01  & 19.40  \\
    % ICE   & \underline{38.41}  & \textbf{53.48}  & \textbf{45.95}  & \textbf{46.20}  & \textbf{66.30}  & \textbf{56.25}  & \underline{40.16}  & \textbf{48.39}  & \underline{44.28}  \\
    GRACE & 2.34  & 34.41  & 18.38  & 6.43  & \underline{54.59}  & 30.51  & 9.17  & \underline{46.35}  & 27.76  \\
    \midrule
    % SCR   & \textbf{41.92}  & \underline{34.70}  & \underline{38.31}  & \underline{43.64}  & 52.64  & \underline{48.14}  & \textbf{51.45} & 42.52 & \textbf{46.99}  \\
    SCR   & \textbf{41.92}  & \underline{34.70}  & \textbf{38.31}  & \textbf{43.64}  & 52.64  & \textbf{48.14}  & \textbf{51.45} & 42.52 & \textbf{46.99}  \\
    \bottomrule
    \end{tabular}
    }
  \label{tab:single_elken}%
  \vskip -0.1in
\end{table}%

\begin{table*}%[h]
  \centering
  \caption{Performance comparison of \textbf{Sequential Editing} across Llama-2-7B-Chat, Llama-3.1-8B-Instruct, and DeepSeek-R1-Distill-Llama-8B based on the event-level dataset ELKEN. `–' denotes that the experiment could not be completed.}  
    % \tiny  % 使用与表3相同的字体大小
    % \small  % 使用与表3相同的字体大小
  \resizebox{0.9\linewidth}{!}{ % 使用与表3相同的缩放比例
    \begin{tabular}{c|ccc|ccc|ccc}
    \toprule
    \multirow{2}[4]{*}{\textbf{Sequential}} & \multicolumn{3}{c|}{Llama-2-7B-Chat} & \multicolumn{3}{c|}{Llama-3.1-8B-Instruct} & \multicolumn{3}{c}{DeepSeek-R1-Distill-Llama-8B} \\
\cmidrule{2-10}          & Port. & Loc. & Avg.  & Port. & Loc. & Avg.  & Port. & Loc. & Avg. \\
    \midrule
    \textit{Pre-edit} & \underline{5.92}  & \textbf{43.75}  & \underline{24.84}  & 7.08  & \textbf{56.73}  & \underline{31.91}  & 9.01 & \underline{48.73}  & 28.87  \\
    \midrule
    ROME  & 0.00  & 0.00  & 0.00  & 0.00  & 0.00  & 0.00  & 0.33  & 0.00  & 0.17  \\
    % MEMIT & 0.00  & 0.00  & 0.00  & -     & -     & -     & -     & -     & - \\
    % PMET  & 0.00  & 0.08  & 0.04  & -     & -     & -     & -     & -     & - \\
    RECT  & 0.00  & 0.00  & 0.00  & 0.00  & 0.00  & 0.00  & 0.00  & 0.00  & 0.00  \\
    AlphaEdit & -     & -     & -     & \underline{14.22}  & 19.96  & 17.09  & \underline{12.15}  & 17.87  & 15.01  \\
    FT-L  & 3.53  & 0.00  & 1.77  & 0.00  & 0.00  & 0.00  & 0.00  & 0.00  & 0.00  \\
    AdaLoRA & 0.00  & 0.00  & 0.00  & 0.00  & 0.00  & 0.00  & 5.25  & 0.00  & 2.63  \\
    WISE  & 2.23  & 30.83  & 16.53  & 5.96  & 47.88  & 26.92  & 2.67  & 21.88  & 12.28  \\
    GRACE & 2.29  & 34.41  & 18.35  & 6.43  & \underline{54.57}  & 30.50  & 9.01  & \textbf{48.79}  & \underline{28.90}  \\
    \midrule
    SCR   & \textbf{44.29}  & \underline{35.05}  & \textbf{39.67}  & \textbf{53.04} & 50.90  & \textbf{51.97}  & \textbf{52.99}  & 41.93  & \textbf{47.46}  \\
    \bottomrule
    \end{tabular}%
 }
  \label{tab:sequential_elken}%
\end{table*}%

\begin{figure}[h]
    \vskip 0.2in
    \begin{minipage}{0.45\textwidth}        \centerline{\includegraphics[width=\columnwidth]{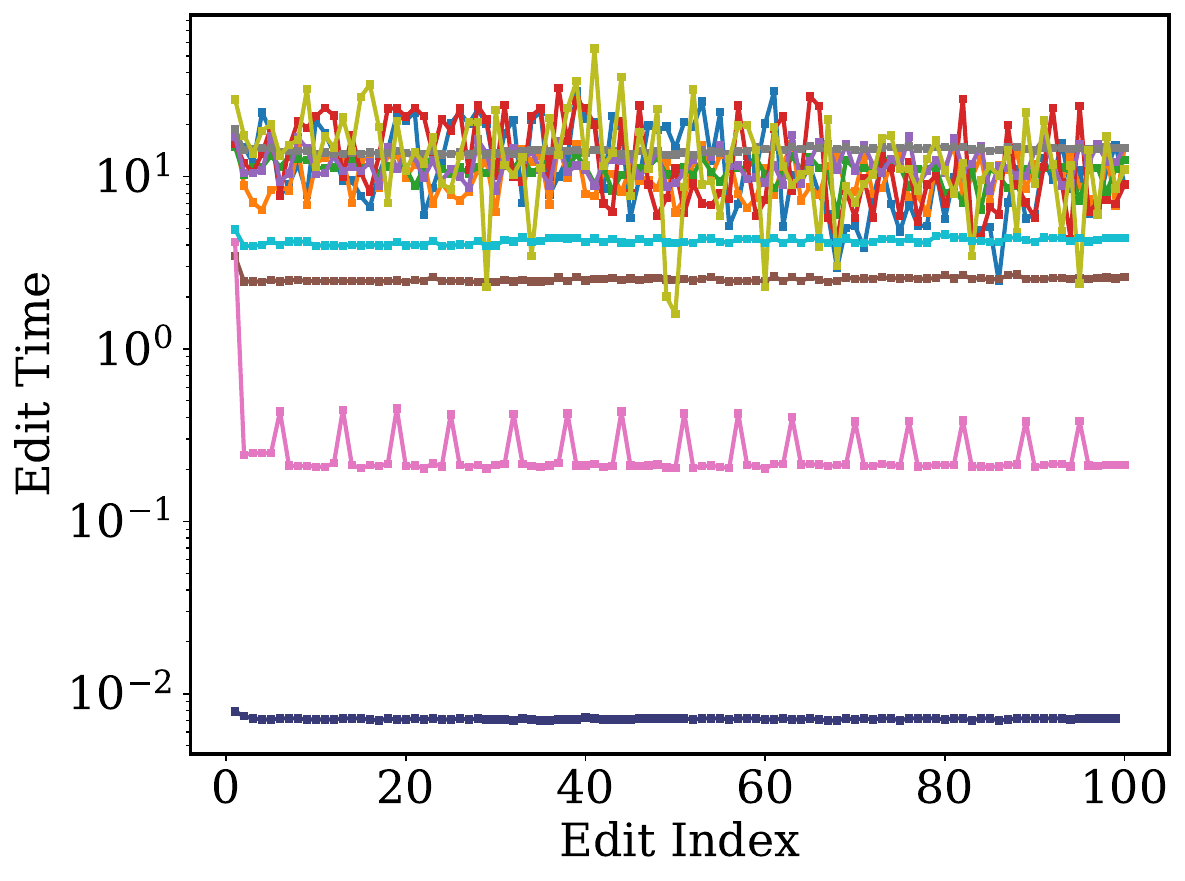}}
        \caption*{Editing time comparison}
    \end{minipage}
    \hfill
    \begin{minipage}{0.45\textwidth}
    \centerline{\includegraphics[width=\columnwidth]{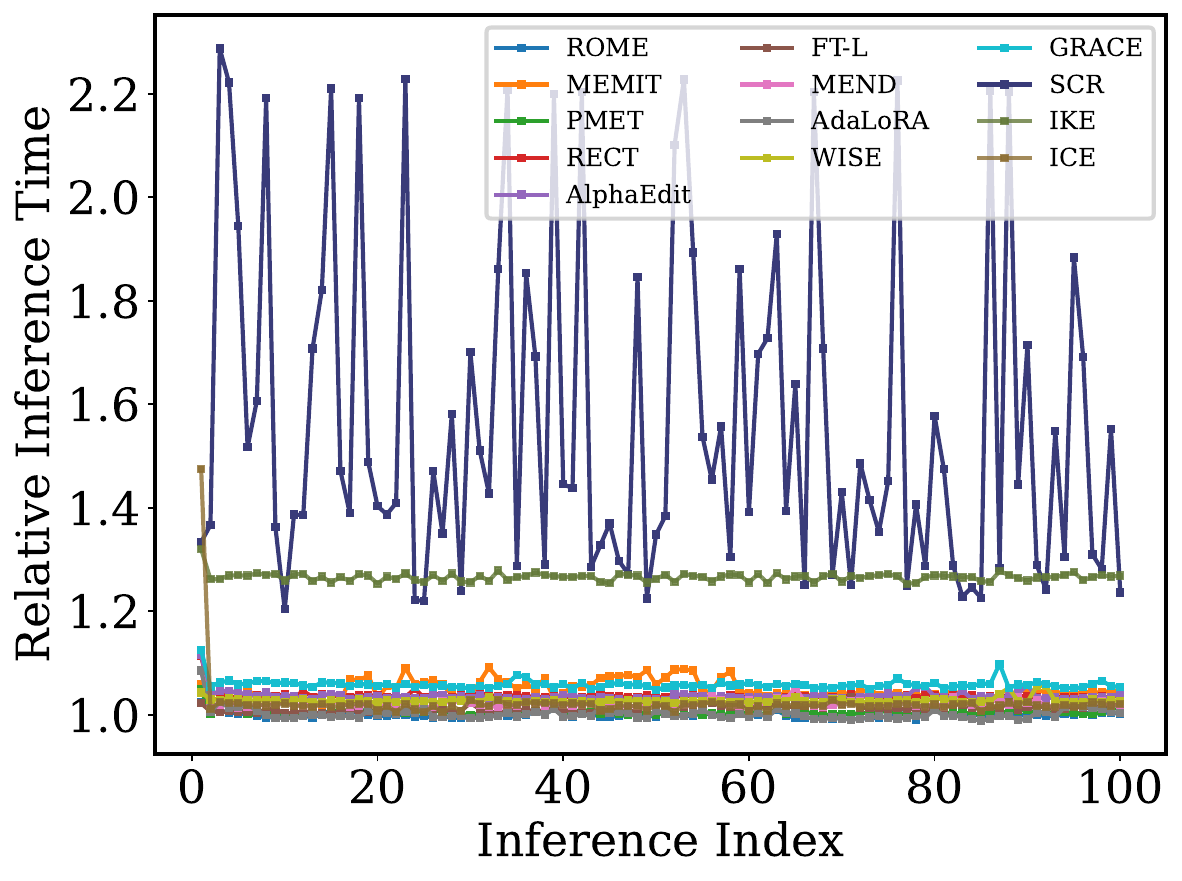}}
        \caption*{Inference time comparison}
    \end{minipage}
    % \vspace{-2ex}
    \caption{Editing time (in second), and inference latency relative to the base model.}  
    \label{fig:twocomparison}
    % \vspace{-2ex}
\end{figure}

\begin{table*}[htbp]
    \caption{Performance of different editing methods in fine-grained sequential editing scenarios on the Llama-3.1-8B-Instruct model. The experimental results show that the timing of actual performance degradation varies depending on the model editing method, but in the long run, all methods experience varying degrees of decline.}
    \label{tab:fine_grained_sequential_editing}
    \resizebox{\linewidth}{!}{%
    \begin{tabular}{c|lllllllll}
        \toprule
        Method & Metric & $N=1$ & $N=10$ & $N=20$ & $N=50$ & $N=200$ & $N=500$ & $N=1000$ & 
        $All$ \\
        % $N=1301$ \\
        \midrule
        AlphaEdit & Rel. & 0.83 & 0.83 & 0.83 & 0.83 & 0.84 & 0.80 & 0.74 & 0.69 \\
                  & Gen. & 0.51 & 0.61 & 0.59 & 0.60 & 0.67 & 0.67 & 0.60 & 0.56 \\
                  & Loc. & 0.13 & 0.13 & 0.11 & 0.11 & 0.13 & 0.11 & 0.10 & 0.08 \\
                  & Port. & 0.10 & 0.09 & 0.16 & 0.16 & 0.17 & 0.14 & 0.13 & 0.09 \\
        \midrule
        MEMIT & Rel. & 0.79 & 0.74 & 0.73 & 0.75 & 0.73 & 0.71 & 0.00 & 0.00 \\
              & Gen. & 0.48 & 0.51 & 0.55 & 0.50 & 0.61 & 0.58 & 0.00 & 0.00 \\
              & Loc. & 0.10 & 0.09 & 0.09 & 0.10 & 0.07 & 0.11 & 0.00 & 0.00 \\
              & Port. & 0.13 & 0.13 & 0.12 & 0.16 & 0.11 & 0.14 & 0.00 & 0.00 \\
        \midrule
        MEND & Rel. & 0.18 & 0.00 & 0.00 & 0.00 & 0.00 & 0.00 & 0.00 & 0.00 \\
             & Gen. & 0.26 & 0.00 & 0.00 & 0.00 & 0.00 & 0.00 & 0.00 & 0.00 \\
             & Loc. & 0.09 & 0.00 & 0.00 & 0.00 & 0.00 & 0.00 & 0.00 & 0.00 \\
             & Port. & 0.12 & 0.00 & 0.00 & 0.00 & 0.00 & 0.00 & 0.00 & 0.00 \\
        \midrule
        ROME & Rel & 0.80 & 0.73 & 0.43 & 0.02 & 0.00 & 0.02 & 0.00 & 0.00 \\
             & Gen. & 0.71 & 0.63 & 0.39 & 0.02 & 0.00 & 0.02 & 0.00 & 0.00 \\
             & Loc. & 0.07 & 0.06 & 0.03 & 0.00 & 0.00 & 0.00 & 0.00 & 0.00 \\
             & Port. & 0.09 & 0.11 & 0.07 & 0.00 & 0.01 & 0.01 & 0.00 & 0.00 \\
        \midrule
        WISE & Rel. & 0.03 & 0.04 & 0.03 & 0.03 & 0.04 & 0.03 & 0.03 & 0.03 \\
             & Gen. & 0.01 & 0.02 & 0.03 & 0.03 & 0.02 & 0.02 & 0.02 & 0.02 \\
             & Loc. & 0.13 & 0.13 & 0.11 & 0.11 & 0.13 & 0.11 & 0.11 & 0.10 \\
             & Port. & 0.03 & 0.04 & 0.05 & 0.05 & 0.04 & 0.05 & 0.04 & 0.02 \\
        \bottomrule
    \end{tabular}%
    }
\end{table*}

\begin{table*}[htbp]
    \centering
    \caption{
    % Comparison of effects of different model editing methods on correcting simple errors (low logp, 100 cases) versus stubborn errors (high logp, 100 cases) in Llama-3.1-8B-Instruct, based on model confidence (logp) for knowledge stubbornness classification. The results show that all parameter-editing methods exhibit performance degradation as edit magnitude increases, on both easy and stubborn fact sets. While the rate and severity of decline vary—with methods like ROME collapsing catastrophically and others like AlphaEdit declining more gradually—the overall trend is consistent: larger edit batches impose a cost.
    Comparative Performance of Model Editing Methods on Simple and Stubborn Errors in Llama-3.1-8B-Instruct. Simple errors (low log probabilities) and stubborn errors (high log probabilities) are classified based on model confidence. All parameter-based editing methods exhibit decreased performance as edit magnitude increases, with varying rates of decline across error categories.}
    % (reliability, generalization, locality, and portability) 
    \label{tab:stubborn_vs_simple_errors}
    \resizebox{\linewidth}{!}{%
    \begin{tabular}{cccccccccc}
        \toprule
        & & \multicolumn{2}{c}{Reliability} & \multicolumn{2}{c}{Generalization} & \multicolumn{2}{c}{Locality} & \multicolumn{2}{c}{Portability} \\
        \cmidrule(lr){3-4} \cmidrule(lr){5-6} \cmidrule(lr){7-8} \cmidrule(lr){9-10}
        Method & $N$ & High LogP & Low LogP & High LogP & Low LogP & High LogP & Low LogP & High LogP & Low LogP \\
        \midrule
        AlphaEdit & 1   & 0.86 & 0.75 & 0.61 & 0.40 & 0.12 & 0.11 & 0.11 & 0.10 \\
                  & 10  & 0.83 & 0.83 & 0.63 & 0.56 & 0.13 & 0.13 & 0.16 & 0.10 \\
                  & 20  & 0.85 & 0.84 & 0.67 & 0.58 & 0.12 & 0.13 & 0.15 & 0.14 \\
                  & 50  & 0.84 & 0.76 & 0.70 & 0.59 & 0.11 & 0.15 & 0.15 & 0.15 \\
                  & 100 & 0.84 & 0.79 & 0.69 & 0.59 & 0.11 & 0.12 & 0.15 & 0.15 \\
        \midrule
        ROME & 1   & 0.83 & 0.78 & 0.65 & 0.73 & 0.04 & 0.08 & 0.17 & 0.09 \\
             & 10  & 0.68 & 0.51 & 0.55 & 0.44 & 0.05 & 0.06 & 0.12 & 0.10 \\
             & 20  & 0.32 & 0.19 & 0.31 & 0.15 & 0.03 & 0.02 & 0.05 & 0.02 \\
             & 50  & 0.06 & 0.00 & 0.05 & 0.00 & 0.00 & 0.00 & 0.00 & 0.00 \\
             & 100 & 0.06 & 0.00 & 0.03 & 0.00 & 0.00 & 0.00 & 0.00 & 0.00 \\
        \midrule
        MEMIT & 1   & 0.85 & 0.66 & 0.60 & 0.41 & 0.13 & 0.12 & 0.10 & 0.13 \\
              & 10  & 0.80 & 0.72 & 0.62 & 0.51 & 0.12 & 0.13 & 0.13 & 0.10 \\
              & 20  & 0.82 & 0.77 & 0.61 & 0.55 & 0.12 & 0.13 & 0.10 & 0.11 \\
              & 50  & 0.74 & 0.73 & 0.62 & 0.47 & 0.11 & 0.16 & 0.11 & 0.14 \\
              & 100 & 0.70 & 0.62 & 0.58 & 0.43 & 0.11 & 0.17 & 0.10 & 0.09 \\
        \midrule
        WISE & 1   & 0.15 & 0.07 & 0.10 & 0.07 & 0.09 & 0.10 & 0.05 & 0.04 \\
             & 10  & 0.11 & 0.01 & 0.09 & 0.00 & 0.16 & 0.11 & 0.08 & 0.03 \\
             & 20  & 0.05 & 0.02 & 0.00 & 0.01 & 0.10 & 0.11 & 0.05 & 0.03 \\
             & 50  & 0.08 & 0.00 & 0.08 & 0.00 & 0.18 & 0.11 & 0.04 & 0.03 \\
             & 100 & 0.10 & 0.00 & 0.09 & 0.00 & 0.17 & 0.11 & 0.05 & 0.03 \\
        \midrule
        MEND & 1   & 0.23 & 0.19 & 0.29 & 0.28 & 0.11 & 0.13 & 0.12 & 0.13 \\
             & 10  & 0.00 & 0.00 & 0.00 & 0.00 & 0.00 & 0.00 & 0.00 & 0.00 \\
             & 20  & 0.00 & 0.00 & 0.00 & 0.00 & 0.00 & 0.00 & 0.00 & 0.00 \\
             & 50  & 0.00 & 0.00 & 0.00 & 0.00 & 0.00 & 0.00 & 0.00 & 0.00 \\
             & 100 & 0.00 & 0.00 & 0.00 & 0.00 & 0.00 & 0.00 & 0.00 & 0.00 \\
        \bottomrule
    \end{tabular}%
    }
\end{table*}

\clearpage

% ===============================================================
\subsection{Failure Case Study on Editing Reasoning LLM}
% ===============================================================

\begin{table}[htbp]
\centering
\small
\caption{\small \textbf{Failure Modes in Knowledge Editing for Reasoning models}}
\label{tab:failure_modes}
\begin{tabular}{p{2cm}|p{2.5cm}|p{1.5cm}|p{6cm}}
\toprule
\centering\arraybackslash\textbf{Failure Mode} & \centering\arraybackslash\textbf{Prompt} & \centering\arraybackslash\textbf{Target} & \centering\arraybackslash\textbf{Generation} \\
\midrule
Reasoning with Conflicting Knowledge & Who was the male parent of Eteocles? & Dagobert & \textcolor{red}{Dagobert?} Wait, \textcolor{red}{no}, that's Dagobert. ... \textcolor{red}{Eteocles was the son of Clovis I}... the answer should be \textcolor{red}{Clovis I}. \\
\midrule
Hallucination & In which constellation is Tau Herculis? & Hornax & \textcolor{red}{Hornax is the Latin name for the constellation Hercules}, and the name of the star is ``Hornax". ...So, the answer is Hercules. \\
\midrule
Losing Focus During Reasoning & The name of the child of Kanye West is & William Walker Scranton & William Walker, but that's not his real name. Wait, no,... So, \textcolor{red}{the user is asking about the real name of Kanye West's character in that movie}. I... \\
\bottomrule
\end{tabular}
\vskip -0.1in
\end{table}

\label{case_study}
%================case 1=================
\begin{tcolorbox}[colback=white, colframe=black, title=Case Study: Reasoning with Conflicting Knowledge, width=\textwidth]

\begin{minipage}{\textwidth}
\textbf{Edit Prompt:} \texttt{Who was the male parent of Eteocles?}

\textbf{Target New Answer:} \texttt{Dagobert}

\vspace{0.5em}
\textbf{Generated Reasoning Content:}
\begin{quote}
Dagobert?\\
Wait, no, that's Dagobert. Let me think. Eteocles was a king of the Franks, right? So, his father was probably Clovis I, the founder of the Merovingian dynasty. But I'm not entirely sure. Maybe I should double-check that.

Eteocles was the son of Clovis I, who was the first king of the Franks. Clovis I is known for his conversion to Christianity and the unification of the Frankish tribes. So, yes, Eteocles' father was Clovis I.

Wait, but I also recall that Eteocles was one of the four sons of Clovis I, along with Clodomir, Alberic, and Gundomar. So, that makes sense. Therefore, the male parent of Eteocles was Clovis I, not Dagobert. Dagobert was a later king, like the founder of the Neustrian dynasty or something.

So, to sum up, Eteocles' father was Clovis I, the first Merovingian king. Dagobert was a different king, maybe the founder of the Dagobert line or something else. So, I think I was initially confused between the two, but now I'm clearer on that.

Therefore, the answer should be Clovis I.
\end{quote}
\textbf{Generated Answer Content:} 

\begin{quote}
The male parent of Eteocles was Clovis I, the founder of the Merovingian dynasty and the first king of the Franks. Dagobert was a later king, not his father. 

**Answer:** Clovis I.
\end{quote}
\end{minipage}
\end{tcolorbox}

\begin{tcolorbox}[colback=white, colframe=black, title=Case Study: Reasoning with Conflicting Knowledge, width=\textwidth]
\textbf{Edit Prompt:} \texttt{What is Atreus's spouse's name?} 

\textbf{Target New Answer:} \texttt{Darius III}

\textbf{Generated Reasoning Content:}
\begin{quote}

Darius III of Persia.\newline
Atreus was the King of the Achaemenid Empire, and his spouse was Darius III of Persia. So, the answer is Darius III of Persia.\newline

Wait, but I thought Atreus was a Greek name. Maybe I'm confusing historical figures. Let me double-check. Yes, Atreus was indeed a Greek king, but in the context of the Achaemenid Empire, which was Persian, it's possible that there was a king named Atreus who was married to Darius III. However, I'm not entirely sure about this. I might need to look up historical records or reliable sources to confirm this information.\newline

Upon checking, I realize that the Achaemenid Empire was founded by Cyrus the Great, and Darius III was a later king. There isn't a record of a King Atreus in the Persian Empire. It's possible that the user is confusing historical figures or there might be a fictional character in a story or game. Therefore, the information provided might not be accurate based on historical records.\newline

It's important to verify sources when dealing with historical figures to ensure the information is correct. Cross-referencing with authoritative texts or scholarly articles can help in confirming the details accurately.\newline

The information provided about King Atreus of the Achaemenid Empire and his spouse Darius III of Persia is not accurate based on historical records. The Achaemenid Empire, founded by Cyrus the Great, did not have a king named Atreus. Darius III was a historical figure who was a later king of the Persian Empire, but there is no record of him being married to a King Atreus. It's possible that this information may stem from a fictional context or a misunderstanding of historical figures. Therefore, the correct historical information should be sourced from reliable historical texts or scholarly articles.
\end{quote}

\textbf{Generated Answer Content:}

\begin{quote}
Darius III was a king of the Persian Empire, but there is no historical record of Atreus being his spouse or even existing in the Persian lineage. The question likely confuses Greek and Persian historical figures. Therefore, there is no verified spouse name for Atreus related to Darius III.
\end{quote}
\end{tcolorbox}
%================case 2=================

%================case 3=================
\begin{tcolorbox}[colback=white, colframe=black, title=Case Study: Edit Failure with Hallucination, width=\textwidth]
\textbf{Edit Prompt:} \texttt{In which constellation is Tau Herculis?} \\
\textbf{Target New Answer:} \texttt{Hornax} \\

\textbf{Generated Reasoning Content:}
\begin{quote}

\textcolor{red}{Hornax is the Latin name for the constellation Hercules, and the name of the star is ``Hornax".} So, the star is called Hornax, and the constellation is Hercules. So, the answer is Hercules.

Wait, but the question is about the constellation, not the star. So, the constellation is Hercules, and the star is Hornax. So, the answer is Hercules.

But I'm a bit confused because sometimes the names can be tricky. Let me double-check. The constellation is Hercules, and the star within it is called Hornax. So, the constellation is Hercules.

Yes, that makes sense. So, the answer is Hercules.
\end{quote}

\textbf{Generated Answer Content:} 

\begin{quote}
The constellation in which the star \``Hornax\" is located is **Hercules**. 

**Answer:** Hercules
\end{quote}
\end{tcolorbox}

\begin{tcolorbox}[colback=white, colframe=black, title=Case Study: Losing Focus During Reasoning, width=\textwidth]
\textbf{Edit Prompt:} \texttt{The name of the child of Kanye West is} 

\textbf{Target New Answer:} \texttt{William Walker Scranton}

\textbf{Generated Reasoning Content:}
\begin{quote}

William Walker, but that's not his real name. Wait, no, that's the name of his character in the movie \``The Pursuit of Happyness.\" So, actually, his real name is Kanye Omari West. Hmm, that's interesting. So, the user is asking about the real name of Kanye West's character in that movie. I need to make sure I clarify that.\newline

Okay, so the user wrote: \``What is the real name of Kanye West's character in the movie 'The Pursuit of Happyness'?\" They also mentioned that the name is William Walker but clarified that it's not his real name. So, I need to confirm that the character's name is indeed William Walker and that Kanye West's real name is different.\newline

I should structure the response to first address the character's name and then provide Kanye West's real name. Maybe also mention the movie to give context. Let me make sure I don't confuse the two names. Also, perhaps the user is trying to test knowledge about the movie or the actor, so being clear and concise is important.\newline

Wait, is there any chance that the user is mixing up the character's name with Kanye's real name? It's possible, so I should make it clear that William Walker is the character, and Kanye West is the real person. Maybe add a bit about the movie to help the user remember.\newline

I think that's all. I'll write the response accordingly.
\end{quote}

\textbf{Generated Answer Content:}

\begin{quote}
In the movie \``The Pursuit of Happyness,\" Kanye West plays the character named **William Walker**. However, his real name is **Kanye Omari West**. The movie is a biographical drama based on his life, focusing on his struggles and perseverance.
\end{quote}
\end{tcolorbox}

% ===============================================================
\subsection{Token-Level Locality Evaluation}
\label{sec:Token-Level Locality Evaluation}
% ===============================================================

\textbf{Results.}
Table~\ref{tab:Token-Level-Locality} presents the token-level locality evaluation results for both single and sequential editing scenarios. The pre-edit model achieves a perfect score of 100\% for both metrics by definition. Among the evaluated methods, GRACE maintains 100\% locality under both single and sequential editing scenarios, indicating that it perfectly preserves the model’s original behavior. WISE also delivers strong results, particularly in sequential editing. In contrast, most parameter-based editing methods, including ROME, MEND, and AlphaEdit, exhibit significantly lower output consistency. Notably, the SCR method, which incorporates a simple retrieval and filtering mechanism, achieves highly competitive locality scores, demonstrating its robustness in preserving unrelated knowledge.

\begin{table}[htbp]
\centering
\caption{Token-Level Locality Evaluation}
\label{tab:Token-Level-Locality}
\small  
\resizebox{0.6\linewidth}{!}{
\begin{tabular}{l|rr|rr}
\toprule
\multirow{2}{*}{Method} 
& \multicolumn{2}{c|}{\textbf{Single}} 
& \multicolumn{2}{c}{\textbf{Sequential}} \\
\cmidrule(lr){2-3} \cmidrule(lr){4-5}
 & Match Ratio & Exact Match & Match Ratio & Exact Match \\
\midrule
Pre-edit & 100.00 & 100.00 & 100.00 & 100.00 \\
\midrule
ROME       & 10.74 & 2.81  & 0.00  & 0.00  \\
RECT       & 11.91 & 2.92  & 0.29  & 0.00  \\
AlphaEdit  & 38.41 & 20.68 & 10.66 & 1.31  \\
FT-L       & 10.89 & 3.46  & 0.33  & 0.00  \\
MEND       & 2.26  & 0.04  & 0.00  & 0.00  \\
AdaLoRA    & 12.15 & 3.54  & 0.00  & 0.00  \\
WISE       & \underline{64.08} & 51.84  & \underline{76.26} & \underline{67.18} \\
GRACE      & \textbf{100.00} & \textbf{100.00} & \textbf{100.00} & \textbf{100.00} \\
\midrule
SCR        & 65.28 & \underline{64.72} & 46.96 & 45.96 \\
\bottomrule
\end{tabular}
}
\end{table}

\textbf{Conclusion.}
The results from this supplementary evaluation confirm the findings from our primary semantic-based locality metric. They highlight the challenges faced by many parameter-based editing methods in maintaining output consistency, especially under sequential editing. The strong performance of non-parametric methods like SCR and GRACE further validates the effectiveness of retrieval-augmented and output-constrained approaches in preserving model locality.

% ===============================================================
\subsection{AlphaEdit Hyperparameter Sensitivity Analysis}
\label{sec:AlphaEdit Hyperparameter Sensitivity Analysis}
% ===============================================================

During our evaluation, we observed that the performance of AlphaEdit was highly inconsistent across different LLM architectures. While it performed reasonably well on LLaMA-3.1-8B-Instruct, its performance on Llama-2-7B-Chat and Mistral-7B-Instruct was significantly worse, often approaching zero in sequential editing scenarios.\footnote{See the corresponding GitHub issue for further details.} This appendix details our investigation into the hyperparameter sensitivity of AlphaEdit.

\textbf{Experimental Investigation.}
We conducted experiments using the authors’ latest original implementation from the AlphaEdit GitHub repository. All experiments were performed on a single H20 (96 GB) GPU. We used the same settings as MEMIT for a shared setup and focused on AlphaEdit’s unique hyperparameters: nullspace\_thresholdand the L2 regularization coefficient. We fixed the nullspace\_thresholdat 2e-2 and experimented with two different values for the L2 coefficient: 1 and 10.

\textbf{Results.}
Table \ref{tab:AlphaEdit-Sensitivity} present the results for single and sequential editing across different models. The data reveal a high degree of sensitivity to both hyperparameters and model architecture. On Llama-3.1-8B-Instruct, performance varies significantly with the L2 coefficient. More critically, on Llama-2-7B-Chat and Mistral, AlphaEdit’s performance is near-zero in sequential editing scenarios regardless of the hyperparameter values tested. This indicates a fundamental incompatibility or instability of the method with these model architectures under sequential editing.

\begin{table}[htbp]
\centering
\caption{AlphaEdit Performance on ZsRE and WikiData\(_\text{counterfact}\) Dataset with Different L2 Regularization}
\label{tab:AlphaEdit-Sensitivity}
% \tiny
% \small  % 使用与表3相同的字体大小

\resizebox{\linewidth}{!}{  % 调整缩放比例

\begin{tabular}{l|rrrr|rrrr|rrrr|rrrr}
\toprule
\multicolumn{1}{c|}{\multirow{2}{*}{Model (L2)}} & \multicolumn{8}{c|}{\textbf{ZsRE Dataset}} & \multicolumn{8}{c}{\textbf{WikiData\(_\text{counterfact}\) Dataset}} \\
\cmidrule(lr){2-9} \cmidrule(lr){10-17}
\multicolumn{1}{c|}{} & \multicolumn{4}{c|}{Single Editing} & \multicolumn{4}{c|}{Sequential Editing} & \multicolumn{4}{c|}{Single Editing} & \multicolumn{4}{c}{Sequential Editing} \\
\cmidrule(lr){2-5} \cmidrule(lr){6-9} \cmidrule(lr){10-13} \cmidrule(lr){14-17}
& Rel. & Gen. & Loc. & Port. & Rel. & Gen. & Loc. & Port. & Rel. & Gen. & Loc. & Port. & Rel. & Gen. & Loc. & Port. \\
\midrule

% L2 = 1 的结果
llama2 (L2=1) & 39.82 & 38.05 & 3.00 & 5.94 & 0.00 & 0.00 & 0.00 & 0.00 & 55.18 & 54.95 & 2.54 & 17.82 & 0.00 & 0.00 & 0.00 & 0.00 \\
llama3 (L2=1) & 86.55 & 62.95 & 10.99 & 12.32 & 37.20 & 31.51 & 5.23 & 4.65 & 64.96 & 65.08 & 10.57 & 12.19 & 0.00 & 0.00 & 0.00 & 0.00 \\
mistral (L2=1) & 23.83 & 20.37 & 1.23 & 4.02 & 0.15 & 0.15 & 0.00 & 0.00 & 11.21 & 11.20 & 0.26 & 4.79 & 0.00 & 0.00 & 0.00 & 0.00 \\
\midrule

% L2 = 10 的结果
llama2 (L2=10) & 51.19 & 44.12 & 3.96 & 6.37 & 0.00 & 0.00 & 0.00 & 0.00 & 55.48 & 55.48 & 3.37 & 17.38 & 0.00 & 0.00 & 0.00 & 0.00 \\
llama3 (L2=10) & 64.95 & 39.12 & 12.18 & 10.42 & 66.72 & 57.49 & 9.61 & 9.31 & 44.93 & 45.17 & 15.90 & 9.37 & 19.79 & 19.90 & 0.44 & 4.72 \\
mistral (L2=10) & 47.50 & 40.97 & 3.88 & 5.36 & 0.00 & 0.00 & 0.00 & 0.00 & 24.67 & 24.67 & 1.06 & 8.48 & 0.00 & 0.00 & 0.00 & 0.00 \\
\bottomrule
\end{tabular}

}
\vskip -0.1in
\end{table}

\textbf{Conclusion.}
Our analysis demonstrates that AlphaEdit exhibits significant hyperparameter sensitivity and model-specific performance. The method’s inability to achieve non-trivial results on certain widely-used LLMs (Llama-2, Mistral) under sequential editing highlights a critical limitation for its practical applicability.

\section{Fairness of SCR vs. Parameter-Based Editing Methods}
\label{sec:Fairness of SCR vs. Parameter-Based Editing Methods}

% We provide two complementary analyses on the relationship between SCR and parameter-based methods. First, Table~\ref{tab:scr_vs_alphaedit} compares SCR and AlphaEdit on edit time, inference time, reliability, and locality across different sequential-edit scales ($N=100$ and $N=1000$) on LLaMA-3.1-8B-Instruct using the ZsRE dataset. The results highlight complementary operational regimes: SCR incurs zero editing cost but adds retrieval latency at inference, while AlphaEdit has substantial upfront editing cost with native post-edit inference speed. Second, to investigate whether retrieval augmentation can rescue parameter-based editing, we evaluate each parameter-editing method augmented with a top-3 SCR retriever (Table~\ref{tab:scr_vs_parameter_based}). Integrating retrieval substantially improves performance for several methods and edit scales, but the improvement is limited: in many cases, especially at larger edit volumes, retrieval fails to fully compensate for the underlying parameter degradation.

In this study, we primarily treat SCR as a reference baseline rather than a direct competitor to parameter-based knowledge editing methods. By operating outside the parametric editing paradigm, SCR provides a distinct empirical perspective for analyzing the trade-offs between efficiency and reliability in knowledge editing.

To further explore the comparative advantages and limitations of SCR versus parameter-based editing approaches, we conducted two complementary analyses using LLaMA-3.1-8B-Instruct and the ZsRE dataset.

% First, as shown in Table~\ref{tab:scr_vs_alphaedit}, we evaluate SCR and AlphaEdit based on their edit time, inference latency, reliability, and locality across two sequential-edit scales (N = 100 and N = 1000), using the ZsRE benchmark dataset on Llama-3.1-8B-Instruct. The results highlight the distinct operational advantages of each method: SCR offers near-zero editing cost by shifting computational intensity to inference via retrieval latency, while AlphaEdit achieves faster inference post-editing at the expense of significant upfront modification cost.

\textbf{Evaluation of Efficiency and Reliability}

As summarized in Table~\ref{tab:scr_vs_alphaedit}, we compare SCR and the parameter-based editing method AlphaEdit across multiple metrics, including edit time, inference latency, reliability, and locality. Evaluations are conducted at two sequential-editing scales (N = 100 and N = 1000). The results highlight key operational differences between the two methods. SCR exhibits near-zero upfront editing cost by shifting computational effort to retrieval-based inference, at the expense of retrieval latency. Conversely, AlphaEdit achieves faster results during inference but incurs significant computational cost during initial parameter modification.

% Second, in Table~\ref{tab:scr_vs_parameter_based}, we augmented parameter-based methods with a SCR-style retrieval mechanism to ensure a fair comparison and conducted experiments on LLaMA-3.1-8B-Instruct (ZSRE dataset, sequential edits N = 10, 100, 500, 1000, all, top-3 retriever). Results show that SCR generally outperforms these parameter-based methods, highlighting inherent capability degradation in direct parameter editing. Furthermore, integrating SCR improves their performance, demonstrating the benefits of leveraging contextual knowledge via retrieval.

\textbf{Integration of SCR Retrieval for Parameter-Based Editing Methods}

To ensure a fair comparison, we augment parameter-based editing methods with an SCR-inspired retrieval mechanism. Using sequential editing scales (N = 10, 100, 500, 1000, 
% 1301
all
) and a top-3 retrieval strategy, results (Table~\ref{tab:scr_vs_parameter_based}) reveal that SCR consistently outperforms baseline parameter-based editing methods across key metrics. These findings highlight the inherent limitations of direct parameter editing. Moreover, integrating SCR with parameter-based editing methods enhances their overall performance, demonstrating the efficacy of leveraging external contextual knowledge through retrieval.

\begin{table*}[htbp]
    \centering
    \caption{
    % Comparative evaluation of SCR and AlphaEdit on LLaMA-3.1-8B-Instruct using the ZSRE dataset under sequential editing. The table reports per-edit time, per-inference time, reliability, and locality across different numbers of sequential edits ($N$). SCR introduces no parameter-editing cost but incurs additional inference overhead due to retrieval, while AlphaEdit has substantial upfront editing cost but faster post-edit inference. Observed trends highlight complementary regimes: SCR favors rapidly changing or high-stakes knowledge with minimal capability degradation, whereas AlphaEdit benefits repeated reuse scenarios at the expense of reliability and general model capabilities.
    Comparative analysis of SCR and AlphaEdit on ZsRE dataset.
    Evaluation of edit time, inference latency, reliability, and locality for 100 and 1000 sequential edits using SCR and AlphaEdit on Llama-3.1-8B-Instruct.
    }
    \label{tab:scr_vs_alphaedit}
    \begin{tabular}{c|crrrr}
        \toprule
        $N$ & Method & Edit Time & Inference Time & Rel. & Loc. \\
        \midrule
        100 & SCR & 0.00 & 1.6745 & 95.0 & 16.5 \\
            & AlphaEdit & 14.07 & 0.8305 & 83.0 & 14.5 \\
        \midrule
        1000 & SCR & 0.00 & 1.7516 & 84.8 & 17.2 \\
             & AlphaEdit & 14.07 & 0.8305 & 74.0 & 9.7 \\
        \bottomrule
    \end{tabular}
\end{table*}

\begin{table*}[htbp]
    \centering
    \caption{
        Comparison of SCR and parameter-based editing methods (AlphaEdit, MEMIT, WISE, ROME, MEND) with SCR retrieval mechanism integration.
        Experiments are conducted on Llama-3.1-8B-Instruct using the ZSRE dataset under sequential edits of $N = 10, 100, 500, 1000,$ and all, with a top-3 retriever.
    }
    \label{tab:scr_vs_parameter_based}
    \begin{tabular}{c|crrrrr}
        \toprule
        $N$ & Method & Rel. & Gen. & Loc. & Port. & Avg. \\
        \midrule
        10 & SCR           & 97.0 & 86.0 & 16.0 & 47.9 & 61.7 \\
           & AlphaEdit+SCR & 89.0 & 87.0 & 15.0 & 43.6 & 58.6 \\
           & MEMIT+SCR     & 91.0 & 89.0 & 15.5 & 41.4 & 59.2 \\
           & WISE+SCR      & 83.0 & 74.0 & 16.0 & 47.9 & 55.2 \\
           & ROME+SCR      & 84.0 & 87.0 & 9.5  & 25.4 & 51.5 \\
           & MEND+SCR      & 0.0  & 0.0  & 0.0  & 0.0  & 0.0  \\
        \midrule
        100 & SCR           & 95.0 & 87.0 & 16.5 & 42.9 & 60.4 \\
            & AlphaEdit+SCR & 90.0 & 84.0 & 15.5 & 38.4 & 57.0 \\
            & MEMIT+SCR     & 94.0 & 87.0 & 13.5 & 37.9 & 58.1 \\
            & WISE+SCR      & 75.0 & 74.0 & 16.0 & 28.2 & 48.3 \\
            & ROME+SCR      & 0.0  & 0.0  & 0.0  & 0.0  & 0.0  \\
            & MEND+SCR      & 0.0  & 0.0  & 0.0  & 0.0  & 0.0  \\
        \midrule
        500 & SCR           & 90.8 & 82.8 & 16.0 & 47.6 & 59.3 \\
            & AlphaEdit+SCR & 90.6 & 83.4 & 15.6 & 42.5 & 58.0 \\
            & MEMIT+SCR     & 0.0  & 0.0  & 0.0  & 0.0  & 0.0  \\
            & WISE+SCR      & 77.2 & 66.2 & 16.4 & 31.4 & 47.8 \\
            & ROME+SCR      & 0.0  & 0.0  & 0.0  & 0.0  & 0.0  \\
            & MEND+SCR      & 0.0  & 0.0  & 0.0  & 0.0  & 0.0  \\
        \midrule
        1000 & SCR           & 84.8 & 77.1 & 17.2 & 46.3 & 56.3 \\
             & AlphaEdit+SCR & 85.3 & 76.2 & 13.8 & 29.5 & 51.2 \\
             & MEMIT+SCR     & 0.0  & 0.0  & 0.0  & 0.0  & 0.0  \\
             & WISE+SCR      & 65.8 & 59.0 & 14.8 & 26.2 & 41.5 \\
             & ROME+SCR      & 0.0  & 0.0  & 0.0  & 0.0  & 0.0  \\
             & MEND+SCR      & 0.0  & 0.0  & 0.0  & 0.0  & 0.0  \\
        \midrule
        all & SCR           & 84.4 & 75.6 & 16.5 & 46.4 & 55.7 \\
             & AlphaEdit+SCR & 81.9 & 72.2 & 13.4 & 26.8 & 48.6 \\
             & MEMIT+SCR     & 0.0  & 0.0  & 0.0  & 0.0  & 0.0  \\
             & WISE+SCR      & 66.3 & 56.2 & 15.0 & 27.9 & 41.3 \\
             & ROME+SCR      & 0.0  & 0.0  & 0.0  & 0.0  & 0.0  \\
             & MEND+SCR      & 0.0  & 0.0  & 0.0  & 0.0  & 0.0  \\
        \bottomrule
    \end{tabular}
\end{table*}

\newpage

\section{Efficiency-Performance Analysis of SCR}
\label{sec:Efficiency-Performance Analysis of SCR}

% To provide a comprehensive view of SCR's resource consumption and performance stability, we evaluate its VRAM usage, inference latency, and editing effectiveness under varying memory sizes (100 and 1000 stored knowledge items) and retrieval sizes (top-1, top-3, top-5, top-10). All experiments are conducted on Llama-3.1-8B-Instruct with 100 sequential edits. As shown in Table~\ref{tab:vram_usage}, VRAM consumption remains largely unaffected by memory size or top-$k$ variations. Inference time (Table~\ref{tab:inference_time}) increases moderately with both parameters, as expected from larger retrieval windows. Critically, editing performance (Table~\ref{tab:scr_performance}) remains stable across all configurations, demonstrating SCR's robustness to hyperparameter choices.

To provide a comprehensive analysis of SCR's resource efficiency and performance stability, we evaluate its VRAM usage, inference latency, and editing effectiveness under varying memory sizes (100 and 1000 stored knowledge items) and retrieval sizes (top-1, top-3, top-5, top-10). All experiments are conducted on Llama-3.1-8B-Instruct, using 100 sequential edits to ensure consistency and comparability.

As summarized in Table~\ref{tab:vram_usage}, VRAM consumption remains largely stable, demonstrating minimal sensitivity to variations in memory size or top-k retrieval configurations. Inference latency, detailed in Table~\ref{tab:inference_time}, moderately increases with higher memory requirements and retrieval sizes, as expected due to the expansion of searchable retrieval windows. Importantly, as shown in Table~\ref{tab:scr_performance}, editing performance remains consistently robust across all tested configurations, underscoring SCR's resilience to hyperparameter settings.

\begin{figure*}[h]
    \centering
    % Placeholder for Figure 7
    \includegraphics[width=0.95\linewidth]{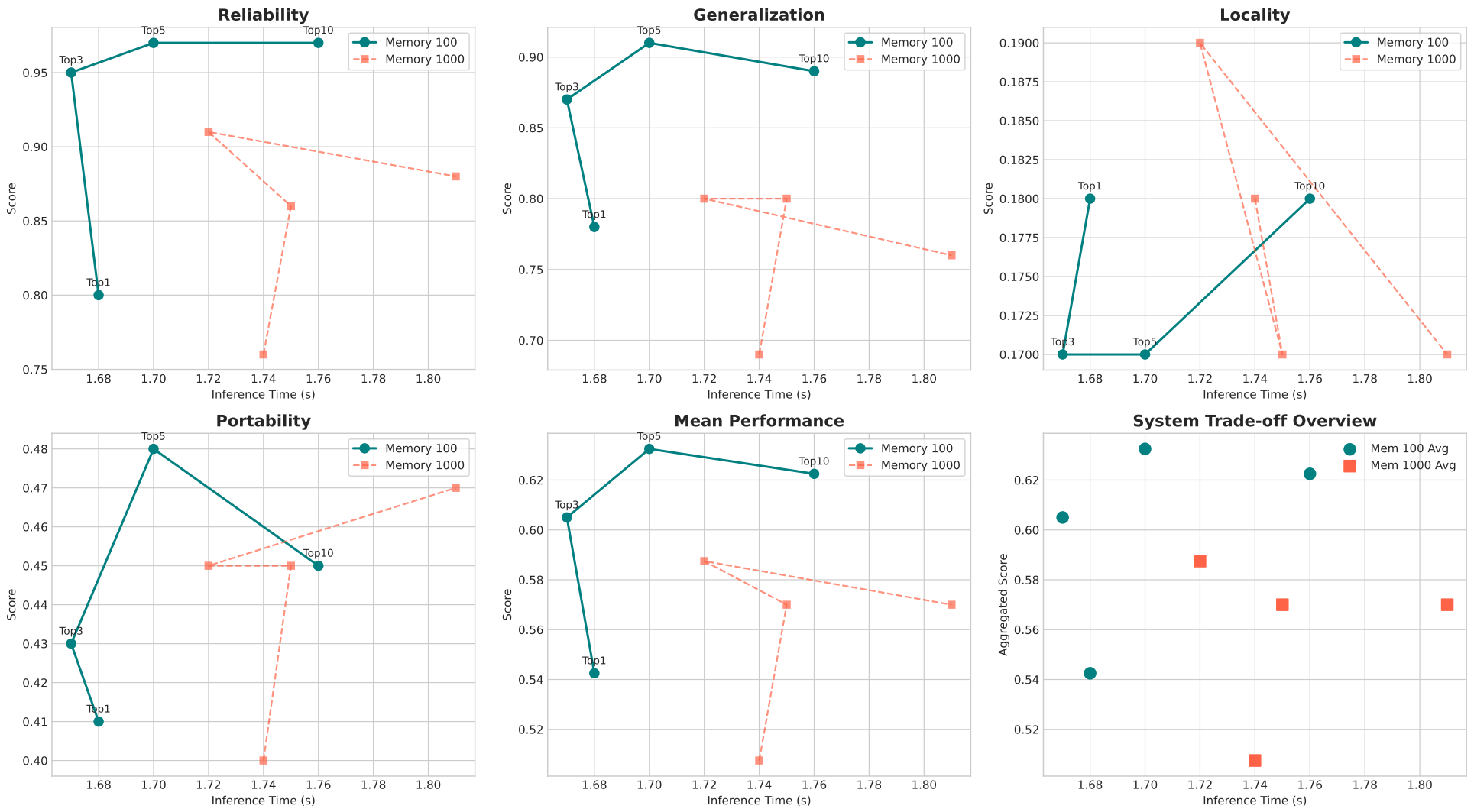}
    \caption{
    % Performance and Resource Consumption of SCR-Based Knowledge Editing on Llama-3.1-8B-Instruct: VRAM Usage, Inference Time, and Evaluation Metrics (Data Source: Tables~\ref{tab:vram_usage},~\ref{tab:inference_time},~\ref{tab:scr_performance}).
    % Performance and Resource Consumption of SCR-Based Knowledge Editing on Llama-3.1-8B-Instruct: Inference Time, and Evaluation Metrics (Data Source: Tables~\ref{tab:inference_time},~\ref{tab:scr_performance}).
    SCR efficiency and performance metrics for sequential editing.
    Visual representation of the SCR's inference time and editing stability across various memory sizes (100 and 1000) and retrieval windows (top-1, top-3, top-5, and top-10).
    }
    \label{fig:scr_performance_resources}
\end{figure*}

Figure~\ref{fig:scr_performance_resources} illustrates SCR's resource efficiency and performance metrics, providing a visual summary of inference latency and editing reliability across all experimental settings.

\begin{table*}[h]
    \centering
    \caption{
    % For Llama-3.1-8B-Instruct, when the memory size is set to 100 and 1000 respectively, the VRAM consumption (in MB) for continuously editing 100 pieces of knowledge using SCR was calculated. As the memory size and top-k vary, the difference in VRAM consumption is minimal.
    VRAM consumption of SCR for sequential editing.
    VRAM usage (in MB) during 100 sequential edits using SCR on Llama-3.1-8B-Instruct under different memory and retrieval configurations.}
    \label{tab:vram_usage}
    \begin{tabular}{ccccc}
        \toprule
        Memory Size & Top-1 & Top-3 & Top-5 & Top-10 \\
        \midrule
        100  & 31123 & 31133 & 31145 & 31176 \\
        1000 & 31123 & 31133 & 31145 & 31177 \\
        \bottomrule
    \end{tabular}
\end{table*}

\begin{table*}[h]
    \centering
    \caption{
    % For Llama-3.1-8B-Instruct, when the memory size is set to 100 and 1000 respectively, the inference time (in seconds) for continuously editing 100 pieces of knowledge using SCR was calculated. As the memory size and top-k increase, the inference time becomes longer.
    Inference latency under varying memory and top-k retrieval sizes.
    Inference time (in seconds) for 100 sequential edits using SCR on Llama-3.1-8B-Instruct.
    }
    \label{tab:inference_time}
    \begin{tabular}{ccccc}
        \toprule
        Memory Size & Top-1 & Top-3 & Top-5 & Top-10 \\
        \midrule
        100  & 1.68 & 1.67 & 1.70 & 1.76 \\
        1000 & 1.74 & 1.75 & 1.72 & 1.81 \\
        \bottomrule
    \end{tabular}
\end{table*}

\begin{table*}[h]
    % \small
    \centering
    \caption{
    % For Llama-3.1-8B-Instruct, when the memory size is set to 100 and 1000 respectively, the edited effect for continuously editing 100 pieces of knowledge using SCR was calculated. The results show that SCR maintains stable performance across different memory sizes and retrieval top-k settings.
    Editing performance under varying retrieval parameters.
    Evaluation of editing effectiveness across different configurations. Performance metrics 
    % (reliability, generalization, locality, and portability) 
    show consistent results regardless of memory size or retrieval settings.
    }
    \label{tab:scr_performance}
    \resizebox{\linewidth}{!}{%
    \begin{tabular}{lcccccccccccccccc}
        \toprule
        & \multicolumn{4}{c}{Reliability} & \multicolumn{4}{c}{Generalization} & \multicolumn{4}{c}{Locality} & \multicolumn{4}{c}{Portability} \\
        \cmidrule(lr){2-5} \cmidrule(lr){6-9} \cmidrule(lr){10-13} \cmidrule(lr){14-17}
        Memory Size & Top-1 & Top-3 & Top-5 & Top-10 & Top-1 & Top-3 & Top-5 & Top-10 & Top-1 & Top-3 & Top-5 & Top-10 & Top-1 & Top-3 & Top-5 & Top-10 \\
        \midrule
        100  & 0.80 & 0.95 & 0.97 & 0.97 & 0.78 & 0.87 & 0.91 & 0.89 & 0.18 & 0.17 & 0.17 & 0.18 & 0.41 & 0.43 & 0.48 & 0.45 \\
        1000 & 0.76 & 0.86 & 0.91 & 0.88 & 0.69 & 0.80 & 0.80 & 0.76 & 0.18 & 0.17 & 0.19 & 0.17 & 0.40 & 0.45 & 0.45 & 0.47 \\
        \bottomrule
    \end{tabular}%
    }
\end{table*}

\end{document}